\documentclass[10pt, logo, copyright]{nvidiatechreport}

\usepackage{amsmath}
\usepackage{amssymb}
\usepackage{amsfonts}
\usepackage{amsthm}
\usepackage{mathtools}
\usepackage{bm}
\usepackage{bbm}
\usepackage{dsfont}
\usepackage{cases} 
\usepackage{nicefrac}
\usepackage{bbding}
\usepackage{pifont}

\usepackage[utf8]{inputenc} %
\usepackage[T1]{fontenc}    %
\usepackage{textcomp}

\usepackage{etoc}  %

\usepackage{graphicx}
\usepackage{wrapfig}
\usepackage{float}
\usepackage{caption}
\usepackage{subcaption}     %
\usepackage{array}
\usepackage{multirow}
\usepackage{multicol}
\usepackage{booktabs}
\usepackage{makecell}
\usepackage{colortbl}

\usepackage{xr-hyper}
\usepackage{hyperref}       %
\usepackage[capitalise]{cleveref}
\crefname{equation}{}{}
\crefname{figure}{Fig.}{Figs.}
\crefname{section}{Sec.}{Secs.}
\crefname{appendix}{App.}{Apps.}
\crefname{table}{Tab.}{Tabs.}
\usepackage{url}

\usepackage{tikz}
\usetikzlibrary{positioning, calc, matrix} 

\usepackage[makeroom]{cancel}

\hypersetup{
  final,
  colorlinks,
  linkcolor=ourred,
  citecolor=ourblue,
  urlcolor=url,
}

\usepackage{algorithm}
\usepackage{algpseudocode}
\usepackage[page,toc,titletoc,title]{appendix}
\usepackage{microtype}
\usepackage{xargs}
\usepackage{xspace}
\usepackage[normalem]{ulem}
\usepackage{soul}
\usepackage{fancyvrb}

\usepackage{lipsum}
\usepackage{blindtext}

\usepackage{xcolor}
\definecolor{ourblue}{rgb}{0.368,0.507,0.71}
\definecolor{ourorange}{rgb}{0.881,0.611,0.142}
\definecolor{ourgreen}{rgb}{0.56,0.692,0.195}
\definecolor{ourgreen2}{rgb}{0.46,0.792,0.195}
\definecolor{ourred}{rgb}{0.923,0.386,0.209}
\definecolor{ourviolet}{rgb}{0.528,0.471,0.701}
\definecolor{ourbrown}{rgb}{0.772,0.432,0.102}
\definecolor{ourlightblue}{rgb}{0.364,0.619,0.782}
\definecolor{ourbrightblue}{RGB}{93,135,237}
\definecolor{ourdarkgreen}{rgb}{0.572,0.586,0.}

\definecolor{ourcyan2}{rgb}{0.125,0.722,0.804}
\definecolor{ourred2}{rgb}{0.863,0.184,0.047}
\definecolor{ouryellow2}{cmyk}{0,0.16,1.0,0.07}
\definecolor{ourviolet2}{cmyk}{0.55,0.56,0,0.47}
\definecolor{ourorange2}{cmyk}{0,0.46,0.89,0.11}
\definecolor{grayseq}{RGB}{120,120,120}

\definecolor{discretecolor}{RGB}{11,83,150}
\definecolor{gaussiancolor}{RGB}{230,145,56}
\definecolor{argmaxcolor}{RGB}{154,0,0}

\definecolor{customred}{RGB}{169, 45, 59}
\definecolor{url}{HTML}{d95225}

\definecolor{olivegreen}{RGB}{165,185,115}
\definecolor{olivegreendark}{RGB}{145,165,95}

\definecolor{maroon}{RGB}{140, 45, 45}
\definecolor{bcolor}{RGB}{0, 110, 55}
\definecolor{nvgreen}{HTML}{77b900}

\newcommand{\B}[1]{\textbf{#1}}
\newcommand{\U}[1]{\underline{#1}}

\usepackage[most]{tcolorbox} %
\usepackage{soul}

\makeatletter
\DeclareRobustCommand\onedot{\futurelet\@let@token\@onedot}
\def\@onedot{\ifx\@let@token.\else.\null\fi\xspace}

\makeatother

\newtcolorbox{corollarybox}{
  breakable,
  enhanced,
  colback=ourbrightblue!12,
  colframe=ourbrightblue!80!black,
  left=1pt,
  right=1pt,
  top=1pt,
  bottom=1pt,
  boxrule=1pt
}

\newtcolorbox[auto counter]{findingbox}[1][]{
  breakable,
  enhanced,
  colback=ourbrightblue!8,
  colframe=ourbrightblue!80!black,
  left=2pt,
  right=2pt,
  top=1pt,
  bottom=1pt,
  boxrule=1pt,
  before skip=0.7em,
  after skip=0.7em,
  before upper={\textbf{Finding~\thetcbcounter: } },
  #1
}

\newtcolorbox{samplebox}[1][]{
  breakable,
  enhanced,
  colback=ourorange!6,
  colframe=ourorange!75!black,
  left=6pt, right=6pt, top=4pt, bottom=4pt,
  boxrule=0.6pt,
  arc=2.5pt,
  fonttitle=\bfseries\small,
  coltitle=black,
  colbacktitle=ourorange!22,
  attach boxed title to top left={yshift=-2.5mm, xshift=5mm},
  boxed title style={size=small, sharp corners=all, boxrule=0.3pt, arc=2pt, colframe=ourorange!75!black},
  fontupper=\small,
  title=Sample,
  #1
}

\newtcolorbox{exoutput}[1][]{
  breakable,
  enhanced,
  colback=olivegreen!14,
  colframe=olivegreen!70!black,
  left=6pt, right=6pt, top=6pt, bottom=4pt,
  boxrule=0.6pt,
  arc=2.5pt,
  fonttitle=\bfseries\footnotesize,
  coltitle=black,
  colbacktitle=olivegreen!32,
  attach boxed title to top left={yshift=-2.5mm, xshift=5mm},
  boxed title style={size=small, sharp corners=all, boxrule=0.3pt, arc=2pt, colframe=olivegreen!70!black},
  fontupper=\footnotesize,
  before skip=0.8em,
  after skip=0.8em,
  title={Example output},
  #1
}

\usepackage{pifont}

\definecolor{topone}{RGB}{112, 149, 239}   %
\definecolor{toptwo}{RGB}{174, 195, 246}   %
\definecolor{topthree}{RGB}{223, 231, 251} %

\definecolor{rankone}{RGB}{99, 137, 236}    %
\definecolor{ranktwo}{RGB}{140, 168, 242}   %
\definecolor{rankthree}{RGB}{174, 195, 246}  %
\definecolor{rankfour}{RGB}{201, 216, 249}  %
\definecolor{rankfive}{RGB}{224, 233, 252}  %

\usepackage{arydshln}

\usepackage{amsthm}
\usepackage{thmtools}
\usepackage{thm-restate}

\crefname{theorem}{Thm.}{Thms.}

\crefname{lemma}{Lem.}{Lems.}

\crefname{assumption}{Assump.}{Assumps.}

\crefname{proposition}{Prop.}{Props.}

\crefname{corollary}{Cor.}{Cors.}

\crefname{definition}{Def.}{Defs.}

\crefname{remark}{Rmk.}{Rmks.}
\crefname{algocf}{Alg.}{Algs.}
\crefname{algorithm}{Alg.}{Algs.}

\usepackage{titlesec}

\titlespacing*{\section}{0pt}{0.5\baselineskip}{0.2\baselineskip}
\titlespacing*{\subsection}{0pt}{0.3\baselineskip}{0.1\baselineskip}
\titlespacing*{\subsubsection}{0pt}{0.2\baselineskip}{0.1\baselineskip}
\titlespacing*{\paragraph}{0pt}{0pt}{0.5em}

\newcommand{\model}{\textit{Sigma}\xspace}
\newcommand{\nemmodel}{\textit{Nemotron-Labs-Sigma}\xspace}

\newcommand{\unreg}{-\textit{unreg}}
\newcommand{\preg}{-\textit{preg}}
\newcommand{\distill}{-\textit{distill}}
\newcommand{\instruct}{-\textit{instruct}\xspace}
\xspaceaddexceptions{\unreg,\preg,\distill,\instruct}

\usepackage{minted}
\usepackage{listings}

\definecolor{elfcomment}{HTML}{3B9C9C} %
\definecolor{elfblue}{HTML}{1A237E}    %
\definecolor{elfmagenta}{HTML}{C2185B} %
\definecolor{elforange}{HTML}{E65100}  %

\lstdefinelanguage{ELF}{
    language=Python,
    basicstyle=\ttfamily\small\color{black},
    commentstyle=\color{elfcomment},
    stringstyle=\color{elforange},
    columns=fullflexible,                  %
    keepspaces=true,                       %
    showstringspaces=false,
    keywordstyle=[1]\color{elfblue},
    morekeywords=[1]{if, else, for, in, range, len},
    keywordstyle=[2]\color{elfmagenta},
    morekeywords=[2]{encode, uniform, sample_t, randn_like, net, mse_loss, corrupt, unembed, ce_loss, randn, argmax, adaptive_bs, gamma_tilde_net, gamma_bounds, sqrt, sigmoid, prior_loss},
    literate=
     *{=}{{\textcolor{elfblue}{=}}}{1}
      {+}{{\textcolor{elfblue}{+}}}{1}
      {-}{{\textcolor{elfblue}{-}}}{1}
      {<}{{\textcolor{elfblue}{<}}}{1}
      {*}{{\textcolor{elfblue}{*}}}{1}
      {/}{{\textcolor{elfblue}{/}}}{1}
      {(t = 1)}{{\textcolor{elfcomment}{(t = 1)}}}{7}, 
}

\usepackage{xcolor}

\newcommand{\algsection}[2]{%
  \Statex
  \colorbox{#1!8}{%
    \parbox{\dimexpr\linewidth-2\fboxsep\relax}{%
      \strut\textbf{#2}\strut
    }%
  }%
}

\usepackage{amsmath,amsfonts,bm}

\def\eqref#1{equation~\ref{#1}}

\def\1{\bm{1}}

\DeclareMathAlphabet{\mathsfit}{\encodingdefault}{\sfdefault}{m}{sl}
\SetMathAlphabet{\mathsfit}{bold}{\encodingdefault}{\sfdefault}{bx}{n}

\DeclareMathOperator{\E}{\mathbb{E}}
\newcommand{\unif}{\operatorname{Unif}}
\newcommand{\SNR}{\operatorname{SNR}}

\newcommand{\R}{\mathbb{R}}

\newcommand{\softmax}{\operatorname{softmax}}
\newcommand{\sigmoid}{\sigma}

\newcommand{\KL}{\operatorname{KL}}

\DeclareMathOperator*{\argmax}{argmax}

\renewcommand{\c}{{\mathbf c}}
\newcommand{\x}{{\mathbf x}}
\renewcommand{\a}{\mathbf{a}}
\renewcommand{\i}{\mathrm{i}}

\newcommand{\ee}{{\mathrm{e}}}
\newcommand{\zero}{{\mathbf 0}}
\newcommand{\one}{{\mathbf 1}}

\newcommand{\z}{{\mathbf z}}
\newcommand{\e}{{\mathbf e}}
\newcommand{\p}{{\mathbf p}}      %
\renewcommand{\r}{{\mathbf r}}    %
\renewcommand{\c}{{\mathbf c}}    %

\newcommand{\I}{{\mathbf I}}
\newcommand{\Emb}{{\mathbf E}}
\newcommand{\beps}{{\boldsymbol{\epsilon}}}
\newcommand{\bell}{{\boldsymbol{\ell}}}

\renewcommand{\d}{{\mathrm{d}}}        
\newcommand{\cL}{\mathcal{L}}
\newcommand{\cV}{\mathcal{V}}
\newcommand{\mask}{{\mathbf m}}   %
\newcommand{\selfcond}{\text{sc}}

\newcommand{\sg}{\operatorname{sg}}

\newcommand{\tsum}{{\textstyle\sum}}

\newcommand{\bcolor}[1]{\textcolor{bcolor}{#1}}

\usepackage[utf8]{inputenc} 
\usepackage[T1]{fontenc}    
\usepackage{hyperref}       
\usepackage{url}            
\usepackage{booktabs}       
\usepackage{amsfonts}       
\usepackage{nicefrac}       
\usepackage{microtype}      
\usepackage{xcolor}         
\usepackage{enumitem} 
\setlist[itemize,enumerate]{leftmargin=1.5em, topsep=1pt, itemsep=2pt, parsep=0pt}
\usepackage{natbib}
\usepackage{fancyhdr}

\hypersetup{
  colorlinks=true,
  linkcolor=OliveGreen,
  citecolor=OliveGreen,
  urlcolor=OliveGreen,
  filecolor=OliveGreen,
  menucolor=OliveGreen,
  anchorcolor=OliveGreen,
}

\title{Large Language Continuous Diffusion Models}

\author{%
\textbf{Zhihan Yang}$^{*,\dagger,1,2}$ \quad
\textbf{Wei Guo}$^{*,\dagger,1,3}$ \quad
\textbf{Jean-Marie Lemercier}$^{*,1}$ \quad
\textbf{Simon Welker}$^{*,1}$
\authorcr
\textbf{Mohammad Mahdi Kamani}$^{*,1}$ \quad
\textbf{Yonggan Fu}$^1$ \quad
\textbf{Sajad Norouzi}$^1$ \quad
\textbf{Julius Berner}$^1$
\authorcr
\textbf{Tomas Geffner}$^1$ \quad
\textbf{Karsten Kreis}$^1$ \quad
\textbf{Yongxin Chen}$^3$ \quad
\textbf{Molei Tao}$^3$ \quad
\textbf{John Thickstun}$^2$
\authorcr
\textbf{Pavlo Molchanov}$^1$ \quad
\textbf{Ante Jukić}$^1$ \quad
\textbf{Arash Vahdat}$^{\star,1}$ \quad
\textbf{Morteza Mardani}$^{\star,1}$
\\
$^1$NVIDIA \quad
$^2$Cornell University \quad
$^3$Georgia Institute of Technology \quad
$^{\dagger}$Work done during internship at NVIDIA.\\
$^{*}$Core contributors, equal contribution. \quad
$^\star$Equal supervision. \quad
Correspondence to:
\href{mailto:mmardani@nvidia.com}{\texttt{mmardani@nvidia.com}}
}

\newcommand{\se}[2]{#1\textsubscript{\scriptsize\textcolor{gray}{\ ($\pm$\ #2)}}}
\newcommand{\sev}[2]{\begin{tabular}[t]{@{}c@{}}#1\\[-2pt]{\scriptsize\textcolor{gray}{$\pm$\ #2}}\end{tabular}}

\begin{document}

\etocdepthtag.toc{main}

\begin{abstract}
Despite the success of discrete diffusion language models (dLMs) for fast parallel decoding, their non-smooth, high-dimensional space hinders trajectory steering for reasoning and inference acceleration. To overcome this, we present \nemmodel, the first family of large-scale (3B/8B) continuous dLMs built on steerable, low-dimensional ODE/SDE latent trajectories. Trained blockwise via likelihood optimization, \nemmodel jointly denoises Gaussian-corrupted token embeddings while learning an optimal embedding geometry, and leverages pretrained autoregressive (AR) weights for warm-starting. At inference, we identify classifier-free guidance and score temperature as essential for high-fidelity reasoning and coding. Against state-of-the-art discrete counterparts (masked dLMs and AR baselines), \nemmodel outperforms them on standard benchmarks after pretraining (e.g., GSM8K, Minerva, HumanEval, MBPP) and on challenging reasoning tasks after supervised fine-tuning (e.g., MATH-500, AIME), with the largest consistent gains on coding. Beyond accuracy, we exploit two structural properties unique to continuous dLMs. First, smooth trajectories degrade gracefully at low NFEs and enable efficient few-step distillation: distribution matching proves strongest among the methods we study, giving roughly $8\times$ acceleration on math and $4\times$ on coding ($8$ and $4$ tokens per forward pass) with negligible accuracy loss and preserved diversity. In addition, the one-step speculative decoding on the (pre-SFT) distilled 8B model yields a further $\sim\! 3.5\times$ throughput gain over the AR counterpart (on GB200), outperforming masked dLM baselines. Second, embedding-space steering controls the quality--diversity trade-off, yielding strong pass@$k$ that keeps improving as $k$ grows. Together, these results establish continuous dLMs as a promising paradigm for efficient language generation. Model family weights are publicly available at \href{https://huggingface.co/collections/nvidia/nemotron-labs-sigma}{\raisebox{-0.25ex}{\includegraphics[height=1em]{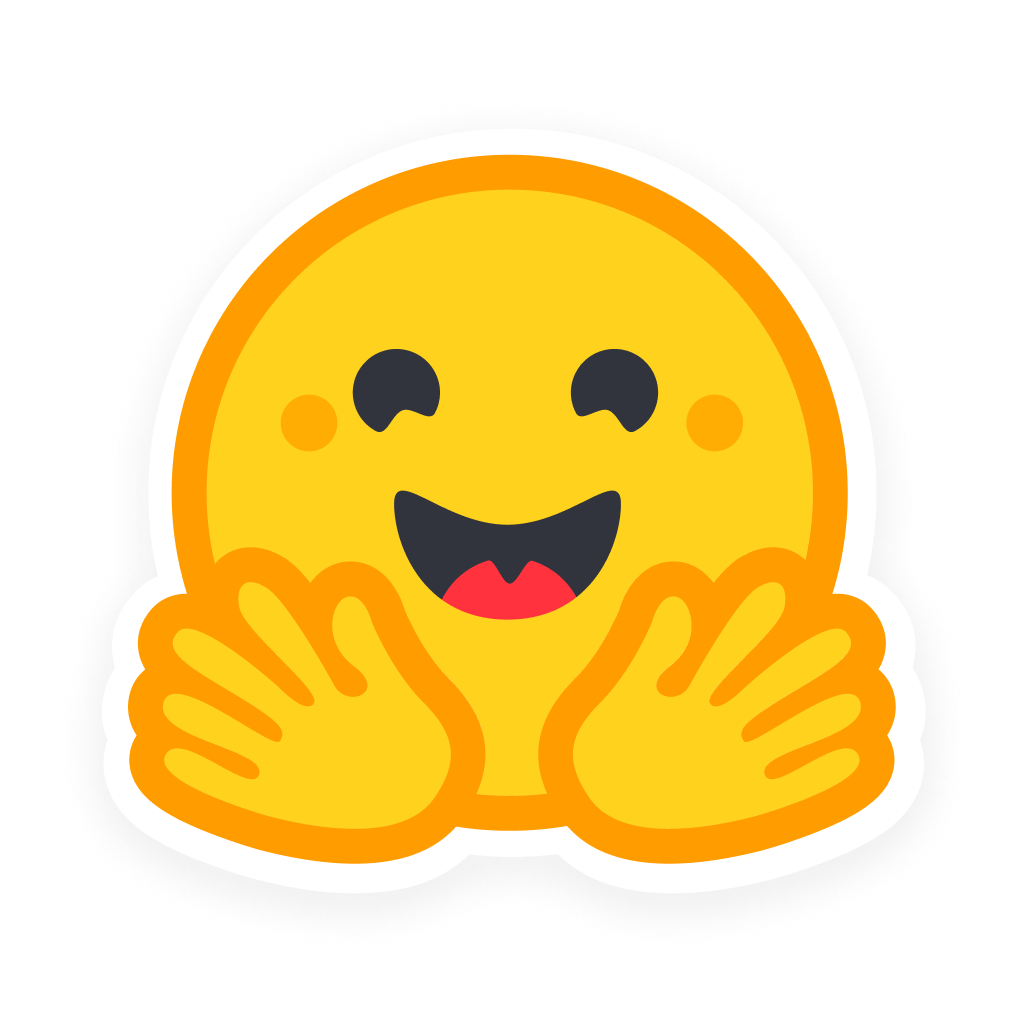}}~\textit{Nemotron-Labs-Sigma}}.
\end{abstract}

\maketitle

\section{Introduction}
\label{sec:introduction}

While diffusion language models (dLMs) enable fast parallel decoding compared to autoregressive (AR) baselines \citep{khanna2025mercury,song2025seed,fu2026nemotron,zhu2026flare,diffusiongemma}, dominant discrete dLMs operate in high-dimensional, categorical state spaces (e.g., vocabulary size $V\ge100\text{k}$) that suffer under limited step budgets and impede trajectory control for complex reasoning and coding tasks. Continuous diffusion offers a principled way to overcome these structural limitations through smooth latent geometries that natively support structured editing \citep{meng2022sdedit}, flexible ODE trajectory steering \citep{zhang2023fast, lu2022dpmsolver,zhang2023gddim}, and efficient few-step distillation \citep{salimans2022progressive,song2023consistency}.

Recently, there has been growing interest in continuous diffusion and flow matching for text generation \citep{gulrajani2023plaid, dieleman2022continuous, chen2026langflow, hu2026elf,  lee2026flow, guo2026continuous, liang2026aurora, yang2026continuous}; see~\cref{app:related-work} for a review. However, existing methods remain limited in scale ($<2\text{B}$ parameters) or focus on \textit{unconditional} generation, lacking evaluations on complex reasoning tasks. Whether continuous dLMs scale to complex conditional reasoning remains open. This is a non-trivial hurdle because continuous models update block embeddings globally, in contrast to the factorized categorical denoising of discrete dLMs that leads to semi-autoregressive token commitment per block.

To overcome these limitations, we present \nemmodel (3B/8B): \textbf{s}cal\textbf{i}ng up \textbf{G}aussian diffusion \textbf{m}odels for l\textbf{a}nguage. For brevity, we refer to the model family simply as \model throughout the remainder of the paper. We propose a family of continuous dLMs operating over a 16-dimensional embedding space trained blockwise via likelihood optimization. Within each block, the model learns to jointly denoise Gaussian-corrupted token embeddings and to structure an optimal embedding geometry. To scale training, we warm-start the model from existing pretrained AR weights. For inference, we explore diffusion trajectory sampling, identifying embedding-space Classifier-Free Guidance (CFG) and Score Temperature (ST) as crucial ingredients for high-fidelity reasoning and coding.

\begin{figure}[t]
\centering
    \begin{minipage}[t]{0.49\linewidth}
        \centering
        \includegraphics[width=\linewidth]{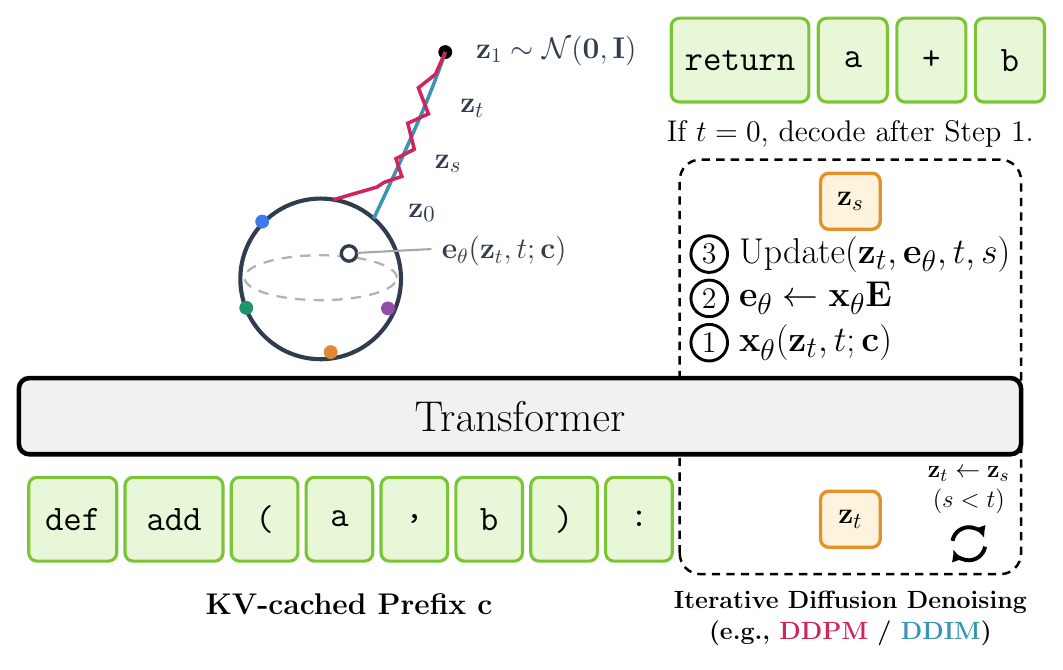}
    \end{minipage}\hfill
    \begin{minipage}[t]{0.49\linewidth}
        \centering
        \includegraphics[width=\linewidth]{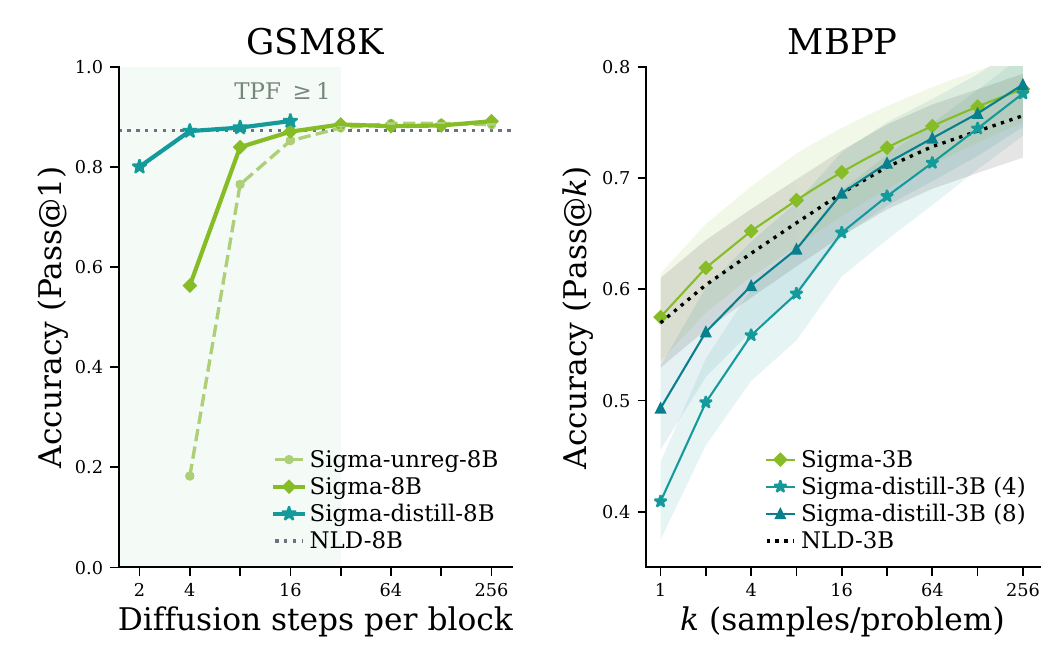}
    \end{minipage}
\caption{ \textbf{Left: Inference pipeline.} \model generates text blockwise via iterative diffusion denoising over 16D continuous latents ($\mathbf{z}_t$) conditioned on a KV-cached prefix ($\mathbf{c}$), followed by categorical readout. Predicted vocabulary probabilities $\mathbf{x}_\theta(\mathbf{z}_t,t;\mathbf{c})$ map to mean predictions $\mathbf{e}_\theta(\mathbf{z}_t,t;\mathbf{c})$ in 16D space via categorical reparameterization. Clean embeddings (dots) lie on the 16D unit sphere, while $\mathbf{e}_\theta$ lies on or within it. \textbf{Middle: Smooth trajectory enables low-NFE reasoning.} GSM8K accuracy vs. NFEs using DDIM sampler for \model\!\!(\unreg)-8B, and sampler \cref{eq:dmd-update} for \model\distill-8B. \textbf{Right: Scaling inference-time compute.} pass@$k$ vs. $k$ on MBPP (solid curves: fixed config across $k$; thin curves: alternative tested configs). }

\label{fig:teaser-inference}
\end{figure}

Our key contributions are summarized as follows:
\begin{itemize}

    \item \textbf{First Large-Scale Continuous dLM:} We present \nemmodel (3B/8B), efficiently trained by warm-starting from pretrained AR weights, and optimizing a regularized continuous diffusion ELBO.

    \item \textbf{Competitive Math \& Coding:} Across six base and ten instruct benchmarks, \model outperforms state-of-the-art discrete dLMs/ARs on reasoning and coding.

    \item \textbf{Inference Steering (CFG \& ST):} By elucidating the inference design space, we show that embedding-space CFG and ST effectively govern the quality-diversity trade-off with minimal latency overhead via single-pass block-causal attention packing.

    \item \textbf{Distillation:} We study trajectory- and distribution-matching distillation (DMD) methods and find that phased DMD delivers the strongest acceleration, speeding up inference by $8\times$ for math and $4\times$ for coding, reaching the equivalent of 8 and 4 tokens per forward (TPFs), respectively.

    \item \textbf{Speculative Decoding:} We apply speculative decoding to both distilled and non-distilled \model~variants, and find that combining distillation with one-step speculative decoding is highly effective, yielding a $\sim\!3\times$ throughput improvement over AR on an NVIDIA GB200 GPU.

\end{itemize}

\section{Background}
\label{sec:prelim-cdlm}

\paragraph{Notation.} Let $\x$ be a sequence of length $L$ over a vocabulary $\cV$ of size $V$. We identify $\x$ with a matrix in $\{0,1\}^{L \times V}$ whose rows $\x^1, ..., \x^L$ are one-hot vectors in $\R^V$. $\Delta^V$ denotes the probability simplex over $\cV$. $\langle\cdot,\cdot\rangle$ denotes the inner product. NELBO stands for negative evidence lower bound.

\paragraph{Plaid.} Continuous dLMs lift discrete tokens into a low-dimensional Euclidean space and run Gaussian diffusion. There have been multiple proposed design choices, see~\cref{app:related-work} for an overview.
We adopt the embedding and diffusion process from Plaid \citep{gulrajani2023plaid}, a Variational Diffusion Model (VDM, \citet{kingma2021variational}) over token embeddings.
Let $\Emb \in \R^{V \times d_e}$ be a learnable embedding matrix whose rows lie on the unit sphere, with $d_e\ll V$.
The embedding of $\x$ is given by $\e := \x\Emb \in \R^{L \times d_e}$ (i.e., standard embedding lookup). The forward process applies variance-preserving Gaussian noise on $\e$:
\begin{align}\label{eq:vdm-forward}
    q(\z_t \mid \x) = \mathcal{N}(\alpha_t \e,\ \sigma_t^2 \I), \qquad \alpha_t, \sigma_t > 0, \qquad \alpha_t^2 + \sigma_t^2 = 1, \qquad t\in[0,1],
\end{align}
with $\SNR(t) = \alpha_t^2/\sigma_t^2$ decreasing monotonically over $t \in [0,1]$. A denoiser $\x_\theta(\z_t, t) \in (\Delta^V)^L \subset \R^{L \times V}$ with output prior returns a categorical distribution over the vocabulary at every position. The reverse process is the exact Gaussian posterior $q(\z_s \mid \z_t, \x)$ \citep[App. A.2]{kingma2021variational} with $\x$ replaced by $\x_\theta(\z_t, t)$. Its NELBO splits into 3 terms: prior KL $\cL_\text{prior}$, reconstruction loss $\cL_\text{rec}$, and diffusion loss $\cL_\text{diff}$:
\begin{align}
-\log p_\theta(\x) \leq \cL_{\text{NELBO}}(\x):=&\KL(q(\z_1 \mid \x) \,\|\, p(\z_1)) + \E_{\z_0 \sim q(\z_0 \mid \x)}\big[\tsum_{l=1}^L- \log \langle \x_\theta^l(\z_0, 0), \x^l \rangle\big] \nonumber\\
& -\tfrac{1}{2}\ \E_{t\sim\unif[0,1]}\E_{\z_t\sim q(\z_t \mid \x)}[\SNR'(t) \| \e_\theta(\z_t, t) - \e \|^2], \label{eq:plaid-nelbo}
\end{align}
where $p(\z_1)=\mathcal{N}(\zero,\I)$ and the \textit{categorical reparameterization} $\e_\theta(\z_t, t) := \x_\theta(\z_t, t) \Emb \in \R^{L\times d_e}$ predicts the posterior mean embedding $\E[\e \mid \z_t, t]$ corresponding to $\x_\theta(\z_t, t)$.

\paragraph{Learned noise schedule.} The schedule admits
$\gamma(t) = \gamma_0 + (\gamma_1 - \gamma_0)\tilde{\gamma}(t)$ with learnable endpoints
$\gamma_0 < \gamma_1$ and a monotone interior shape $\tilde{\gamma} \in [0,1]$, giving
 $\alpha_t^2 = \sigmoid(-\gamma(t))$, $\sigma_t^2 = \sigmoid(\gamma(t))$ and
$\SNR(t) = \ee^{-\gamma(t)}$.\footnote{$\sigma(t)=1/(1+\ee^{-t})$ is the sigmoid function.} The diffusion term depends on the endpoints alone and is invariant to $\tilde{\gamma}(t)$, which instead controls its Monte-Carlo (MC) variance~\citep{kingma2021variational}, so
they are learned for different reasons. Learning the interior-shape distributes denoising difficulty over $t$~\citep{yang2026continuous}.

\paragraph{Learned embeddings.} The
NELBO~\cref{eq:plaid-nelbo} depends on $\Emb$ via~\cref{eq:vdm-forward}, where embeddings become likelihood-trained parameters, which separates Plaid from continuous dLMs built on fixed (e.g., one-hot)
encodings. Not learning embeddings significantly worsens Plaid's performance~\citep{yang2026continuous}.

\paragraph{Self-conditioning (SC).}
Proposed in~\citet{chen2023analog}, SC lets the denoiser see its own previous estimate. Let $\x_\theta(\z_t, t; \e_{\selfcond})$ denote a denoiser taking an additional predicted clean embedding $\e_{\selfcond}\in\R^{L\times d_e}$, and $\zero$ for the empty condition (omitted by default in our notation). During training, $25\%$ of training samples are self-conditioned: a first pass without gradient tracking produces $\e_{\selfcond} := \x_\theta(\z_t, t; \zero)\Emb$, and only the second pass, $\x_\theta(\z_t, t; \sg(\e_{\selfcond}))\Emb$, receives gradients. The remaining samples use $\zero$ as SC input, so the denoiser learns both branches. During validation, all samples enable SC. Since $\e_{\selfcond}$ is itself a function of $(\z_t, t)$, the NELBO remains valid. Notice that during training, $25\%$ SC increases the non-embedding FLOPs by only about $8\%$, which is minimal. Furthermore, during inference, one directly feeds the previous-step's prediction as SC, which incurs no overhead.

\paragraph{Scaling.} RePlaid~\citep{yang2026continuous} realigns Plaid with the architectures of modern
discrete dLMs/ARs and finds that, once aligned, continuous diffusion scales competitively with masked diffusion.

\section{Large Language Continuous Diffusion Models}
\label{sec:llaconda}

\begin{figure*}[t]
    \centering
    \begin{subfigure}[b]{0.45\linewidth}
        \centering
        \includegraphics[width=\linewidth]{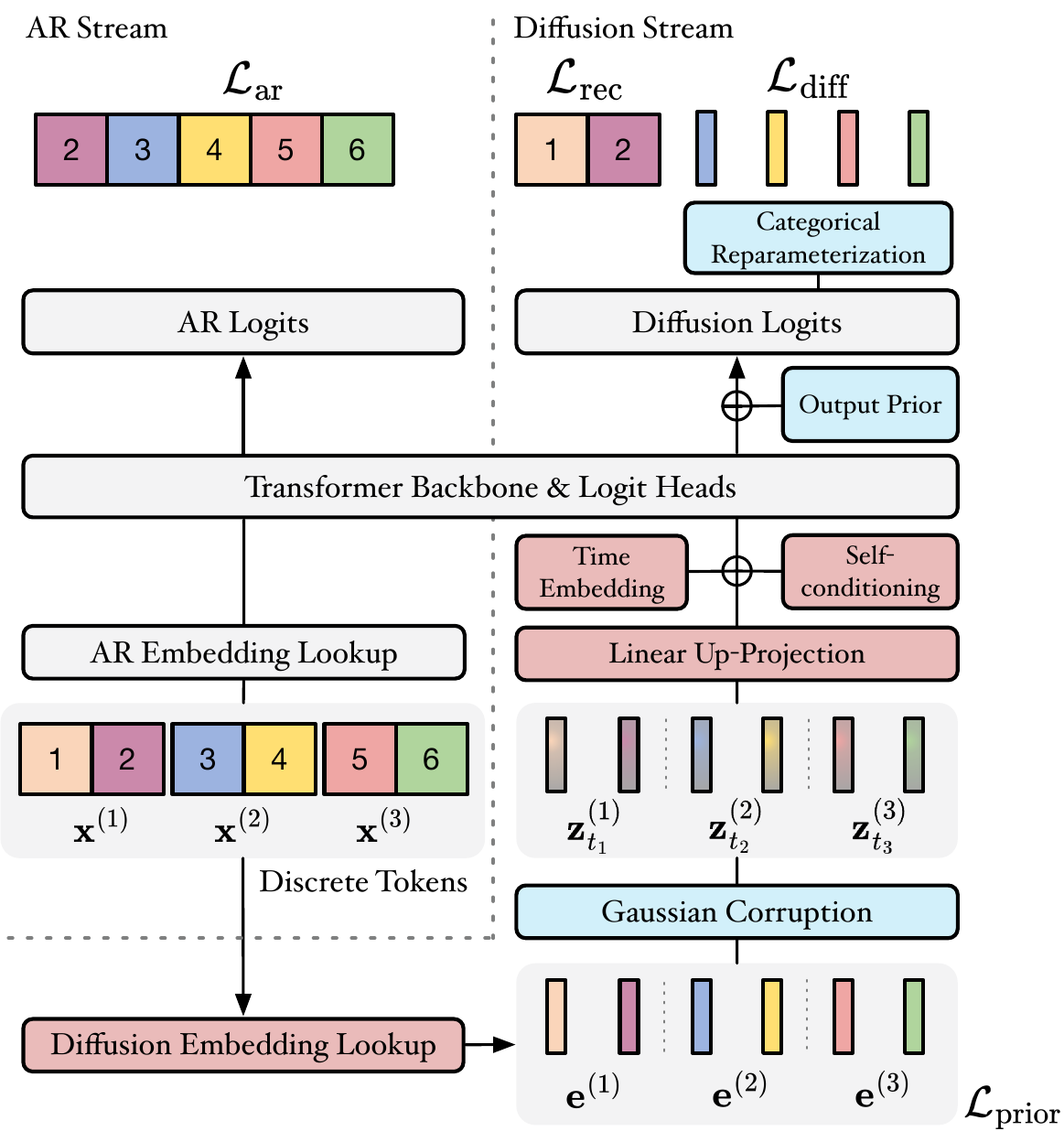}
        \caption{ Training procedure of \model.}
        \label{fig:teaser-pretrain}
    \end{subfigure}
    \begin{subfigure}[b]{0.4\linewidth}
        \centering
        \includegraphics[width=\linewidth]{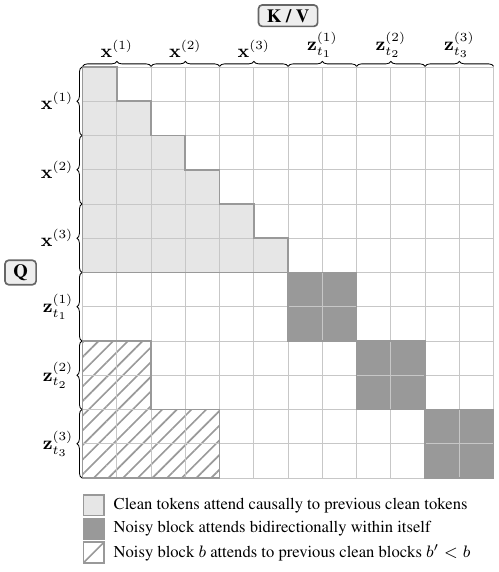}
        \caption{ Attention mask for training.}
        \label{fig:teaser-pretrain-attention}
    \end{subfigure}
    \caption{ \textbf{Training procedure of \model (a) and its transformer attention mask (b)} for an example sequence $\x$ of length $L=6$ with block size $L'=2$.
    During pretraining, we feed the clean discrete tokens $\mathbf{x}$ to the AR stream (a, left part) and the noisy continuous embeddings $\mathbf{z}$ to the diffusion stream (a, right part). 
    For the diffusion stream, each block is randomly chosen to either compute the reconstruction loss (1st block, $t=0$) or the diffusion loss (2nd \& 3rd blocks, $t\sim\unif[0,1]$); see~\cref{eq:ht}.
    Reconstruction loss is the cross entropy between the logits and the ground truth tokens, while for diffusion loss we use categorical reparameterization to obtain the mean embedding prediction and compute the mean squared error against the ground truth diffusion embeddings.
    }
    \label{fig:pretrain}
\end{figure*}

\subsection{Blockwise Probabilistic Model}
\label{sec:probabilistic}

Inspired by Nemotron-Labs-Diffusion (NLD, \citet{fu2026nemotron}), a state-of-the-art open-sourced masked dLM, we adopt a blockwise formulation for our continuous diffusion language model: the sequence $\x$ is partitioned into $B$ contiguous blocks $\{\x^{(\bcolor{b})}\}_{\bcolor{b}=1}^B$ of size $L'=L/B$, where the model denoises one block at a time given its clean prefix. As in \citet{arriola2025block}, we factorize the likelihood autoregressively \emph{across} blocks and then bound each factor: %
\begin{align}\label{eq:block-nelbo}
    -\log p_\theta(\x)
    &= \tsum_{b=1}^{B} -\log p_\theta(\x^{(\bcolor{b})} \mid \x^{(\bcolor{<b})})
    \leq \tsum_{b=1}^{B} \cL_{\text{NELBO}}(\x^{(\bcolor{b})}; \x^{(\bcolor{<b})})
    =: \cL_{\text{BD}}(\x),
\end{align}
where we obtain $\cL_{\text{NELBO}}(\x^{(\bcolor{b})}; \x^{(\bcolor{<b})})$ by conditioning the denoiser on $\x^{(\bcolor{<b})}$ as follows:
\begin{align}\label{eq:nelbo}
\cL_{\text{NELBO}}(\x^{(\bcolor{b})}; \x^{(\bcolor{<b})}) &= \KL(q(\z_1^{(\bcolor{b})} \mid \x^{(\bcolor{b})}) \,\|\, p(\z_1^{(\bcolor{b})})) \nonumber\\
& + \E_{q(\z_0^{(\bcolor{b})} \mid \x^{(\bcolor{b})})}\big[\tsum_{l=1}^{L'}
- \log \langle \x_\theta^{l}(\z_0^{(\bcolor{b})}, 0; \x^{(\bcolor{<b})}), \x^{(\bcolor{b,l})} \rangle\big] \nonumber\\
& -\tfrac{1}{2}\E_{t\sim\unif[0,1]}\E_{q(\z_t^{(\bcolor{b})} \mid \x^{(\bcolor{b})})}[\SNR'(t) \| \e_\theta(\z_t^{(\bcolor{b})}, t; \x^{(\bcolor{<b})}) - \e^{(\bcolor{b})} \|^2].
\end{align}
\paragraph{Training.} The reconstruction loss at $t=0$ and diffusion loss at $t\in(0,1)$ do not share a single forward pass. RePlaid evaluates one of them per $\x^{(\bcolor{b})}$ via Horvitz--Thompson estimator~\citep{horvitz1952generalization} with $R \sim \mathrm{Ber}(p_\text{rec})$, i.e., $\x^{(\bcolor{b})}$ is assigned to $\cL_{\text{rec}}$ w.p. $p_{\text{rec}}\in (0,1)$:
\begin{align}\label{eq:ht}
    \cL_{\text{NELBO}}(\x^{(\bcolor{b})}; \x^{(\bcolor{<b})}) = \E_{R\sim \mathrm{Ber}(p_\text{rec})} \big[
    \cL_{\text{prior}}(\x^{(\bcolor{b})})
    +\tfrac{R}{p_{\text{rec}}}\cL_\text{rec}(\x^{(\bcolor{b})}; \x^{(\bcolor{<b})})
    +\tfrac{1-R}{1 - p_{\text{rec}}}\cL_\text{diff}(\x^{(\bcolor{b})}; \x^{(\bcolor{<b})})
    \big]. 
\end{align}
Notice that this is an unbiased estimator if $\cL_\text{diff}$ and $\cL_\text{rec}$ are unbiased. We set $p_{\text{rec}} = 0.1$, empirically found to minimize the training objective variance of RePlaid.

To construct the unbiased estimator for $\cL_{\text{BD}}(\x)$ in \cref{eq:block-nelbo} for training, we sample $R^{(\bcolor{b})}$ i.i.d. for each block. If $R^{(\bcolor{b})}=0$, we sample $t^{(\bcolor{b})}$ and $\z_{t^{(\bcolor{b})}}^{(\bcolor{b})}$; otherwise, we set $t=0$ and sample $\z_0^{(\bcolor{b})}$. The noise schedule endpoints are learned to optimize $\cL_{\text{NELBO}}(\x^{(\bcolor{b})}; \x^{(\bcolor{<b})})$ while the schedule interior shape is learned to optimize its MC variance, with both shared across blocks. As mentioned before, we apply SC only to $25\%$ of samples (full sequence with all blocks). See~\cref{alg:train} for the training algorithm.

\subsection{Pretraining}\label{sec:pretrain}
NLD utilizes AR initialization and regularization to enhance downstream performance~\citep{fu2026nemotron}. However, adapting these techniques to continuous dLMs like \model is non-trivial, as the activation dynamics of continuous and discrete models differ fundamentally; it is thus unclear whether discrete AR initialization remains as effective for continuous representation space. To establish a rigorous baseline, we first align \model's data, architecture, and optimization with discrete dLMs for a controlled comparison at scale (\cref{subsec:unified-recipe}). In~\cref{sec:pretrain-from-scratch}, we evaluate the impact of AR initialization when pretraining \model\unreg-3B%
\footnote{``\textit{unreg}'' means no regularization loss for the diffusion embedding $\Emb$. This additional loss will be introduced in \cref{sec:few-step-susreg}.}
with \cref{eq:block-nelbo}, followed by an investigation of AR regularization for training our flagship 3B/8B models.

\begin{figure*}[t]
    \centering
    \begin{subfigure}{0.32\linewidth}
        \centering
        \includegraphics[width=\linewidth]{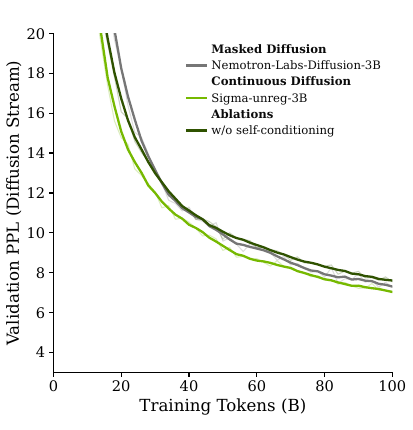}
        \caption{\centering From scratch + no AR reg.\\Block size: 64}
        \label{fig:scratch-100b}
    \end{subfigure}
    \begin{subfigure}{0.32\linewidth}
        \centering
        \includegraphics[width=\linewidth]{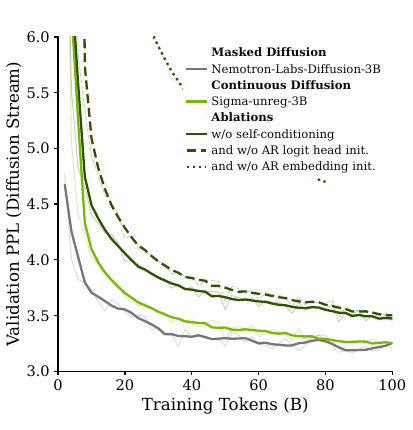}
        \caption{\centering From AR init. + no AR reg.\\Block size: 64}
        \label{fig:aradapt-100b}
    \end{subfigure}
    \begin{subfigure}{0.31\linewidth}
        \centering
        \includegraphics[width=\linewidth]{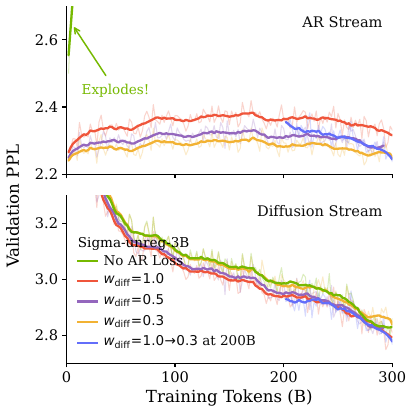}
        \caption{\centering From AR init. + AR reg.\\Block size: 32}
        \label{fig:dlm-weight-schedule}
    \end{subfigure}
    \caption{ \textbf{Pretraining \model\unreg-3B}: (a) from scratch, (b) from AR initialization, and (c) from AR initialization with AR regularization. PPL for AR and diffusion streams is computed via $\cL_{\text{ar}}(\x)$ in~\cref{eq:ar_loss} and $\cL_{\text{BD}}(\x)$ in~\cref{eq:block-nelbo}, respectively. Early investigations (a, b) use block size 64; our flagship models (c) use block size 32 to match the inference block size of NLD. Proper AR initialization/regularization leads to the best \model model.
    }
    \label{fig:pretrain-ablations}
\end{figure*}

\subsubsection{Unified Pretraining of Continuous and Discrete dLMs}\label{subsec:unified-recipe}

\paragraph{Data.} As with NLD, we adopt the open-sourced pretraining data from~\citet{basant2025nvidia}. We use the Mistral-NeMo tokenizer ($V = 131{,}072$) ~\citep{mistral_nemo_2024}, batch size $512$, and $L=4096$.

\paragraph{Architecture.} \model adopts the transformer backbone of NLD and its AR baseline, ensuring the three families are parameter- and FLOP-matched. As with NLD, \model uses a block causal attention mask (\cref{fig:teaser-pretrain-attention}). \model uniquely adds a unit-norm 16-dimensional diffusion embedding table with a linear up-projection, a categorical reparameterization mapping $V$-dimensional probabilities to low-dimensional soft embeddings, a self-conditioning linear projection, Fourier time conditioning, and a learnable noise schedule (see \cref{sec:prelim-cdlm}). Together, these account for less than $1\%$ of the total parameters. We provide thorough details in \cref{app:pretrain}.

\paragraph{Training.} We train our models for 150k steps, following the optimization procedure detailed in \cref{app:pretrain}. Given a batch size of 512, 150k steps correspond to over 300B tokens. Training consumed $\sim\!9{,}600$ and $\sim\!19{,}200$ NVIDIA GB200 GPU hours for the 3B and 8B models, respectively.

\subsubsection{AR Initialization and Regularization}
\label{sec:pretrain-from-scratch}

\paragraph{AR Initialization.} To evaluate the impact of initialization, we compare pretraining \model (\textit{i}) from scratch and (\textit{ii}) from an AR baseline pretrained on 1T tokens. Due to the dimension mismatch between high-dimensional AR embeddings and \model's 16-dimensional space, we randomly initialize the diffusion embedding layer. As shown in \cref{fig:scratch-100b,fig:aradapt-100b}, \model\unreg-3B ($L'=64$) with self-conditioning achieves competitive perplexity relative to NLD-3B ($L'=64$) under both regimes (\textit{i}) and (\textit{ii}). However, AR initialization significantly accelerates convergence and yields lower final training loss.

\paragraph{AR Regularization.} Continuous diffusion training inherently suffers from high gradient variance and slow convergence. To mitigate variance and accelerate optimization, we follow NLD by adding AR regularization, optimizing the joint objective $\cL_{\text{ar}} + w_{\text{diff}} \cL_{\text{BD}}$ (\cref{fig:dlm-weight-schedule}; see \cref{app:pretrain} for the $w_{\text{diff}}$ weighting schedule), where
\begin{equation}\label{eq:ar_loss}
    \cL_{\text{ar}} = 
    \tsum_{l=2}^{L} 
    - \log \langle \x_{\theta}^l(\x^{<l}) , \x^l \rangle
\end{equation}
is computed over the AR stream using causal attention (\cref{fig:pretrain}). This architecture employs two distinct readout heads, with the AR regularization computed exclusively from the AR head. While this dual-head formulation improves auxiliary AR decoding capability, our primary objective remains maximizing diffusion decoding performance. 

Finally, input sequences utilize AR embeddings for the AR stream and 16-dimensional diffusion embeddings for the diffusion stream (\cref{fig:pretrain}), facilitating AR initialization and efficient KV-caching. As shown in \cref{fig:dlm-weight-schedule}, AR regularization significantly improves training stability; with a properly tuned $w_{\text{diff}}$, it even outperforms pure perplexity optimization for the diffusion stream. This confirms that AR regularization stabilizes training while yielding a higher-quality model. Consequently, we adopt AR initialization, AR regularization, and a block size of $L'=32$ for all subsequent \model experiments.

\section{Language Generation via Continuous Diffusion Sampling}
\label{sec:inference}
Drawing on the success of continuous diffusion samplers using SDE/ODE solvers~\citep{song2021scorebased}, we systematically explore decoding samplers for continuous dLMs, evaluating performance across standard mathematical reasoning and coding benchmarks.

\subsection{Block Diffusion Sampling}
\label{sec:inference-solver}

To generate a response $\r$ given a prompt $\p$, we denoise one block at a time.
Let $\c := (\p, \r^{(\bcolor{<b})})$ be the clean context available when generating block $b$, stored in an attention key-value (KV) cache using the AR embeddings.
An offset block-causal mask determines who can see whom (\cref{fig:cfg-attention}).
Positions in $\c$ use causal attention, so their KV can be cached and reused across diffusion steps, regardless of whether the prompt length is a multiple of $L'$. Positions in the noisy block $\z_t^{(\bcolor{b})}$ attend bidirectionally within the block and causally to the entire prefix $\c$. After a block is decoded, the AR embeddings of its committed tokens are appended to the causal KV cache for the next block during the next block's first diffusion step, at no extra NFE.

Sampling block $b$ starts from $\z_1^{(\bcolor{b})} \sim \mathcal{N}(\zero, \I)$ and steps from $t$ to $s < t$ along a discretization of $[1,0]$.
At each step the denoiser yields a predicted clean embedding $\hat\e_\theta := \e_\theta(\z_t^{(\bcolor{b})}, t; \c) = \x_\theta(\z_t^{(\bcolor{b})}, t; \c) \Emb$ and the implied noise prediction $\hat\beps_\theta := \frac{1}{\sigma_t}(\z_t^{(\bcolor{b})} - \alpha_t \hat\e_\theta)$.
We introduce two canonical samplers for the reverse process.
The \textit{DDPM} sampler uses the ancestral kernel of VDM~\citep{kingma2021variational}. It draws $\z_s^{(\bcolor{b})}$ from the closed-form Gaussian posterior $q(\z_s^{(\bcolor{b})} \mid \z_t^{(\bcolor{b})}, \x^{(\bcolor{b})})$ with the clean embedding replaced by $\hat\e_\theta$.
The \textit{DDIM} sampler~\citep{song2021denoising} leverages the predictions $\hat\e_\theta$ and $\hat\beps_\theta$ to deterministically update the state as $\z_s^{(\bcolor{b})} = \alpha_s \hat\e_\theta + \sigma_s \hat\beps_\theta$.
For both samplers, after the final step, tokens are read out as $\argmax$ of the conditional logit output $\x_\theta(\z_0^{(\bcolor{b})}, 0; \c)$ and appended to $\c$, so $\text{NFEs}=\text{steps}+1$.
See \cref{alg:sample} for the complete sampling algorithms and \cref{sec:app-samplers} for more details. 

\subsection{Trajectory Steering}
\label{sec:inference-cfg-st}

To sharpen model predictions, we incorporate two reverse-process steering methods to our blockwise continuous diffusion model with no impact to its training efficiency and minimal impact to its inference efficiency.

\begin{figure*}[t]
    \centering
    \begin{subfigure}[t]{0.48\linewidth}
        \centering
        \includegraphics[width=0.9\linewidth]{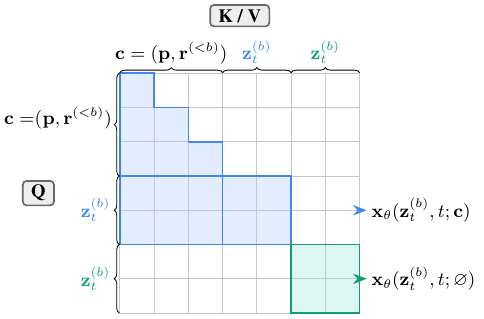}
        \caption{ Packed attention mask while generating block $b$ ($L'=2$). Prefix $\c$ is causal (KV-cached). Conditional $\z_t^{(b)}$ attends to all of $\c$ and bidirectionally within the block; unconditional copy attends only to itself.}
        \label{fig:cfg-attention}
    \end{subfigure}\hfill
    \begin{subfigure}[t]{0.50\linewidth}
        \centering
        \includegraphics[width=0.9\linewidth]{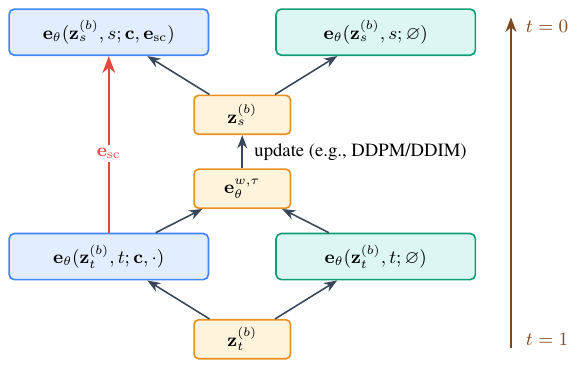}
        \caption{ One reverse step $t\to s$. CFG and ST mix the two denoisers into $\e^{w,\tau}_\theta$; an update yields $\z_s^{(\bcolor{b})}$. SC (red) feeds the previous conditional mean (not sharpened by CFG or ST) to the next conditional branch.}
        \label{fig:cfg-steering}
    \end{subfigure}
    \caption{\textbf{Inference} for blockwise continuous diffusion: packed CFG attention (a) and one reverse step (b).}
    \label{fig:inference-steering}
    \vspace{-0.5em}
\end{figure*}

\paragraph{Classifier-free guidance (CFG).} CFG~\citep{ho2022classifierfreediffusionguidance} is the standard tool for class-conditional or text-to-image diffusion models. In CFG, one queries the model twice to obtain a conditional and an unconditional prediction and then linearly extrapolates them.
For our block diffusion model, the clean context $\c$ serves as the natural conditioning information, and CFG extrapolates away from the unconditional prediction for the current block $\bcolor{b}$:
\begin{align}\label{eq:cfg}
    \e^{w}_\theta(\z_t^{(\bcolor{b})}, t; \c) := \e_\theta(\z_t^{(\bcolor{b})}, t; \varnothing)
    + w\,\big[\e_\theta(\z_t^{(\bcolor{b})}, t; \c) - \e_\theta(\z_t^{(\bcolor{b})}, t; \varnothing)\big],
    \tag{CFG}
\end{align}
where $\varnothing$ drops the entire clean context $\c$ (so $\z_t^{(\bcolor{b})}$ attends only within its own block) and $w \geq 1$ is the guidance strength ($w=1$ falls back to the plain conditional prediction).
We bring two technical considerations. First, at batch size $1$ transformer inference is memory-bound, so we pack conditional and unconditional predictions into a single forward pass at almost no wall-clock cost~\citep{liu2025tidar}, using the packed mask of~\cref{fig:cfg-attention}. We measure this overhead to be less than $5\%$ per forward pass at block size $32$ and $64$ (\cref{app:inference-cfg-fwd-cost}).
Second, unlike in image diffusion models and existing dLMs that require train-time context dropout \citep{potaptchik2026discrete, hu2026elf}, \textbf{our CFG is a pure inference-time trick}: due to sequence packing in pretraining, the first block typically starts in the middle of a document and naturally sees no context.
We discuss existing approaches in \cref{app:related-work}.

\paragraph{Score temperature (ST).} ST~\citep{dieleman2022continuous} is a heuristic that sharpens the conditional score itself, rescaling the score $\nabla\log p_t(\z_t^{(\bcolor{b})}\mid\c)$ (or equivalently, the noise prediction) by $\frac{1}{\tau}$ for $\tau\in(0,1]$. One can show that it corresponds to the following operation on the mean prediction:
\begin{align}\label{eq:st}
    \e^{\tau}_\theta(\z_t^{(\bcolor{b})}, t; \c) := 
    \tfrac{\z_t^{(\bcolor{b})}}{\alpha_t} + \tfrac{1}{\tau}\big[\e_\theta(\z_t^{(\bcolor{b})}, t; \c) - \tfrac{\z_t^{(\bcolor{b})}}{\alpha_t}\big].
    \tag{ST}
\end{align}

\begin{wrapfigure}{r}{0.4\linewidth}
    \vspace{-1em}
    \centering
    \includegraphics[width=0.9\linewidth]{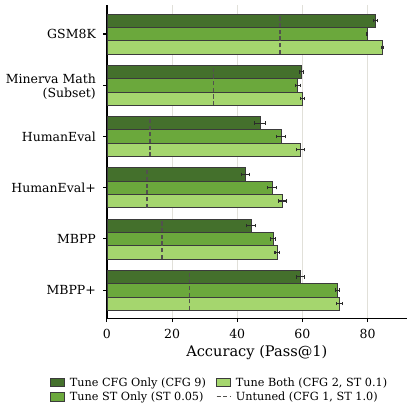}
    \caption{ \textbf{CFG and ST tuning} on \model\unreg-3B for DDPM sampling. Combining both techniques yields larger gains over applying either in isolation, and markedly outperforms sampling without CFG or ST. See \cref{fig:ddpm-256-sweep} for the full tuning grid.}
    \label{fig:cfg-st-tuning}
    \vspace{-2em}
\end{wrapfigure}

Intuitively, since $\E[\z_t^{(\bcolor{b})}|\x^{(\bcolor{b})}] = \alpha_t \e^{(\bcolor{b})}$, $\frac{1}{\alpha_t}\z_t^{(\bcolor{b})}$ serves as the naive bad prediction to extrapolate from.
ST does not incur further model queries, and $\tau = 1$ leaves the sampler unchanged. We also recognize ST as a special case of Temporal Score Rescaling proposed in~\citet{xu2025temporal}.

Finally, we remark that due to the commutativity of \cref{eq:cfg,eq:st}, we can combine them into a single operation $\e^{w, \tau}_\theta(\z_t^{(\bcolor{b})}, t; \c)$, which is essentially using the score of the composed distribution $(p_t(\cdot \mid \c)^w p_t(\cdot \mid \varnothing)^{1-w})^{1/\tau}$. We observed that using either knob is weaker than using both of them, see \cref{fig:cfg-st-tuning}.

\paragraph{Self-conditioning (SC).} At inference, SC is applied only to the conditional prediction: when querying the model at time $s$, the previous step's conditional prediction $\e_\theta(\z_t^{(\bcolor{b})}, t; \c)$ (before applying CFG and ST) serves as the SC input for the conditional prediction, while the unconditional prediction receives $\zero$ as the SC input (\cref{fig:cfg-steering}). 

\paragraph{Sharpening in logit space.} 
While in the above, we apply guidance directly in the embedding domain, we can also sharpen in logit space.
In \cref{app:inference-logit-space}, we show that the logit-space CFG and temperature do provide sharpening, but the embedding-space knobs provide stronger performance, hence we keep \cref{eq:cfg,eq:st} as the default option.

\subsection{Embedding Regularization}  
\label{sec:few-step-susreg}
We observe that during training, embeddings collapse to approximately 5 effective dimensions (i.e., ${>}90\%$ explained variance in the first 5 principal components), a behavior typically not observed in AR language model embeddings but consistent with observations in RePlaid \citep{yang2026continuous}.
This may arise because constant embeddings are a trivial solution for the denoising loss.
We hypothesize that this embedding collapse exacerbates token ambiguity under prediction errors from coarse numerical precision and low-NFE sampling, which motivates regularizing the embeddings toward a uniform distribution on the unit sphere.
For our regularized model \model, we therefore apply a multi-scale extension of \textit{Sketched Uniform Spherical Regularization} (SUSReg, \citet{nicollier2026sphere}) to the embedding matrix $\Emb$ during pretraining, penalizing deviations from the uniform distribution on the unit sphere.
We also add a local repulsion loss to push apart very close embedding vectors, and make other minor training changes, see \cref{app:pretrain-susreg} for full details.
We ablate a simpler variant as in the original SUSReg \citep{nicollier2026sphere}, denoted \model\preg, which also avoids embedding collapse but is less effective at improving low-NFE quality and distillation (\cref{tab:acc-nfe-3b,tab:acc-nfe-8b}); we 
report its full benchmark results in \cref{tab:base-variants}.%

\begin{wrapfigure}{r}{0.6\linewidth}
    \vspace{-3em}
    \centering
    \includegraphics[width=\linewidth]{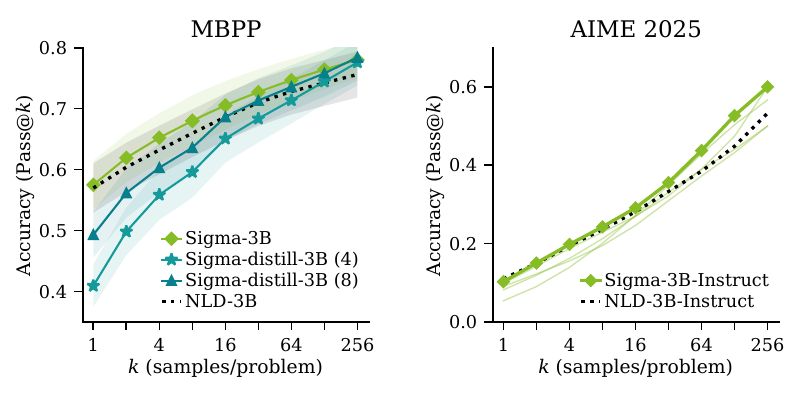}
    \vspace{-2em}
    \caption{
    \textbf{Left: Pass@$k$ accuracy on MBPP for \model-3B and \model\distill-3B}. All models use pass@1 selected sampler. \textbf{Right: Pass@$k$ accuracy on AIME 2025 for \model-3B-Instruct}, highlighting best CFG/ST pair chosen from five pre-defined combinations.
     }
    \label{fig:passk-mbpp-main}
    \vspace{-1em}
\end{wrapfigure}

\paragraph{Results.} As shown in \cref{tab:base-3b,tab:base-8b}, the resulting regularized models \model-3B/8B achieve the highest average reasoning scores, outperforming NLD \citep{fu2026nemotron}. Furthermore, they significantly improve few-step sampling (see \cref{tab:acc-nfe-3b}) and distillation (see \cref{app:dmd-additional-results}) compared to \model\unreg-3B/8B, supporting our hypotheses.
Despite exhibiting substantially different embedding geometries from the unregularized models (\cref{fig:mollwide}), interestingly we find that regularization incurs almost no validation PPL penalty.

\subsection{Mathematical Reasoning and Coding Performance}\label{subsec:3b-base-math-code}

\begin{table}[t]
    \centering
    \caption{
    \textbf{Pass@$1$ accuracy of \model\unreg-3B and \model-3B.}
    The \model diffusion rows are reported as mean (over 10 seeds) $\pm$ two standard errors. 
    \textbf{Bold}: the best value across all methods; \underline{underline}: the best value within each mode.
    }
    \label{tab:base-3b}
    \resizebox{\linewidth}{!}{%
    \begin{tabular}{lccccccc}
    \toprule
    & \multicolumn{2}{c}{\textbf{Math}} & \multicolumn{4}{c}{\textbf{Coding}} & \\
    \cmidrule(lr){2-3} \cmidrule(lr){4-7}
    Method & GSM8K & Minerva Math & HumanEval & HumanEval+ & MBPP & MBPP+ & Avg. \\
    \midrule
    AR initialization & 83.70 & \textbf{66.28} & 51.22 & 48.78 & 60.60 & 77.78 & 64.73 \\
    \midrule
    \multicolumn{8}{l}{\emph{AR mode}} \\
    \quad Nemotron-Labs-Diffusion-3B & \underline{83.78} & 65.06 & 53.05 & 47.56 & \textbf{\underline{60.80}} & \textbf{\underline{78.84}} & 64.85 \\
    \quad \textbf{\model\unreg-3B (Ours)} & 83.40 & \underline{65.10} & \underline{57.32} & \underline{51.83} & 59.00 & 76.19 & \underline{65.47} \\ %
    \quad \textbf{\model-3B (Ours)} & 81.20 & 63.98 & 51.83 & 48.17 & 56.80 & 75.66 & 62.94 \\ %
    \midrule
    \multicolumn{8}{l}{\emph{Diffusion mode}} \\
    \quad LLaDA-8B                                 & 70.96 & 27.30 & 32.32 & 27.44 & 40.80 & 51.85 & 41.78 \\
    \quad Dream-7B                                 & 77.18 & 39.60 & 54.88 & 49.39 & 56.80 & \underline{74.60} & 58.74 \\
    \quad Nemotron-Labs-Diffusion-3B & 83.40 & 63.96 & 56.10 & 53.05 & 57.60 & 74.34 & 64.74 \\
    \quad \textbf{\model\unreg-3B (Ours)} & & & & & & & \\ %
    \qquad DDPM $256$ steps/block ($w{=}2,\tau{=}0.1$) & \se{84.50}{0.31} & \se{64.60}{0.28} & \se{\textbf{\underline{59.45}}}{1.30} & \se{\textbf{\underline{53.84}}}{1.29} & \se{52.32}{0.73} & \se{71.32}{0.91} & 64.34 \\ %
    \qquad DDIM $32$ steps/block ($w{=}3,\tau{=}0.3$) & \se{81.91}{0.32} & \se{62.76}{0.18} & \se{56.59}{1.31} & \se{50.67}{1.00} & \se{47.26}{0.46} & \se{63.49}{0.77} & 60.45 \\ %
    \quad \textbf{\model-3B (Ours)} & & & & & & & \\ %
    \qquad DDPM $256$ steps/block ($w{=}3,\tau{=}0.1$) & \se{\B{\U{85.57}}}{0.29} & \se{\U{64.98}}{0.19} & \se{57.26}{0.82} & \se{\B{\U{53.84}}}{0.55} & \se{\U{58.00}}{0.43} & \se{73.62}{0.51} & \B{\U{65.55}} \\ %
    \qquad DDIM $32$ steps/block ($w{=}3,\tau{=}0.3$) & \se{84.48}{0.21} & \se{64.52}{0.21} & \se{55.37}{1.28} & \se{51.71}{1.23} & \se{53.00}{0.34} & \se{71.30}{0.45} & 63.40 \\ %
    \bottomrule
    \end{tabular}}
\end{table}

\begin{table}[h]
    \centering
    \caption{
     \textbf{Pass@$1$ accuracy of \model\unreg-8B and \model-8B.}
    The \model diffusion rows are reported as mean (over 10 seeds) $\pm$ two standard errors. 
    \textbf{Bold}: the best value across all methods; \underline{underline}: the best value within each mode.
    }
    \label{tab:base-8b}
    \resizebox{\linewidth}{!}{%
    \begin{tabular}{lccccccc}
    \toprule
    & \multicolumn{2}{c}{\textbf{Math}} & \multicolumn{4}{c}{\textbf{Coding}} & \\
    \cmidrule(lr){2-3} \cmidrule(lr){4-7}
    Method & GSM8K & Minerva Math & HumanEval & HumanEval+ & MBPP & MBPP+ & Avg. \\
    \midrule
    AR initialization                & 85.90 & \textbf{69.18} & 59.15 & 53.05 & 68.20 & \textbf{83.33} & 69.80 \\
    \midrule
    \multicolumn{8}{l}{\emph{AR mode}} \\
    \quad Llama-3.1-8B                             & 54.06 & 18.22 & 35.37 & 28.66 & 48.80 & 61.90 & 41.17 \\
    \quad Ministral3-8B                            & 80.21 & 44.58 & 42.68 & 38.41 & 61.60 & 76.98 & 57.41 \\
    \quad Qwen3-8B                                 & 86.73 & 52.94 & \underline{64.63} & \underline{56.71} & \textbf{\underline{69.40}} & \underline{83.07} & 68.91 \\
    \quad Nemotron-Labs-Diffusion-8B               & \underline{88.25} & 66.00 & 60.37 & 53.05 & 68.20 & 82.54 & \underline{69.73} \\
    \quad \textbf{\model\unreg-8B (Ours)}     & 86.88 & \underline{66.72} & 51.83 & 48.78 & 64.00 & 80.95 & 66.53 \\ %
    \quad \textbf{\model-8B (Ours)}           & 84.84 & 65.18 & 56.10 & 52.44 & 63.80 & 82.54 & 67.48 \\ %
    \midrule
    \multicolumn{8}{l}{\emph{Diffusion mode}} \\
    \quad LLaDA-8B                                 & 70.96 & 27.30 & 32.32 & 27.44 & 40.80 & 51.85 & 41.78 \\
    \quad Dream-7B                                 & 77.18 & 39.60 & 54.88 & 49.39 & 56.80 & 74.60 & 58.74 \\
    \quad Nemotron-Labs-Diffusion-8B               & 87.26 & 65.16 & \underline{62.80} & 57.32 & \underline{67.00} & \underline{81.75} & 70.22 \\
    \quad \textbf{\model\unreg-8B (Ours)}     &       &       &       &       &       &       &       \\ %
    \qquad DDPM 256 steps/block ($w{=}2,\tau{=}0.1$) & \se{\textbf{\underline{89.07}}}{0.28} & \se{\underline{67.47}}{0.20} & \se{61.40}{1.73} & \se{57.44}{1.62} & \se{64.78}{0.54} & \se{77.83}{0.72} & 69.67 \\
    \qquad DDIM 32 steps/block ($w{=}3,\tau{=}0.3$) & \se{87.78}{0.33} & \se{67.13}{0.15} & \se{58.54}{1.04} & \se{53.17}{1.27} & \se{59.54}{0.57} & \se{74.15}{0.43} & 66.72 \\
    \quad \textbf{\model-8B (Ours)}     &       &       &       &       &       &       &       \\ %
    \qquad DDPM 256 steps/block ($w{=}3,\tau{=}0.1$) & \se{88.57}{0.12} & \se{66.72}{0.16} & \se{65.12}{1.22} & \se{60.61}{1.15} & \se{65.74}{0.44} & \se{80.19}{0.76} & \B{\U{71.16}} \\
    \qquad DDIM 32 steps/block ($w{=}3,\tau{=}0.3$) & \se{88.48}{0.14} & \se{65.91}{0.15} & \se{\B{\U{65.37}}}{0.57} & \se{\B{\U{60.91}}}{0.74} & \se{62.68}{0.30} & \se{77.72}{0.32} & 70.18 \\
    \bottomrule
    \end{tabular}%
    }
\end{table}

\label{sec:inference-result}

We evaluate on two math and four coding benchmarks: GSM8K, Minerva Math, HumanEval, HumanEval+, MBPP, and MBPP+. Results are reported in~\cref{tab:base-3b}. 
For diffusion generation, \model models always use SC, and we report performance on DDPM sampling with 256 steps/block with EDM $\rho=7$ time discretization, and DDIM sampling with 32 steps/block with uniform-$\gamma$ time discretization.
See~\cref{app:inference-eval-setup} for details on the harness, scoring pipeline, error bar conventions, and the samplers for baselines, and \cref{app:inference-time-disc} for time discretization schemes.

\paragraph{Results.}
In AR mode, \model\unreg-3B attains the best six-task average across all methods; in diffusion mode, our DDPM-256 runs are comparable and out-perform NLD-3B on average and sets the best base-model numbers on GSM8K, HumanEval, and HumanEval+; they also outperform LLaDA-8B and Dream-7B by a noticeable margin, despite being $\sim\!2\times$ smaller in parameters. Empirically, we observe \textit{less} left-to-right bias within a block for \model compared with the confidence-based masked dLM sampling (\cref{fig:commitment-order}). To confirm the benefit of AR regularization (\cref{sec:pretrain-from-scratch}), we also trained \model\unreg-3B with $\cL_{\text{BD}}$ alone \cref{eq:block-nelbo} without $\cL_{\text{ar}}$ \cref{eq:ar_loss}; results are uniformly worse across all tasks (\cref{fig:ddpm-256-sweep-no-ar-loss}). We refer readers to \cref{tab:base-8b} for the same evaluation on 8B base models.

\paragraph{Pass@$k$ analysis.}
While pass@$1$ accuracy in \cref{tab:base-3b} captures single-draw quality, test-time search and RL draw multiple rollouts for the same prompt, where diverse rollouts are needed for richer learning signals and increase the chance of discovering a correct solution. Pass@$k$ does not directly measure diversity, but tests whether additional samples expand correct-solution coverage. 
We report results in \cref{fig:passk-mbpp-main} on MBPP at the 3B scale, using the same sampler configuration that performs best on pass@$1$ accuracy with 256 NFEs DDIM sampler. 
There, \model-3B increasingly surpasses as more trials are available, showing that the model has a useful output diversity. 
See \cref{app:inference-passk} for the full study.

\section{Few-Step Generation}
\label{sec:few-step}

\begin{figure*}[t]
    \centering
    \includegraphics[width=0.9\linewidth]{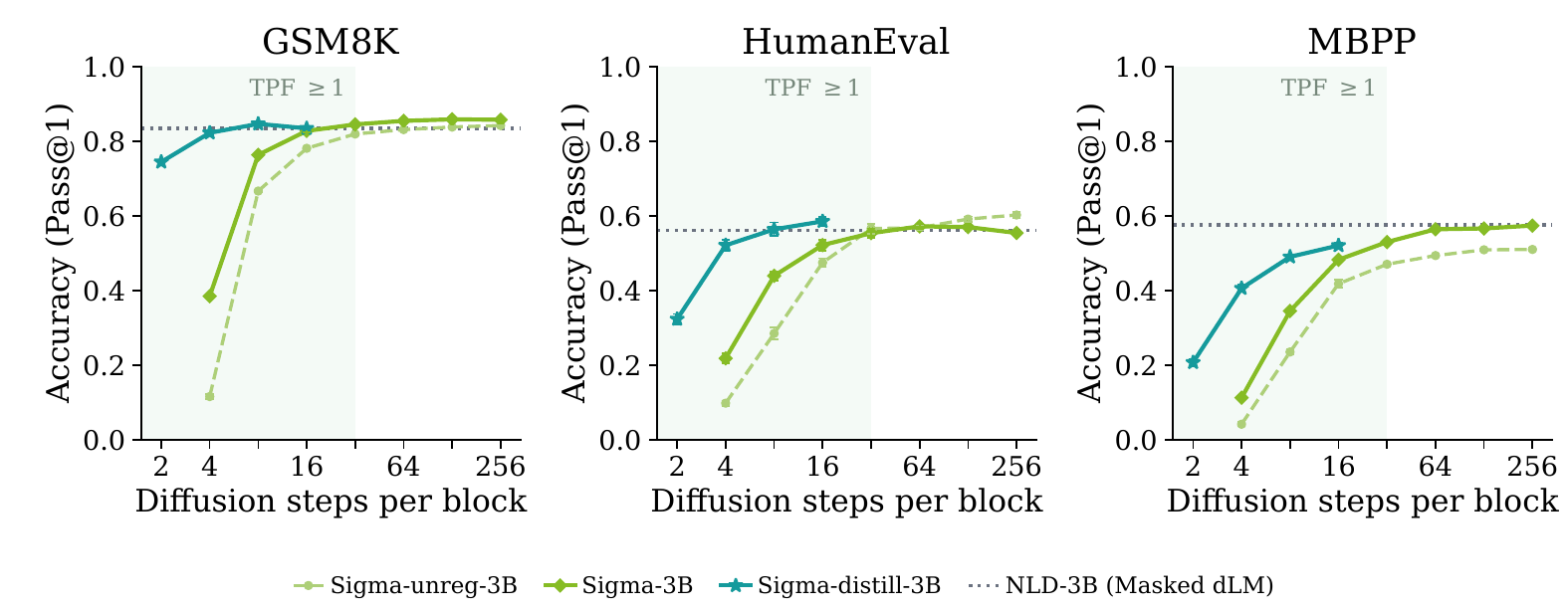}
    \caption{ \textbf{Pass@$1$ accuracy of \model\unreg-3B, \model-3B, and \model\distill-3B}
    Undistilled models use DDIM with best per-NFE (CFG, ST), shared across the three tasks. 
    Full results on base tasks in \cref{tab:acc-nfe-3b}.
    }
    \label{fig:3b-nfe}
    \vspace{-1em}
\end{figure*}

\begin{figure*}[t]
    \centering
    \includegraphics[width=0.9\linewidth]{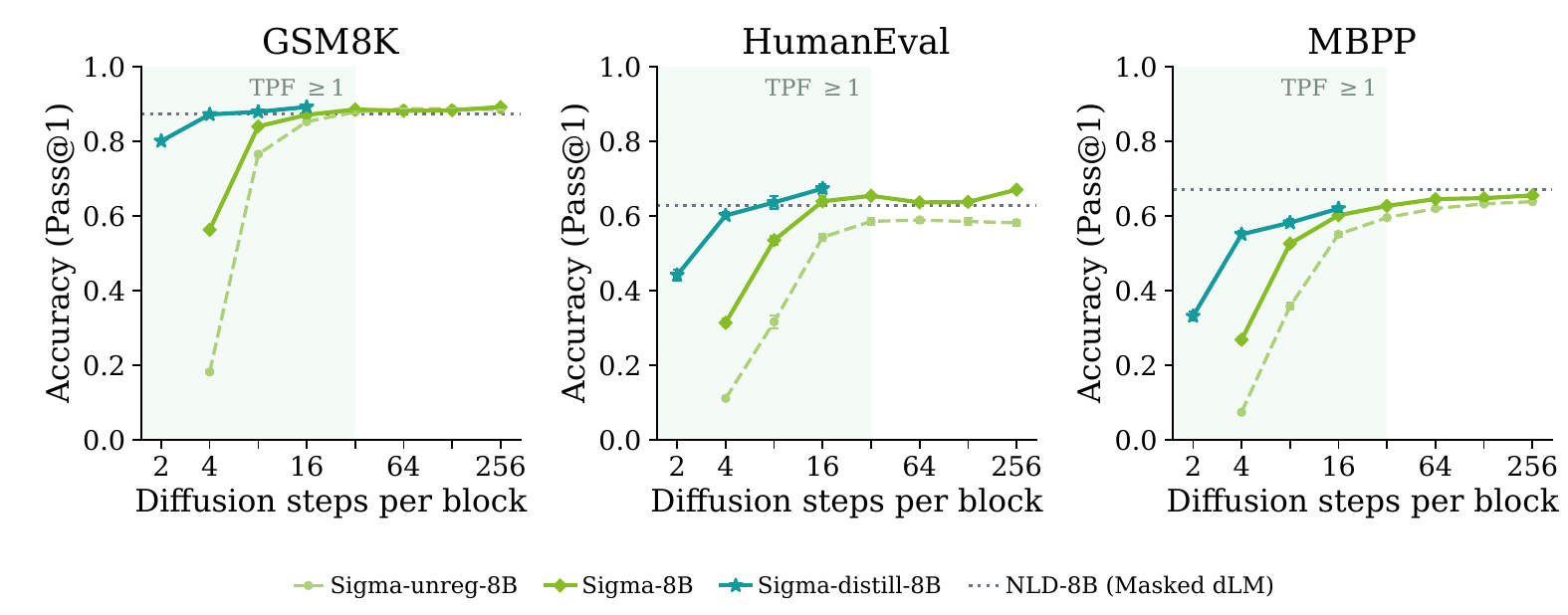}
    \caption{
    \textbf{Pass@$1$ accuracy of \model\unreg-8B, \model-8B and \model\distill-8B.}
    Undistilled models use DDIM with best per-NFE (CFG, ST), shared across the three tasks. 
    Full results on base tasks in \cref{tab:acc-nfe-8b}.
    }
    \label{fig:8b-nfe}
    \vspace{-1em}
\end{figure*}

\subsection{ODE Sampler and Time Discretization}
\label{sec:few-step-ddim}

In previous results, we mainly run 256 DDPM steps per block to establish the strongest performance. In low-NFE regime, it is generally known that deterministic samplers paired with well-tuned time discretization lead to better performance. To investigate the tradeoffs for \model, we choose three representative tasks (GSM8K, HumanEval, and MBPP), and vary the diffusion sampling steps from $4$ to $256$. For each sampler-discretization-budget combination, we sweep CFG and ST; we then report, for each combination, the CFG-ST pair that achieves the best average accuracy over the two tasks; see \cref{app:inference-eval-setup} for more details.

\paragraph{Results.} As shown in \cref{fig:few-step-sweep}, DDIM substantially outperforms DDPM at low NFEs while matching DDPM at high NFEs. The uniform-$\gamma$ discretization (\cref{app:inference-time-disc}) performs the best for a wide range of budgets. 
Using the same optimal hyperparameters from the two tasks, we evaluate the 16/32-step performance across all six tasks in~\cref{tab:base-3b,tab:base-8b}; \cref{tab:acc-nfe-3b,tab:acc-nfe-8b} show performance at other budgets.

\subsection{Distribution Matching Distillation for Few-Step Generation}
\label{sec:dmd}

Although DDIM has better performance than DDPM at a reduced generation budget, the pretrained models still lose quality at very low NFEs (especially ${\le} 16$ steps, see \cref{fig:3b-nfe}). We therefore distill \model into a few-step student \model\distill{} using Distribution-Matching Distillation (DMD, \citet{yin2024one,yin2024improved}). Notably, the student uses self-conditioning and a purely stochastic sampler that alternates between clean embedding estimation and fresh Gaussian noise injection, which is different from DDIM sampler. Furthermore, our DMD students are trained to match the target distribution without using CFG, which saves inference FLOPs per NFE compared with the guided teacher, and needs no ST rescaling.

\paragraph{Results.} As shown in~\cref{fig:3b-nfe,fig:8b-nfe,tab:acc-nfe-3b,tab:acc-nfe-8b}, our distillation recipe yields significant acceleration: with only 8 NFEs ($\text{TPF}=4$), \model\distill{} matches both NLD \citep{fu2026nemotron} and our best-performing 256-NFE DDPM sampler on GSM8K and HumanEval at both 3B and 8B scales. For MBPP, which is a full code generation task, syntactic bugs are introduced at low NFE, but \model\distill-3B is still within 3\% accuracy of the 256-step sampler at 8 NFEs. Interestingly, despite significant acceleration, we observe that \model\distill-3B does not lose any diversity compared with its 256-step DDPM teacher (see \cref{app:dmd-additional-results}). 
As a result, the distilled model exhibits strong pass@$k$ performance: both 4- and 8-NFE \model\distill-3B catch up with the 256-NFE DDIM \model-3B teacher at $k=256$ (see \cref{fig:passk-mbpp-main}). 
Together, this suggests the distilled model is well suited for accelerated RL-based alignment and post-training.

\section{Self-Speculative Decoding}
\label{sec:specdec}
\definecolor{specNLD}{HTML}{5B6670}       %
\definecolor{specDrafter}{HTML}{9932CC}   %
\definecolor{specDistilled}{HTML}{1874CD} %
\definecolor{specAccept}{HTML}{B5102F}    %
\providecommand{\tbd}{\textcolor{red}{TBD}}
\model trains a block-diffusion stream and a causal AR stream on one backbone, so it can act as its own drafter.
In each round, one denoiser call from pure noise drafts a block of $L'=32$ tokens.
The AR stream then scores the block in one causal pass, keeps the longest prefix that matches its own greedy tokens, and appends one token of its own~\citep{cdlmapp_leviathan2023,fu2026nemotron}.
The output is therefore the greedy AR output.
A round of two forward passes commits $\mathrm{AL}+1$ tokens, where the acceptance length (AL) counts the accepted draft tokens that reach the returned answer.

At both 3B and 8B scales, we compare NLD to \model\distill{} (\cref{sec:dmd}) to a post-trained model \model\distill-\textit{spec}, which pairs the same distillation objective as \model\distill{} with an acceptance objective (full details in \cref{app:specdec-lk}).
We serve with SGLang~\citep{cdlmapp_zheng2024sglang} in BF16 on one NVIDIA GB200, one request at a time, and evaluate on GSM8K, HumanEval, and MBPP.
AL is pooled over the prompts whose outputs neither loop nor reach the length cap.
\Cref{app:cdlm3b-specdec} details the protocol and gives further results for the undistilled \model with trained drafters.

\begin{figure}[t]
    \centering
    \includegraphics[width=0.85\linewidth]{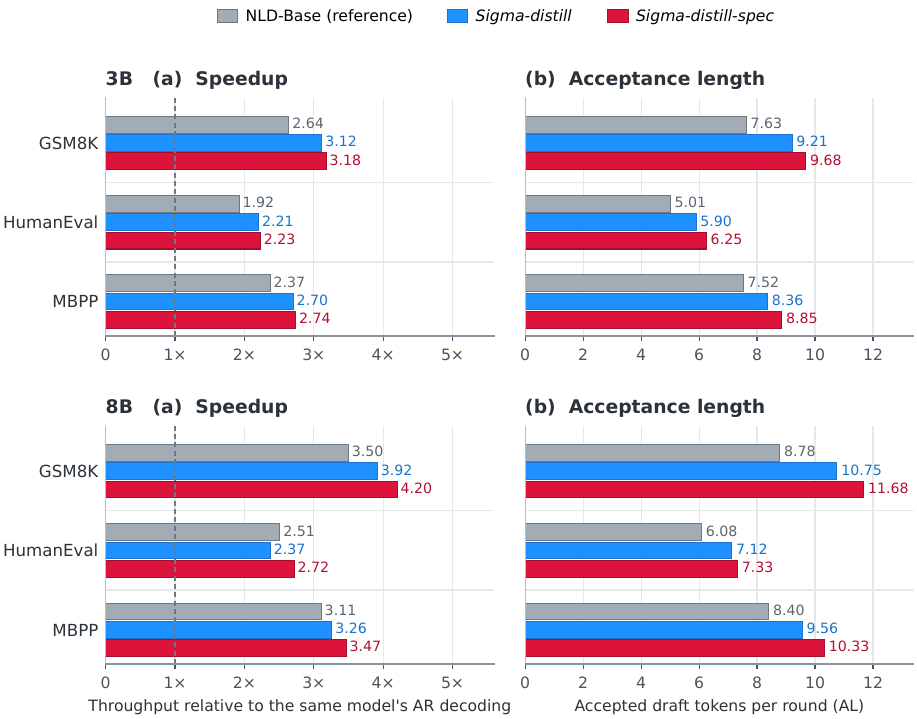}
    \caption{
    \textbf{Self-speculative decoding at 3B (top) and 8B (bottom).}
    (a) Throughput relative to the same model's AR decoding.
    (b) Accepted draft tokens per round (AL).
    }
    \label{fig:specdec-main}
\end{figure}

\begin{table}[t]
    \centering
    \caption{
    \textbf{Accuracy and AL gain over NLD.}
    Accuracy (\%) is that of AR decoding, which speculative decoding reproduces (GSM8K: flexible extract; code: pass@$1$).
    AL gain is relative to NLD of the same size, over the prompts both models keep, with 95\% bootstrap intervals.
    }
    \label{tab:specdec-main}
    \footnotesize
    \setlength{\tabcolsep}{3.5pt}
    \begin{tabular}{@{}llcccccc@{}}
    \toprule
    & & \multicolumn{2}{c}{\textbf{GSM8K}} & \multicolumn{2}{c}{\textbf{HumanEval}} & \multicolumn{2}{c}{\textbf{MBPP}} \\
    \cmidrule(lr){3-4} \cmidrule(lr){5-6} \cmidrule(lr){7-8}
    & Model & Acc. & AL gain & Acc. & AL gain & Acc. & AL gain \\
    \midrule
    \multirow{3}{*}{3B}
    & \textcolor{specNLD}{NLD-3B-Base} & 83.8 & -- & 52.4 & -- & 61.6 & -- \\
    & \textcolor{specDistilled}{\model\distill-3B} & 82.9 & +20.8\% {\scriptsize[18.9, 22.5]} & 56.1 & +18.5\% {\scriptsize[7.4, 30.1]} & 58.2 & +10.5\% {\scriptsize[7.1, 13.9]} \\
    & \textcolor{specAccept}{\model\distill-\textit{spec}-3B} & 82.2 & +26.7\% {\scriptsize[24.8, 28.7]} & 55.5 & +26.3\% {\scriptsize[15.6, 36.1]} & 58.8 & +17.4\% {\scriptsize[14.1, 20.7]} \\
    \midrule
    \multirow{3}{*}{8B}
    & \textcolor{specNLD}{NLD-8B-Base} & 88.8 & -- & 61.6 & -- & 66.2 & -- \\
    & \textcolor{specDistilled}{\model\distill-8B} & 86.4 & +22.6\% {\scriptsize[20.6, 24.6]} & 56.1 & +15.0\% {\scriptsize[4.1, 27.6]} & 64.0 & +13.6\% {\scriptsize[10.3, 17.3]} \\
    & \textcolor{specAccept}{\model\distill-\textit{spec}-8B} & 86.4 & +33.0\% {\scriptsize[31.0, 35.1]} & 57.3 & +21.0\% {\scriptsize[9.8, 34.9]} & 63.8 & +22.7\% {\scriptsize[19.1, 26.5]} \\
    \bottomrule
    \end{tabular}
\end{table}

\paragraph{Results.}
\Cref{fig:specdec-main} shows speedup and AL, and \cref{tab:specdec-main} shows accuracy and the AL gain over NLD.
Without any drafter training, \model\distill-3B accepts 11--21\% more draft tokens per round than NLD-3B, and \model\distill-8B accepts 14--23\% more than NLD-8B.
We also show in \cref{app:specdec-undistilled} that \model\distill-3B also accepts 15--22\% more tokens than the \model-3B, more than what an adapter trained specifically for improving drafting adds to that model. \model\distill-8B gains also more than the adapter on GSM8K and MBPP and as much on HumanEval (\cref{app:specdec-undistilled}).
Further acceptance training (joint distillation and drafter fine-tuning) yields \model\distill-\textit{spec}, which  accepts another 5--9\% more tokens per round and decodes up to 3.18$\times$ (3B) and 4.20$\times$ (8B) faster than AR decoding, ahead of NLD on every benchmark and model size.
In all these experiments, the accuracy remains that of the AR mode in each model, since verification returns the greedy AR output. Therefore the speedup comes at no cost relative to AR decoding with the same model.

\section{Supervised Fine-Tuning}
\label{sec:sft}
We apply supervised fine-tuning (SFT) to the pretrained \model models with the joint AR and block-diffusion losses, restricting supervision to assistant responses. The complete data processing, optimization recipe, and continuous-diffusion-specific controls are given in~\cref{app:sft}.

\paragraph{Results.} As shown in \cref{tab:sft-3b,tab:sft-8b}, the average performance of \model\instruct models is on par with NLD at both 3B and 8B scale, and surpasses other open-source masked dLMs like Dream and LLaDA at the 8B scale. For both 3B checkpoints, we partially re-sweep CFG and ST after SFT and observe that the same optima selected on the corresponding pre-SFT models remain statistically tied for best after SFT (\cref{tab:sft-cfg-st-sweep}); we therefore retain the pre-SFT settings for the full evaluation.

\begin{table}[h]
    \centering
    \caption{
    \textbf{Pass@$1$ accuracy of \model\unreg-3B\instruct and \model-3B\instruct.}
    The \model diffusion rows are reported as mean (over 10 seeds) $\pm$ two standard errors. 
    The four reference models at the top are AR models.
    Baseline numbers are from \citet{fu2026nemotron}.
    In \emph{diffusion mode}, \underline{underline} marks best value per task.
    }
     \label{tab:sft-3b}
    \resizebox{\linewidth}{!}{%
    \begin{tabular}{lccccccccccc}
    \toprule
    & \multicolumn{4}{c}{\textbf{Math}} & \multicolumn{3}{c}{\textbf{Coding}} & \multicolumn{2}{c}{\textbf{Science \& Knowledge}} & \textbf{IF} & \\
    \cmidrule(lr){2-5} \cmidrule(lr){6-8} \cmidrule(lr){9-10} \cmidrule(lr){11-11}
    Method & GSM8K & Math500 & AIME24 & AIME25 & HumanEval & MBPP & LCB-CPP & \makebox[4.2em]{MMLU} & \makebox[4.2em]{GPQA} & IFEval & Avg. \\
    \midrule
    Llama-3.2-3B-Instruct & 62.09 & 32.60 & 0.00 & 0.00 & 65.85 & 66.93 & 9.25 & 62.45 & 27.78 & 69.69 & 39.66 \\
    Phi-4-mini-Instruct & 76.57 & 70.80 & 10.00 & 6.67 & 77.44 & 75.13 & 9.91 & 53.37 & 40.91 & 71.16 & 49.20 \\
    Ministral3-3B-Instruct-2512 & 87.41 & 71.90 & 12.50 & 12.71 & 64.79 & 68.39 & 12.50 & 63.21 & 28.79 & 57.55 & 47.98 \\
    Qwen3-4B & 92.19 & 68.40 & 10.63 & 10.63 & 75.91 & 72.49 & 15.58 & 77.05 & 37.18 & 72.20 & 53.23 \\
    \midrule
    \multicolumn{12}{l}{\emph{AR mode}} \\
    \quad Nemotron-Labs-Diffusion-3B-Instruct & 87.87 & 77.60 & 23.33 & 16.67 & 76.22 & 71.16 & 21.37 & 71.70 & 39.39 & 69.39 & 55.47 \\
    \quad \textbf{\model\unreg-3B\instruct (Ours)} & 88.17 & 71.60 & 20.00 & 10.00 & 71.34 & 77.51 & 13.00 & 71.36 & 38.38 & 73.62 & 53.50 \\ %
    \quad \textbf{\model-3B\instruct (Ours)} & 85.29 & 71.60 & 3.33 & 3.33 & 73.17 & 78.04 & 14.76 & 69.98 & 34.34 & 70.74 & 50.46 \\ %
    \midrule
    \multicolumn{12}{l}{\emph{Diffusion mode}} \\
    \quad Nemotron-Labs-Diffusion-3B-Instruct & 88.40 & \underline{74.80} & \underline{16.67} & 10.00 & 74.39 & 73.54 & \underline{16.30} & 72.06 & 33.84 & 68.93 &  52.89 \\
    \quad \textbf{\model\unreg-3B\instruct (Ours)} & & & & & & & & & & & \\
    \qquad DDPM (256 steps, $w{=}2,\tau{=}0.1$) & \sev{89.32}{0.24} & \sev{73.20}{0.45} & \sev{15.67}{2.84} & \sev{11.00}{1.90} & \sev{\underline{77.32}}{1.05} & \sev{71.75}{0.72} & \sev{12.97}{0.62} & \sev{\underline{72.58}}{0.08} & \sev{\underline{37.68}}{1.70} & \sev{70.11}{0.64} &  53.16 \\
    \qquad DDIM (32 steps, $w{=}3,\tau{=}0.3$) & \sev{88.10}{0.24} & \sev{71.42}{0.88} & \sev{14.00}{2.46} & \sev{8.67}{1.93} & \sev{73.60}{1.17} & \sev{68.76}{0.69} & \sev{11.56}{0.61} & \sev{71.37}{0.08} & \sev{34.24}{1.42} & \sev{68.18}{0.32} & 50.99 \\
    \quad \textbf{\model-3B\instruct (Ours)} & & & & & & & & & & & \\
    \qquad DDPM (256 steps, $w{=}3,\tau{=}0.1$) & \sev{\underline{89.64}}{0.10} & \sev{73.08}{0.62} & \sev{13.67}{1.99} & \sev{11.33}{1.03} & \sev{77.01}{1.24} & \sev{\underline{74.97}}{0.59} & \sev{13.50}{0.42} & \sev{70.64}{0.12} & \sev{36.01}{1.46} & \sev{\underline{73.26}}{0.59} &  \underline{53.31} \\
    \qquad DDIM (32 steps, $w{=}3,\tau{=}0.3$) & \sev{88.82}{0.15} & \sev{72.10}{0.74} & \sev{15.33}{2.70} & \sev{\underline{13.33}}{2.11} & \sev{73.96}{0.91} & \sev{73.31}{0.82} & \sev{12.16}{0.56} & \sev{70.20}{0.07} & \sev{34.80}{1.99} & \sev{69.00}{0.49} & 52.30 \\
    \bottomrule
    \end{tabular}%
    }
\end{table}

\begin{table}[h]
    \centering
    \caption{
    \textbf{Pass@$1$ accuracy of \model\unreg-8B\instruct and \model-8B\instruct.}
    The \model diffusion rows are reported as mean (over 10 seeds) $\pm$ two standard errors. The three reference models at the top are AR models. 
    NLD-8B-Instruct's diffusion mode results are reproduced under aligned protocol, while other baseline numbers are from~\citet[Tab.~5]{fu2026nemotron}. In \emph{diffusion mode}, 
    \underline{underline} marks best value per task.
    }
    \label{tab:sft-8b}
    \resizebox{\linewidth}{!}{%
    \begin{tabular}{lccccccccccc}
    \toprule
    & \multicolumn{4}{c}{\textbf{Math}} & \multicolumn{3}{c}{\textbf{Coding}} & \multicolumn{2}{c}{\textbf{Science \& Knowledge}} & \textbf{IF} & \\
    \cmidrule(lr){2-5} \cmidrule(lr){6-8} \cmidrule(lr){9-10} \cmidrule(lr){11-11}
    Method & GSM8K & Math500 & AIME24 & AIME25 & HumanEval & MBPP & LCB-CPP & \makebox[4.2em]{MMLU} & \makebox[4.2em]{GPQA} & IFEval & Avg. \\
    \midrule
    Qwen2.5-7B-Instruct                        & 91.89 & 75.10 & 13.75 &  6.88 & 77.44 & 81.55 & 12.33 & 74.86 & 37.12 & 74.58 & 54.55 \\
    Qwen3-8B                                   & 92.42 & 84.80 & 30.21 & 22.08 & 81.71 & 81.88 & 21.09 & 76.66 & 49.24 & 87.38 & 62.75 \\
    Ministral3-8B-Instruct-2512                & 92.42 & 83.60 & 27.71 & 24.58 & 71.04 & 78.97 & 20.76 & 73.90 & 42.87 & 64.31 & 58.02 \\
    \midrule
    \multicolumn{12}{l}{\emph{AR mode}} \\
    \quad Nemotron-Labs-Diffusion-8B-Instruct           & 94.01 & 88.00 & 33.33 & 33.33 & 80.49 & 85.19 & 28.85 & 79.85 & 44.44 & 68.65 & 63.61 \\
    \quad \textbf{\model\unreg-8B\instruct (Ours)} & 91.74 & 80.80 & 16.67 & 20.00 & 78.05 & 84.92 & 21.59 & 78.22 & 46.97 & 79.50 & 59.85 \\
    \quad \textbf{\model-8B\instruct (Ours)} & 91.58 & 79.40 & 26.67 & 16.67 & 81.10 & 80.69 & 12.33 & 76.27 & 43.94 & 78.06 & 58.67 \\
    \midrule
    \multicolumn{12}{l}{\emph{Diffusion mode}} \\
    \quad LLaDA-8B Instruct                    & 79.91 & 39.20 &  0.00 &  0.00 & 49.40 & 41.00 &  4.19 & 65.50 & 33.30 & 59.90 & 37.24 \\
    \quad Dream-7B Instruct                    & 81.00 & 43.00 &  0.00 &  3.33 & 55.50 & 58.80 &  1.25 & 67.00 & 33.00 & 62.50 & 40.54 \\
    \quad SDAR-8B Chat                         & 91.30 & 78.60 & 16.67 & 10.00 & 78.70 & 72.00 & 13.44 & \underline{78.60} & 40.20 & 61.40 & 54.09 \\
    \quad Nemotron-Labs-Diffusion-8B-Instruct           & 91.58 & \underline{84.60} & \underline{30.00} & \underline{23.33} & \underline{83.54} & 82.01 & \underline{21.15} & 76.65 & 33.33 & 73.02 &  \underline{59.92} \\
    \quad \textbf{\model\unreg-8B\instruct (Ours)} &       &       &       &       &       &       &       &       &       &       &       \\
    \qquad DDPM (256 steps, $w{=}2,\tau{=}0.1$) & \sev{\underline{93.24}}{0.20} & \sev{79.80}{0.55} & \sev{23.67}{3.19} & \sev{18.00}{2.15} & \sev{77.99}{1.06} & \sev{80.40}{0.80} & \sev{20.79}{0.52} & \sev{77.91}{0.11} & \sev{42.37}{1.80} & \sev{\underline{79.83}}{0.42} & 59.40 \\
    \qquad DDIM (32 steps, $w{=}3,\tau{=}0.3$) & \sev{92.90}{0.18} & \sev{78.20}{0.33} & \sev{19.00}{0.97} & \sev{19.00}{2.50} & \sev{78.35}{1.62} & \sev{79.74}{0.98} & \sev{18.33}{0.61} & \sev{77.60}{0.09} & \sev{41.87}{1.74} & \sev{76.69}{0.38} & 58.17 \\
    \quad \textbf{\model-8B\instruct (Ours)} &       &       &       &       &       &       &       &       &       &       &       \\
    \qquad DDPM (256 steps, $w{=}3,\tau{=}0.1$) & \sev{91.96}{0.21} & \sev{78.40}{0.67} & \sev{23.33}{2.67} & \sev{17.00}{1.75} & \sev{82.50}{1.17} & \sev{\underline{82.04}}{0.67} & \sev{20.09}{0.63} & \sev{76.57}{0.08} & \sev{42.37}{1.17} & \sev{77.18}{0.38} &  59.14 \\
    \qquad DDIM (32 steps, $w{=}3,\tau{=}0.3$) & \sev{91.90}{0.17} & \sev{76.70}{0.67} & \sev{21.67}{3.02} & \sev{14.33}{3.53} & \sev{80.55}{0.58} & \sev{79.47}{0.55} & \sev{19.38}{0.48} & \sev{76.15}{0.08} & \sev{\underline{42.93}}{1.75} & \sev{75.96}{0.32} & 57.90 \\
    \bottomrule
    \end{tabular}%
    }
\end{table}

\section{Conclusions}
\label{sec:conclusion}
We introduced \nemmodel, a family of large-scale blockwise continuous dLMs obtained by efficient continual pretraining from an AR checkpoint, together with a reverse-diffusion sampling recipe that combines DDPM/DDIM samplers with CFG, ST, and SC.
In diffusion mode, \model (3B/8B) outperforms same-size masked dLMs and AR baselines trained under an aligned pipeline, both after pretraining and after SFT, with the most consistent gains on coding.
A sphere-normalized embedding regularization further improves low-NFE sampling and makes model distillation markedly more effective.

Beyond accuracy, the smoothness of continuous trajectories buys efficiency.
Distribution matching is the strongest of the few-step distillation methods we study: using only 8 NFEs ($\text{TPF}=4$), \model\distill-3B almost matches both NLD \citep{fu2026nemotron} and our best-performing 256-NFE DDPM teacher, and it does so without collapsing the teacher's output distribution, preserving diversity and yielding pass@$k$ that keeps improving as $k$ grows. %
Because \model carries a causal AR stream on the same backbone, the \textit{distilled} student can also serve as its own one-call drafter: self-speculative decoding returns the greedy AR output at a $3.46\times$ average speedup over AR decoding at 8B, against $3.04\times$ for NLD.%

Together, these results establish continuous dLMs as a practical path to efficient language generation.
Looking ahead, the controllable quality--diversity trade-off and the shared diffusion/AR backbone make them particularly attractive for RL-based post-training, which we leave to future work.

\bibliography{iclr2027_conference}
\bibliographystyle{iclr2027_conference}

\appendix
\crefalias{section}{appendix}
\crefalias{subsection}{appendix}
\crefalias{subsubsection}{appendix}

\clearpage

\etocdepthtag.toc{appendix}
\etocsettagdepth{main}{none}
\etocsettagdepth{appendix}{subsubsection}
\tableofcontents

\clearpage

\section{Related Work}
\label{app:related-work}

[Back to~\cref{sec:introduction} or~\cref{sec:prelim-cdlm} in the main paper.]

This appendix provides a literature review on continuous diffusion language models, the family to which \model belongs.

\paragraph{Continuous diffusion language models.}
Continuous dLMs lift discrete tokens into a Euclidean space and run a Gaussian (or Riemannian) diffusion there. A central attraction for doing so is that the machinery developed for image diffusion, such as efficient ODE solvers, guidance, and distillation techniques, transfers with little modification. Existing methods differ mainly in \emph{where} the diffusion process lives:

\begin{itemize}
    \item Diffusing \emph{token embeddings}. Diffusion-LM \citep{li2022diffusion} pioneered this approach with Gaussian diffusion over learned word embeddings and a rounding step back to tokens; CDCD \citep{dieleman2022continuous} refined the objective with score interpolation and a categorical cross-entropy loss over softmax logits, and self-conditioned embedding diffusion \citep{strudel2022self} added self-conditioning on the previous denoising estimate. Plaid \citep{gulrajani2023plaid} tied these ideas together around a likelihood target by pairing low-dimensional sphere-normalized embeddings with a learned noise schedule and categorical reparameterization; RePlaid \citep{yang2026continuous} aligns this recipe with modern DiT architectures and shows it scales competitively with discrete dLMs. Meanwhile, LangFlow \citep{chen2026langflow} follows the same learnable-embedding but uses cross-entropy training loss, while \citet{wu2026discrete} adopts a stochastic-localization formulation. $\mathbb{S}$-FLM \citep{deschenaux2026language} and Spherical Flows \citep{chemseddine2026spherical} likewise use learnable token embeddings but further constrain the diffusion trajectory itself to the embedding hypersphere. A parallel strand fixes the embedding to be \emph{one-hot} rather than learned: FLM \citep{lee2026flow} and DFM \citep{potaptchik2026discrete} run flow matching from a Gaussian source to one-hot vocabulary targets in $\R^V$, and further distill the flow into a few-step flow map for fast generation.
    \item Diffusing an \emph{autoencoded latent}. Here, a text autoencoder produces a compact continuous latent variable and diffusion is run in that latent space. LD4LG \citep{lovelace2023latent} runs latent diffusion on a frozen encoder--decoder (BART and T5); COSMOS \citep{meshchaninov2025cosmos}, ELF \citep{hu2026elf} Cola-DLM \citep{guo2026continuous}, and AURORA-LM \citep{liang2026aurora} scale this axis with different latent designs and training objectives; LDLM \citep{meshchaninov2026train} learns the encoder and the diffusion prior jointly rather than sequentially.
    \item Diffusing on the \emph{simplex or other statistical manifolds}. These methods parameterize a categorical distribution over the vocabulary and diffuse its parameters directly, keeping the noise process on-manifold and avoiding a rounding step. SSD-LM \citep{han2022ssd} runs simplex diffusion in a semi-autoregressive block structure, Dirichlet diffusion \citep{avdeyev2023dirichlet} uses a Jacobi diffusion process on the probability simplex, and later variants \citep{cheng2024categorical, davis2024fisher, jo2025continuous} extend the geometry to Fisher--Rao and Riemannian flow matching.
    \item Reaching discrete data through \emph{dequantization or a Gaussian--discrete duality}. Argmax flows \citep{hoogeboom2021argmax} lift categorical data into a continuous space via argmax-based variational dequantization, so continuous flows can be trained under a discrete likelihood; analog bits \citep{chen2023analog}, BitLM \citep{zhuang2026bitlm}, and CoBit \citep{batzolis2026cobit} instead map each token to a bitstring and diffuse from there; Duo \citep{sahoo2025the} establishes a duality between continuous Gaussian and uniform-state discrete diffusion, letting the two families share training and sampling machinery.
\end{itemize}
\model sits in the token-embedding class, inheriting the learnable embedding and variational diffusion process of Plaid and RePlaid.

\paragraph{Scaling continuous dLMs to billions of parameters.} Recent work pushes continuous dLMs into the billion-parameter regime, but along axes different from ours: for example, Cola-DLM \citep{guo2026continuous} ($\sim$2B, hierarchical latent-prior modeling), AURORA-LM \citep{liang2026aurora} (1B, flow matching over an autoencoded latent), and Categorical Flow Maps \citep{davis2026scaling} (1.7B, self-distilled for few-step generation). A separate line changes the base representation and scales through an existing LM backbone: BitLM \citep{zhuang2026bitlm} places a bitwise diffusion head atop a causal LM up to 8B, TESS~2 \citep{tae2502tess} adapts an autoregressive Mistral-7B into a generalist diffusion model with a simplex cross-entropy loss, and DSL-LLaDA \citep{yu2026dsl} scales continuous denoising via discrete stochastic localization \citep{wu2026discrete} on top of the LLaDA-8B masked-diffusion backbone \citep{nie2025large}. In contrast, \model is, to our knowledge, the first likelihood-trained continuous dLM at 3B/8B, and the first evaluated head-to-head with a matched masked dLM \citep{fu2026nemotron} on conditional reasoning and code.

\paragraph{CFG in dLMs.} Continuous dLMs realize the guidance condition in two ways. When there is no prompt (e.g., unconditional generation), several works repurpose the self-conditioning signal as the condition: ELF \citep{hu2026elf} and AURORA-LM \citep{liang2026aurora} extrapolate between predictions made with and without the previous-step estimate. For conditional generation, DFM \citep{potaptchik2026discrete}, AURORA-LM \citep{liang2026aurora}, Cola-DLM \citep{guo2026continuous}, and BitLM \citep{zhuang2026bitlm} extrapolate between a prompt-conditioned and an unconditional prediction, as in image diffusion models. A cost for this is that enabling it requires dropping the prompt context during training. FLM \citep{lee2026flow} also leverages autoguidance \citep{karras2024guiding} for prediction sharpening.
For discrete dLMs, LLaDA \citep{nie2025large} popularized an ``unsupervised'' CFG that runs one forward with full prompt and one with the prompt masked, then linearly extrapolates the two logit vectors before renormalizing; this has since been refined by more general discrete guidance mechanisms \citep{schiff2025simple, nisonoff2024unlocking}.
Our steering (\cref{sec:inference-cfg-st}) differs on three points: (\textit{i}) the guidance condition is the clean prefix itself, and the unconditional branch simply drops that prefix so the noisy block attends only within itself; (\textit{ii}) the extrapolation acts on the predicted clean embedding (equivalently the score), not on logits; (\textit{iii}) it is a pure inference-time operation that requires no train-time dropout because sequence packing leaves the first block without context.
\paragraph{Autoregressive and block video diffusion.} Our block-causal factorization (i.e., autoregressive across blocks, bidirectional Gaussian denoising within a block over a KV-cacheable clean prefix) closely mirrors recent video generation models that couple causal structure across frames with denoising within them \citep{chen2024diffusion, yin2025from, huang2025self}. The bidirectional video-diffusion lineage \citep{ho2022video} was the original target of this conversion, and the autoregressive arm of Cosmos \citep{agarwal2025cosmos} likewise structures video generation causally across frames. Diffusion Forcing \citep{chen2024diffusion} trains a causal model with independent per-frame noise levels, unifying next-token prediction with full-sequence diffusion and enabling stable rollouts past the training horizon. CausVid \citep{yin2025from} and Self Forcing \citep{huang2025self} convert bidirectional video diffusion into causal, KV-cached autoregressive generators for fast streaming synthesis while curbing error accumulation along the rollout. These works share our central design tension (i.e., coupling causal, autoregressive structure with within-block diffusion) and several of their techniques (per-block noise schedules, rolling KV caches, distribution-matching distillation) transfer naturally to the language setting.

\clearpage

\clearpage
\section{Training and Inference Algorithms}
\label{app:algorithms}

[Back to~\cref{sec:pretrain} or~\cref{sec:inference} in the main paper.]

We collect the two procedures described in the main paper:
\begin{itemize}
    \item \cref{alg:train} is one optimization step of the blockwise NELBO~\cref{eq:block-nelbo,eq:nelbo} under the Horvitz--Thompson estimator~\cref{eq:ht} (\cref{sec:probabilistic});
    \item \cref{alg:sample} is blockwise reverse sampling (\cref{sec:inference-solver}) with CFG, ST, SC, and an optional commit stride (\cref{sec:inference-cfg-st}).
\end{itemize}

\begin{algorithm}[h]
\caption{\model training step for one sequence}
\label{alg:train}
\begin{algorithmic}[1]
\Require sequence $\x \in \{0,1\}^{L \times V}$;
block size $L'$, number of blocks $B = L/L'$;
denoiser parameters $\theta$ (transformer, heads, and up-projections);
embedding matrix $\Emb\in\R^{V\times d_e}$;
noise schedule endpoints $\gamma_0 < \gamma_1$ and monotone interior shape $\tilde\gamma_\psi:[0,1]\to[0,1]$, giving $\gamma(t) = \gamma_0 + (\gamma_1 - \gamma_0)\tilde\gamma_\psi(t)$;
reconstruction probability $p_{\text{rec}}\in(0,1)$ (default $0.1$);
self-conditioning probability $p_{\selfcond}\in(0,1)$ (default $0.25$);
diffusion loss weight $w_{\text{diff}}>0$ and optional AR-loss weight $w_{\text{ar}}\ge0$;
optional embedding regularizer $s_\star\in\{s,s_{\mathrm{MS}}\}$ with weight $\lambda_{\text{susreg}}\ge0$;
optional repulsion weight $\lambda_{\mathrm{rep}}\ge0$ with radius $r$ computed from $\sg(\gamma_0)$ (settings in \cref{app:pretrain-susreg})

\State $\e \gets \x\Emb \in \R^{L \times d_e}$ \Comment{Embedding representation, each row has unit-norm}
\For{$b = 1, \dots, B$} \Comment{independent noise level and estimator branch per block}
  \State $R^{(\bcolor{b})} \sim \mathrm{Ber}(p_{\text{rec}})$
  \State $t^{(\bcolor{b})} \gets 0$ \textbf{if} $R^{(\bcolor{b})} = 1$ \textbf{else} $t^{(\bcolor{b})} \sim \unif[0,1]$
  \State $\beps^{(\bcolor{b})} \sim \mathcal{N}(\zero, \I)$, \quad
         $\z^{(\bcolor{b})} \gets \alpha_{t^{(\bcolor{b})}} \e^{(\bcolor{b})} + \sigma_{t^{(\bcolor{b})}} \beps^{(\bcolor{b})}$
         \Comment{Gaussian noising \cref{eq:vdm-forward}}
\EndFor
\State $S \sim \mathrm{Ber}(p_{\selfcond})$, \quad $\e_{\selfcond} \gets \zero$
\If{$S = 1$} \Comment{detached SC pass}
  \State $\e_{\selfcond} \gets \sg\big(\x_\theta(\{\z^{(\bcolor{b})}\}, \{t^{(\bcolor{b})}\}; \x, \zero)\,\Emb\big)$
\EndIf
\State $\x_\theta \gets \x_\theta(\{\z^{(\bcolor{b})}\}, \{t^{(\bcolor{b})}\}; \x, \e_{\selfcond})$,
       \quad $\e_\theta \gets \x_\theta \Emb$
       \Comment{one forward for all blocks; block-causal mask}
\For{$b = 1, \dots, B$}
  \State $\cL_{\text{rec}}^{(\bcolor{b})} \gets -\sum_{l=1}^{L'} \log \langle \x_\theta^{(\bcolor{b},l)}, \x^{(\bcolor{b},l)} \rangle$
         \Comment{used only when $R^{(\bcolor{b})} = 1$, i.e., $t^{(\bcolor{b})} = 0$}
  \State $\cL_{\text{diff}}^{(\bcolor{b})} \gets -\frac{1}{2}\,\SNR'(t^{(\bcolor{b})})\,\|\e_\theta^{(\bcolor{b})} - \e^{(\bcolor{b})}\|^2$
         \Comment{used only when $R^{(\bcolor{b})} = 0$}
  \State $\cL_{\text{prior}}^{(\bcolor{b})} \gets \tfrac{L'}{2}\big[d_e(\sigma_1^2 - 1 - \log \sigma_1^2) + \alpha_1^2\big]$
         \Comment{closed form of $\KL(q(\z_1^{(\bcolor{b})} \mid \x^{(\bcolor{b})}) \,\|\, p(\z_1^{(\bcolor{b})}))$}
  \State $\cL^{(\bcolor{b})} \gets
         \frac{1 - R^{(\bcolor{b})}}{1 - p_{\text{rec}}} \cL_{\text{diff}}^{(\bcolor{b})}
       + \frac{R^{(\bcolor{b})}}{p_{\text{rec}}} \cL_{\text{rec}}^{(\bcolor{b})}
       + \cL_{\text{prior}}^{(\bcolor{b})}$
         \Comment{Horvitz--Thompson estimator \cref{eq:ht}}
\EndFor
\State $\cL_{\text{BD}} \gets \sum_{b=1}^{B} \cL^{(\bcolor{b})}$
       \Comment{block-diffusion NELBO estimator \cref{eq:block-nelbo}}
\If{$\lambda_{\text{susreg}} > 0$} %
  \State $\cL_{\text{BD}} \gets \cL_{\text{BD}} + \lambda_{\text{susreg}}\, L\, s_\star(\Emb)$
         \Comment{\model\preg: $\lambda_{\text{susreg}}=0.005$, $s_\star=s$; \model: $\lambda_{\text{susreg}}=1$, $s_\star=s_{\mathrm{MS}}$}
\EndIf
\If{$\lambda_{\mathrm{rep}} > 0$} %
  \State $\cL_{\text{BD}} \gets \cL_{\text{BD}} + \lambda_{\mathrm{rep}}\, L\, s_{\mathrm{rep}}(\Emb)$
         \Comment{local repulsion loss \cref{eq:repulsion}, used by \model only}
\EndIf
\State $\cL \gets w_{\text{diff}}\, \cL_{\text{BD}} + w_{\text{ar}}\, \cL_{\text{ar}}(\x)$
       \Comment{joint objective (\cref{sec:pretrain-from-scratch})
       }
\State $\cL_{\text{var}} \gets \frac{(1-S)w_{\text{diff}}}{(1-p_{\text{rec}})L'}\sum_{b\,:\,R^{(\bcolor{b})} = 0} (\cL_{\text{diff}}^{(\bcolor{b})})^2$
       \Comment{Noise schedule interior shape learned to minimize MC variance}
\State update $\theta$ with $\nabla_\theta \cL$
       \Comment{backbone group, peak lr $10^{-5}$}
\State update $\Emb$ with $\nabla_{\Emb} \cL$
       \Comment{Riemannian Adam for \model, AdamW otherwise; no weight decay}
\State update $(\gamma_0,\gamma_1)$ with $\nabla_{(\gamma_0,\gamma_1)}\cL$
       \Comment{AdamW, weight decay $0.1$; ordering projection for \model}
\State update $\psi$ with $\nabla_\psi \cL_{\text{var}}$
       \Comment{projected Adam for \model, AdamW otherwise; no weight decay}
\end{algorithmic}
\end{algorithm}

At training, all $B$ blocks are denoised in a \emph{single} forward pass: the offset block-causal mask of \cref{fig:teaser-pretrain-attention} makes block $b$'s noisy latent attend to the clean prefix $\x^{(\bcolor{<b})}$ and bidirectionally within itself, so a per-block time $t^{(\bcolor{b})}$ and a per-block Horvitz--Thompson coin $R^{(\bcolor{b})}$ cost no extra compute. At inference, the conditional and unconditional branches are likewise packed into one forward as in \cref{fig:cfg-attention}, so the guidance computation costs slightly more wall-clock time per step (\cref{tab:cfg-fwd-cost}) at batch size one.

The losses are sequence sums, so per-token embedding penalties carry a factor $L$. For MS-SUSReg, $\lambda_{\text{susreg}}=1$ because $s_{\mathrm{MS}}$ already includes the scale weights. SC-selected sequences stop NELBO gradients to the diffusion embedding and schedule while retaining regularizer gradients. MS-SUSReg balances the $25\%$ SC fraction across each global batch.

We further clarify on the learning of the noise schedule, since its two parts are optimized for two different objectives.
The endpoints $\gamma_0, \gamma_1$ enter $\cL_{\text{diff}}$ through $\SNR'(t)$ and $(\alpha_t, \sigma_t)$, so they receive an ordinary NELBO gradient and are trained to tighten the bound while the interior shape $\tilde\gamma_\psi$ does not. As noted in \cref{sec:prelim-cdlm}, the diffusion term is invariant to it in expectation, so $\nabla_\psi \cL$ is zero on average and carries only noise. Following \citet{kingma2021variational}, $\psi$ is instead trained to minimize the Monte-Carlo variance of the diffusion estimator, which for a mean-invariant estimator reduces to minimizing its second moment $\cL_{\text{var}}$ in \cref{alg:train}. In code, each block's schedule gradient is scaled by twice its detached mean diffusion loss; reconstruction blocks and SC-selected sequences do not update the shape.
Variant-specific optimizer and learning-rate settings are given in \cref{app:pretrain-susreg}.

\begin{algorithm}[h]
\caption{Blockwise reverse sampling of \model with CFG, ST, SC, and commit stride}
\label{alg:sample}
\begin{algorithmic}[1]
\Require prompt $\p$; block size $L'$; commit stride $S \in \{1, \dots, L'\}$ ($S = L'$ recovers plain block decoding, \cref{app:inference-stride-sampling});
total generation length $T_{\text{gen}}$; steps $N$ with
discretization $1 = t_0 > t_1 > \cdots > t_N = 0$ (\cref{app:inference-time-disc}); guidance weight $w \geq 1$;
score temperature $\tau \in (0,1]$; sampler $\in \{\text{DDPM}, \text{DDIM}\}$
\State $\c \gets \p$; populate the causal KV cache from $\p$; $b \gets 0$
\While{$|\r| < T_{\text{gen}}$} \Comment{$|\r|$: number of tokens committed so far}
  \State $b \gets b + 1$
  \State $\z_{t_0}^{(\bcolor{b})} \sim \mathcal{N}(\zero, \I) \in \R^{L' \times d_e}$,
         \quad $\e_{\selfcond} \gets \zero$
  \For{$i = 1, \dots, N$}
    \State $t \gets t_{i-1}$, \quad $s \gets t_i$
    \State $(\e_{\mathrm{c}}, \e_{\varnothing}) \gets
           (\e_\theta(\z_t^{(\bcolor{b})}, t; \c, \e_{\selfcond}),\;
                \e_\theta(\z_t^{(\bcolor{b})}, t; \varnothing, \zero))$
           \Comment{one packed forward, \cref{fig:cfg-attention}}
    \State $\e_{\selfcond} \gets \e_{\mathrm{c}}$
           \Comment{SC reuses the \emph{unsteered} conditional mean, \cref{fig:cfg-steering}}
    \State $\hat\e_\theta \gets \frac{w}{\tau} \e_{\mathrm{c}} + \frac{1-w}{\tau} \e_{\varnothing}
           + \big(1 - \frac{1}{\tau}\big) \frac{\z_t^{(\bcolor{b})}}{\alpha_t}$
           \Comment{Embedding mean prediction after CFG \& ST}
    \If{sampler is DDIM}
      \State $\hat\beps_\theta \gets \frac{1}{\sigma_t}(\z_t^{(\bcolor{b})} - \alpha_t \hat\e_\theta)$,
             \quad $\z_s^{(\bcolor{b})} \gets \alpha_s \hat\e_\theta + \sigma_s \hat\beps_\theta$
             \Comment{\cref{eq:ddim}}
    \Else
      \State $\kappa \gets 1 - \ee^{\gamma(s) - \gamma(t)}$, \quad $\beps \sim \mathcal{N}(\zero, \I)$
      \State $\z_s^{(\bcolor{b})} \gets (1 - \kappa)\frac{\alpha_s}{\alpha_t}\z_t^{(\bcolor{b})}
             + \kappa\,\alpha_s \hat\e_\theta + \sqrt{\kappa(1 - \alpha_s^2)}\,\beps$
             \Comment{\cref{eq:ddpm}}
    \EndIf
  \EndFor
  \State $k_b \gets \min(S,\; T_{\text{gen}} - |\r|)$
         \Comment{tokens to commit: the last window may commit fewer than $S$}
  \State $\r^{(\bcolor{b},l)} \gets \argmax_{v} \x_\theta^{l, v}(\z_{t_N}^{(\bcolor{b})}, t_N; \c, \e_{\selfcond})$, \; $l \in \{1,...,k_b\}$
         \Comment{argmax decoding}
  \State $\c \gets (\c, \r^{(\bcolor{b})})$; append the KV of $\r^{(\bcolor{b})}$ to the causal cache; discard the tail
         \Comment{the $L' - k_b$ tail positions are re-drawn as fresh noise in the next window}
\EndWhile
\State \Return $\r = (\r^{(1)}, \dots, \r^{(b)})$
\end{algorithmic}
\end{algorithm}

The outer loop of \cref{alg:sample} runs $B := \lceil T_{\text{gen}} / S \rceil$ times, and each window costs $N + 1$ denoiser forward passes ($N$ diffusion steps plus one decoding readout), for a total of $B(N + 1)$ forward passes. The effective NFE per committed token is thus $\approx (N + 1)/S$, 
see \cref{app:inference-stride-sampling} for a full compute-vs-quality sweep.
Setting $w = 1$ disables CFG, $\tau = 1$ disables ST, and never updating $\e_{\selfcond}$ (i.e., leaving it as $\zero$) disables SC; each of these recovers a strictly weaker sampler, as shown in \cref{fig:ddpm-256-sweep}.

\clearpage
\section{Implementation Details and Additional Results for Pretraining}
\label{app:pretrain}

[Back to~\cref{sec:pretrain} in the main paper.]

\paragraph{Architecture.} Following NLD~\citep{fu2026nemotron}, we adopt the transformer architecture of Ministral~\citep{liu2026ministral}: 
\begin{itemize}
    \item 3B: 26 layers, width 3072, FFN
9216, 32 query and 8 key/value heads (head dimension 128), SwiGLU, RMSNorm, YaRN-scaled rotary
embeddings ($\theta = 10^6$), and untied embeddings. This adds up to 3.0B non-embedding and 3.8B total
parameters. 
    \item 8B: 34 layers, width 4096, FFN 14336, same 32 query and 8 key/value heads (head
dimension 128) and same SwiGLU, RMSNorm, YaRN rotary embeddings and untied embeddings. This adds up to 7.4B
non-embedding and 8.5B total parameters.
\end{itemize}

\paragraph{Optimization.} For \model\unreg/\model\preg, $\tilde{\gamma}$ and $\Emb$ use AdamW with $\beta_1 = 0.9$, $\beta_2 = 0.95$, no weight decay (rows of $\Emb$ unit-norm), and peak learning rate $\eta_{\text{peak}}=1\times10^{-2}$; $\gamma_0, \gamma_1$ use the same setting but with weight decay $0.1$~\citep{yang2026continuous}. For other parameters, we follow NLD and use AdamW with $\beta_1 = 0.9$, $\beta_2 = 0.95$, weight decay $0.1$, and $\eta_{\text{peak}}=1\times10^{-5}$. For all parameters, we use gradient clipping value $1.0$. The learning rate schedule is Warmup-Stable-Decay over 150k steps: $1\%$ linear warmup to $\eta_{\text{peak}}$, constant
through step 120k, then exponential decay to $10\%$ of $\eta_{\text{peak}}$.

\paragraph{Self-conditioning (SC).} SC injects the model's own detached clean-embedding prediction
into the projected noised input stream with a bias-free, zero-initialized linear layer mapping $\R^{d_e}$ to the hidden width $\R^{d}$: a first pass with $\e_{\selfcond} = \zero$, then a second pass with $\sqrt{d_e}\,\sg(\e_{\selfcond})$ through that linear layer.

\paragraph{Time conditioning.} For diffusion stream, a single $\gamma$ is sampled per block and shared across its tokens. We encode $\gamma$ using fixed sinusoidal features with frequencies $\exp(\mathrm{linspace}(-5,5,32))$, concatenating $[\sin,\cos]$ into 64 dimensions. A bias-free, zero-initialized linear layer projects this encoding to the hidden width, where it is added to the projected noised input.

\paragraph{Numerical precision.} Transformer weights and matrix multiplications, including the linear logit heads, use \texttt{BF16}; gradient accumulation and communication use \texttt{FP32}. The diffusion embedding, noise schedule, gamma endpoints, and input/time/SC projections are stored in \texttt{FP32} for MS-SUSReg and \texttt{BF16} for \model\unreg/\model\preg. Embedding normalization, corruption, input/time/SC projections, and logit postprocessing compute in \texttt{FP32}; projected inputs are cast to \texttt{BF16} for the transformer. Schedule arithmetic and NELBO accumulation use \texttt{FP64}, following~\citet{gulrajani2023plaid, yang2026continuous}. Adam moments and master copies of \texttt{BF16} weights are \texttt{FP32}. Optimizer state is sharded across data-parallel ranks, except for MS-SUSReg's replicated schedule and endpoint states.

\paragraph{AR regularization.} For training \model\unreg-3B, we searched over $w_{\text{diff}} \in \{0.3, 1.0\}$. Interestingly, when $w_{\text{diff}}=1.0$, adding the AR regularization even improved diffusion PPL (\cref{fig:dlm-weight-schedule}). However, we found that switching to $w_{\text{diff}}=0.3$ at 200B tokens (equivalently, $100\text{k}$ steps) offers the best convergence for both losses (\cref{fig:dlm-weight-schedule}). We thus adopted this schedule ($w_{\text{diff}}=1.0$ for the first $100\text{k}$ steps and $w_{\text{diff}}=0.3$ for the last $50\text{k}$ steps) for all of our \model\unreg-3B/8B and \model\preg-3B/8B models (see \cref{app:pretrain-susreg}). For \model-3B/8B, we keep $w_{\text{diff}}=1.0$ all the way through $150\text{k}$ steps for simplicity of the training objective.

\subsection{Learning Regularized Embeddings}
\label{app:pretrain-susreg}

[Back to~\cref{sec:few-step-susreg} in the main paper.]

The central goal for our embedding-regularized model variants is to improve the low-NFE performance and distillability by improving the structure of the embeddings.
To this end, we develop the toolbox described in the following: A multiscale variant of Sketched Uniform Spherical Regularization (SUSReg) and a local repulsion loss to combat global and local embedding collapse; a modified Riemannian Adam optimizer step for the embedding table that respects the per-row hyperspherical geometry; a simpler noise schedule parameterization based on Bernstein polynomials that can be trivially initialized to a linear (log-SNR) shape, together with a custom projected optimizer step; and a cosine learning rate schedule for the embeddings to aid their stable learning.

\paragraph{Sketched Uniform Spherical Regularization (SUSReg).}

To avoid the $V=131072$ embedding vectors in $d_e=16$ dimensional unit sphere being too tightly clustered for accurate decoding (as observed in their low-rank structure in \citet{yang2026continuous}), we propose to add an optional regularization term based on SUSReg \citep{nicollier2026sphere} to the embeddings, which regularizes their distribution $p_\Emb := \frac{1}{V}\sum_{v=1}^{V}\delta_{\e_v}$ toward $\unif(\mathbb{S}^{d_e-1})$, the uniform distribution on the unit sphere in $\R^{d_e}$.

For $M$ directions $\a_m\sim\unif(\mathbb{S}^{d_e-1})$, we compare the empirical characteristic function of each one-dimensional projection, 
\begin{align*}
    \phi_{\a_m}(t) = \frac{1}{V}\sum_v \ee^{\i t \langle\a_m, \e_v\rangle},
\end{align*}
to the target characteristic function of a one-dimensional projection of $\unif(\mathbb{S}^{d_e-1})$, whose marginal density is known in closed form:
\begin{align*}
    \phi_{d_e}(t) = \int_{-1}^{1} \cos(t x)\rho_{d_e}(x)\,\d x,
    \quad
    \rho_{d_e}(x)
    =\frac{\Gamma(\frac{d_e}{2})}{\sqrt{\pi}\Gamma(\frac{d_e-1}{2})}
    (1-x^2)^{\frac{d_e-3}{2}}\one_{|x|\le1},
\end{align*}
under an Epps--Pulley Gaussian-weighted $L^2$ metric:
\begin{align}\label{eq:susreg}
s(\Emb) := \frac{V}{M}\sum_{m=1}^{M}\int_{-t_{\max}}^{t_{\max}}
\bigl|\phi_{\a_m}(t) - \phi_{d_e}(t)\bigr|^2\ee^{-t^2/2}\,\d t.
\end{align}
The multiplier $V$ compensates for the $1/V$ scaling of the expected squared characteristic-function discrepancy under i.i.d. uniform embeddings.
We add $\lambda_\text{susreg}s(\Emb)$ per token on the diffusion arm, contributing $w_{\mathrm{diff}}\lambda_\text{susreg}s(\Emb)$ to the joint objective (\cref{alg:train}). Plain SUSReg-3B/8B uses $\lambda_\text{susreg}=0.005$, $M=512$ freshly sampled directions, and $P=17$ trapezoidal points over $[-5,5]$. The target characteristic function is cached using 512-point numerical quadrature. The $O(VM(d_e+P))$ cost is negligible relative to the transformer.

\begin{figure*}[h]
    \centering
    \includegraphics[width=0.8\linewidth]{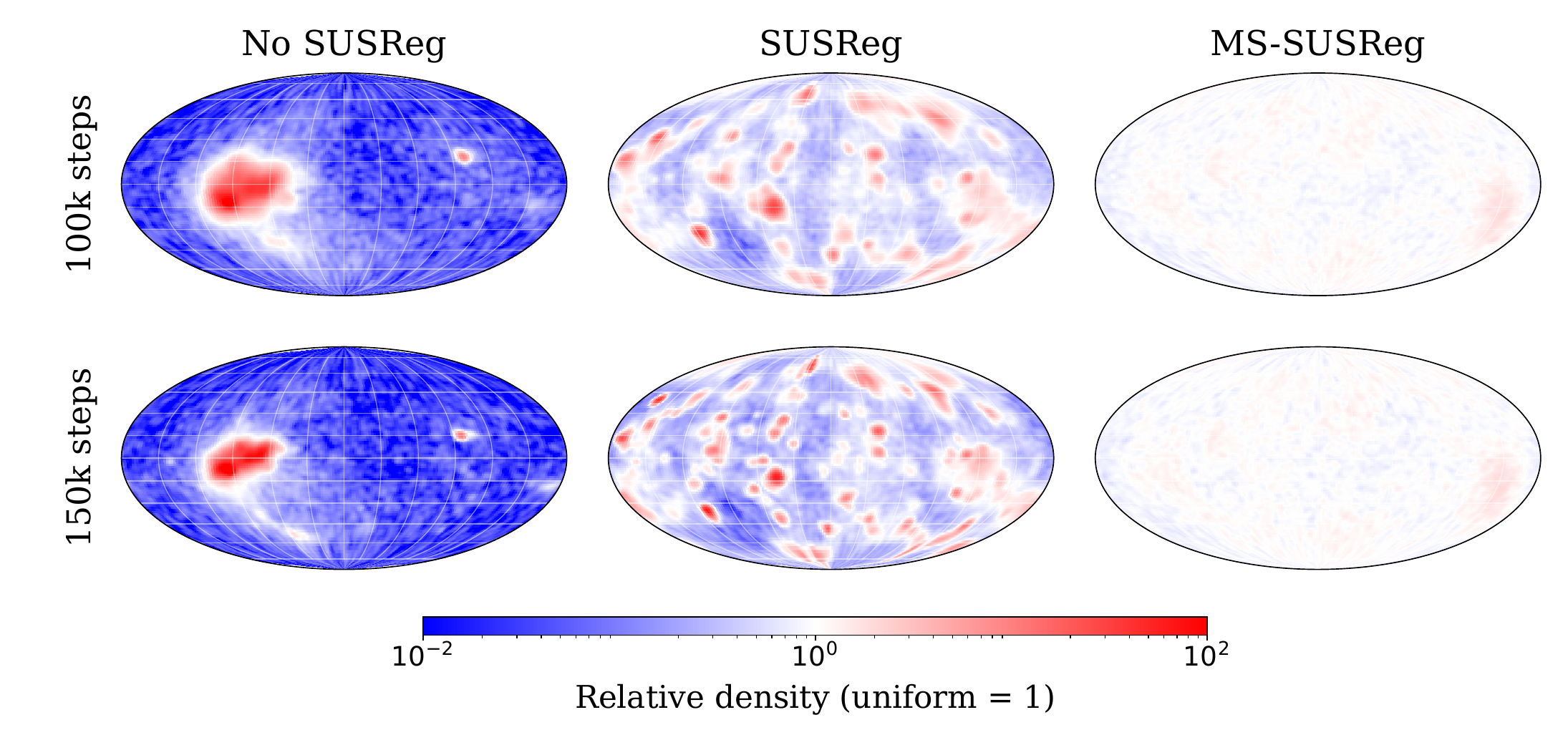}
    \caption{ Density plot of a random $\mathbb{S}^{15} \to \mathbb{S}^2$ Mollweide projection of all learned embeddings living on $\mathbb{S}^{15}$.
    The baseline collapses to a small subregion of the 16D space, while SUSReg distributes density more evently, but still contains many concentrated clusters. MS-SUSReg achieves a comparatively much more uniform distribution. Note that while the MS-SUSReg cloud looks uniform in density, it is far from simple random initialization, see the conceptual hierarchical clustering shown in \cref{fig:susreg-dendrogram}.
    The PCA effective dimensions of three embedding matrices at 150k steps are $3.4008,15.9593,15.9986$, respectively.
    }
    \label{fig:mollwide}
\end{figure*}

\begin{figure*}[h]
    \centering
    \includegraphics[width=\linewidth]{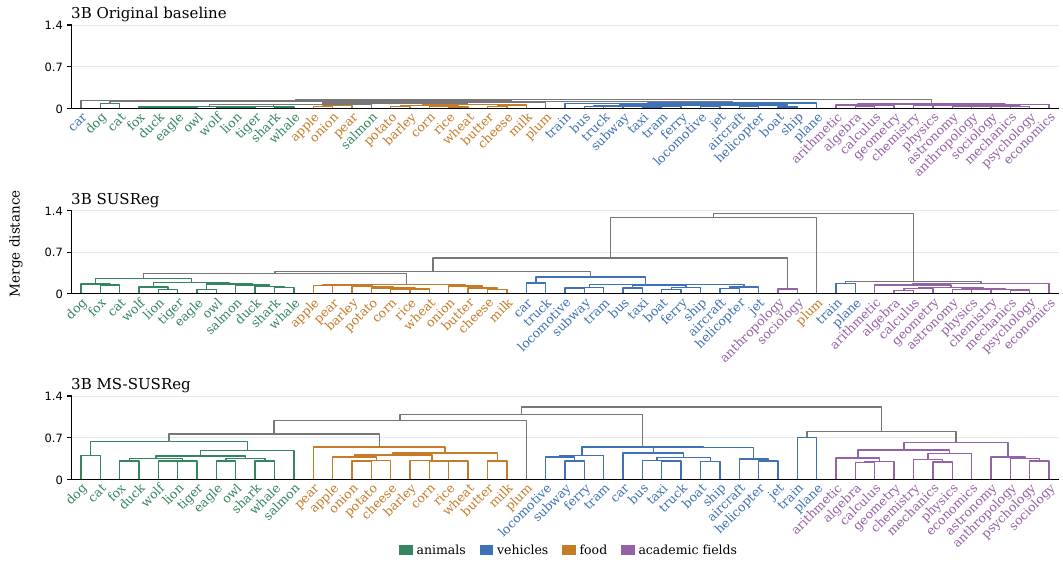}
    \caption{Dendrogram on a set of single-token example words, constructed by bottom-up average-linkage clustering of the selected unit-normalized embedding rows in the 3B model variants (Unregularized baseline, SUSReg, MS-SUSReg), using Euclidean distances, with the cluster merge distance shown on the y-axis. Each shown word represents the lowercase token form prefixed by a single space, e.g., \texttt{" dog"}. \textbf{Top:} The unregularized embedding table exhibits very low clustering distances, consistent with subspace collapse, and also contains some counterintuitive clusterings that mix categories. \textbf{Middle:} The SUSReg run greatly improves upon both issues, but most merges still occur at low distances and some unusual groupings (train, plum) remain. \textbf{Bottom:} The MS-SUSReg run exhibits the cleanest clustering in this example, corresponding well to natural categories, and also shows larger merge distances already at lower levels of the hierarchy.}
    \label{fig:susreg-dendrogram}
\end{figure*}

\paragraph{Multiscale-SUSReg (MS-SUSReg).}
The SUSReg regularizer \cref{eq:susreg} contains the Gaussian kernel $\ee^{-t^2/2}$, which corresponds to a specific low-pass weighting of the frequencies in the compared characteristic functions. This means the loss is more sensitive to larger-scale structural mismatches (global uniformity) than to the finer scales (local clustering). We generalize this by introducing a kernel width parameter $\beta > 0$:
\begin{align}\label{eq:susreg-beta}
s_\beta(\Emb) &:=
\frac{V}{M}\sum_{m=1}^{M}\int_{-\beta t_{\max}}^{\beta t_{\max}} \frac{1}{\beta}
\bigl|\phi_{\a_m}(t) - \phi_{d_e}(t)\bigr|^2\ee^{-t^2/(2\beta^2)}\,\d t
\\ &= \frac{V}{M}\sum_{m=1}^{M}\int_{-t_{\max}}^{t_{\max}}
\bigl|\phi_{\a_m}(\beta u) - \phi_{d_e}(\beta u)\bigr|^2\ee^{-u^2/2}\,\d u
\end{align}
where the second form is from a simple change of variables with $u=t/\beta$. %
The original SUSReg loss \cref{eq:susreg} corresponds to $\beta=1$. To regularize both the global and the local structure, we can then sum $K$ loss terms like \cref{eq:susreg-beta} with different weights $w_k$ and kernel widths $\beta_k$:
\begin{align}\label{eq:ms-susreg}
s_{\mathrm{MS}}(\Emb) :=
\frac{V}{M}\sum_{m=1}^{M} \sum_{k=1}^K w_k \int_{-t_{\max}}^{t_{\max}}
\bigl|\phi_{\a_m}(\beta_k t) - \phi_{d_e}(\beta_k t)\bigr|^2\ee^{-t^2/2}\,\d t
\end{align}
which we call the \emph{Multiscale-SUSReg} (MS-SUSReg) loss. Its scale weights $w_k$ are absolute coefficients: the per-token objective contains $w_{\mathrm{diff}}s_{\mathrm{MS}}(\Emb)$, without an additional SUSReg multiplier.
Each bandwidth uses $P_k=1+\lceil\beta_k(P-1)\rceil$ quadrature points to preserve frequency resolution: 17 and 257 points for $\beta=1$ and $16$. The scales share $M=512$ random directions and embedding projections, use $t_{\max}=5$, and are active from step zero. Cached targets give a total cost of $O(VM(d_e+\sum_k P_k))$.

\paragraph{Local repulsion regularization.}
MS-SUSReg encourages distributional uniformity through sampled projections, but does not explicitly enforce separation between individual embedding pairs. Nearly coincident embeddings can therefore remain difficult to distinguish under diffusion noise. To address this directly, we add a local separation penalty related to maximum hyperspherical separation \citep{liu2021learning}. Namely, we add a per-token weighted loss term $w_{\mathrm{diff}}\lambda_{\mathrm{rep}}s_{\mathrm{rep}}(\Emb)$, where
\begin{align}
s_{\mathrm{rep}}(\Emb) := \frac{1}{2V}\sum_{u<v}\max\{0,r-\|\e_u-\e_v\|_2\}^2
\label{eq:repulsion}
\end{align}
penalizes pairwise chord distances below an exclusion radius $r$. We set $r=\max\{\sqrt{2(1-c)},2\Phi^{-1}(1-p)\ee^{\gamma_0/2}\}$, where $\Phi$ is the standard normal CDF, and $c=0.9995$, $p=0.01$, and $\lambda_{\mathrm{rep}}=20$.
Unlike maximum hyperspherical separation, which maximizes the smallest pairwise distance, our penalty acts on all pairs within $r$.
The adaptive term depending on $\gamma_0$ defines a target separation corresponding to a per-competitor confusion probability of $p$ under the diffusion forward marginal distribution at the minimum noise level $\gamma_0$. We disable gradients through $r$ here so this loss only affects the embeddings and not $\gamma_0$.
We evaluate all unordered pairs in tiles across data-parallel workers without storing the full pairwise matrix. Both embedding regularizers act from step zero, including on SC-selected sequences, and are excluded from reported NELBO/perplexity.

\paragraph{Riemannian Adam step.}
Since the effective embeddings have unit norm, their geometry depends only on their directions. The naive embedding representation, which normalizes each unconstrained row on the fly when used, creates a per-row dependence of the effective learning rate on arbitrary underlying pre-normalization row norms. We find that these row norms grow over training, resulting in an uncontrolled LR decay on the embedding rows. With the following Riemannian Adam step, we optimize the embedding rows directly on the sphere to make every update respect the on-sphere constraint and avoid this uncontrolled influence on the embedding LR. For each row $\e_v$, we project its total training-loss gradient $\mathbf{g}_v$ onto the tangent space at $\e_v$ as $\mathbf{g}_v^\perp=\mathbf{g}_v-\langle\mathbf{g}_v,\e_v\rangle\e_v$. Similar to \citet{becigneul2019riemannian}, we adapt the step size separately for each embedding vector. We apply a bias-corrected on-manifold Adam variant using a tangent first moment and one scalar second moment per row, estimated from $\|\mathbf{g}_v^\perp\|_2^2/d_e$. This scalar second moment rescales all tangent directions equally. After the update, we project the row back onto the sphere by normalization and parallel-transport the first moment along the shortest spherical geodesic to the new tangent space, preserving momentum as the embedding direction changes.

\paragraph{Projected noise-schedule updates.} Our main regularized model (\model-3B/8B) uses a degree-32 Bernstein polynomial for the noise schedule shape with increments $\psi_i\ge10^{-3}/32$, $\sum_i\psi_i=1$:
\begin{align*}
\tilde\gamma_\psi(t) &= \sum_{j=0}^{32} c_j \binom{32}{j}t^j(1-t)^{32-j}, \qquad c_0=0, \quad c_j=\sum_{i=1}^{j}\psi_i.
\end{align*}
Initializing $\psi_i=1/32$ gives a linear schedule. Adam updates on the variance objective are projected onto this simplex; endpoint AdamW updates use weight decay $0.1$ and enforce $\gamma_1-\gamma_0\ge10^{-3}$. The embedding and shape have no weight decay. All three groups use Adam coefficients $(\beta_1,\beta_2)=(0.9,0.95)$ and $\epsilon=10^{-8}$.

\paragraph{Training details for the regularized models.}
\textit{For the plain SUSReg ablation models (\model\preg-3B/8B):}%
\footnote{``\textit{preg}'' stands for plain SUSReg.}
we add the described SUSReg loss in~\cref{eq:susreg} and freeze the embeddings and schedule parameters after 100k steps, but otherwise follow the original model's training recipe including the change to $w_{\mathrm{diff}} = 0.3$ at 100k steps.
\textit{For MS-SUSReg (\model-3B/8B):} we use $(\beta_1,w_1)=(1,10^{-2})$, $(\beta_2,w_2)=(16,10^{-4})$, and repulsion weight $\lambda_{\mathrm{rep}}=20$. The embedding uses Riemannian Adam; the schedule uses the projected updates above. Their learning rates warm up to $3\times10^{-3}$ over 1.5k steps, decay by cosine to $10^{-4}$ at 30k, remain constant through 120k, then follow WSD decay to $10^{-5}$ at 150k. Embeddings and schedules remain trainable throughout, with $w_{\mathrm{diff}}=1$. Global gradient-norm clipping is 3 for 3B and 1 for 8B.

\paragraph{Full pretrained results for all three variants.}
\Cref{tab:base-variants} reports the six generative reasoning tasks of \cref{tab:base-3b,tab:base-8b} for all three pretrained variants at both scales. In diffusion mode, plain SUSReg is roughly neutral against \model\unreg{} at DDPM-256 but gains $1.40$ (3B) and $0.79$ (8B) average points at DDIM-32, consistent with our hypothesis that uniform embeddings mainly buy robustness to coarse discretization; MS-SUSReg improves on both at every operating point and scale.

\begin{table}[h]
    \centering
    \caption{\textbf{Pass@$1$ accuracy of all three pretrained \model variants at 3B and 8B.} Same six generative reasoning tasks, harness, and operating points as \cref{tab:base-3b,tab:base-8b}. \model diffusion rows are mean (over $10$ sampling lanes) $\pm$ two standard errors. \textbf{Bold} marks the best value per task within each mode and model size.}
    \label{tab:base-variants}
    \resizebox{\linewidth}{!}{%
    \begin{tabular}{lccccccc}
    \toprule
    & \multicolumn{2}{c}{\textbf{Math}} & \multicolumn{4}{c}{\textbf{Coding}} & \\
    \cmidrule(lr){2-3} \cmidrule(lr){4-7}
    Method & GSM8K & Minerva Math & HumanEval & HumanEval+ & MBPP & MBPP+ & Avg. \\
    \midrule
    \multicolumn{8}{l}{\textbf{3B models}} \\
    \quad AR initialization & 83.70 & 66.28 & 51.22 & 48.78 & 60.60 & 77.78 & 64.73 \\
    \cmidrule(lr){1-8}
    \multicolumn{8}{l}{\emph{AR mode}} \\
    \quad Nemotron-Labs-Diffusion-3B & \B{83.78} & 65.06 & 53.05 & 47.56 & \B{60.80} & \B{78.84} & 64.85 \\
    \quad \model\unreg-3B & 83.40 & 65.10 & \B{57.32} & \B{51.83} & 59.00 & 76.19 & \B{65.47} \\
    \quad \model\preg-3B & 82.18 & \B{65.64} & 53.66 & 49.39 & 60.00 & 77.78 & 64.77 \\
    \quad \model-3B & 81.20 & 63.98 & 51.83 & 48.17 & 56.80 & 75.66 & 62.94 \\
    \cmidrule(lr){1-8}
    \multicolumn{8}{l}{\emph{Diffusion mode}} \\
    \quad Nemotron-Labs-Diffusion-3B & 83.40 & 63.96 & 56.10 & 53.05 & 57.60 & \B{74.34} & 64.74 \\
    \quad \model\unreg-3B & & & & & & & \\
    \qquad DDPM $256$ steps/block ($w{=}2,\tau{=}0.1$) & \se{84.50}{0.31} & \se{64.60}{0.28} & \se{\B{59.45}}{1.30} & \se{\B{53.84}}{1.29} & \se{52.32}{0.73} & \se{71.32}{0.91} & 64.34 \\
    \qquad DDIM $32$ steps/block ($w{=}3,\tau{=}0.3$) & \se{81.91}{0.32} & \se{62.76}{0.18} & \se{56.59}{1.31} & \se{50.67}{1.00} & \se{47.26}{0.46} & \se{63.49}{0.77} & 60.45 \\
    \quad \model\preg-3B & & & & & & & \\
    \qquad DDPM $256$ steps/block ($w{=}3,\tau{=}0.2$) & \se{84.61}{0.20} & \se{63.82}{0.39} & \se{56.59}{1.38} & \se{50.43}{1.02} & \se{56.34}{0.70} & \se{73.52}{0.99} & 64.22 \\
    \qquad DDIM $32$ steps/block ($w{=}4,\tau{=}0.4$) & \se{83.17}{0.30} & \se{63.28}{0.15} & \se{55.24}{1.36} & \se{50.91}{0.99} & \se{51.64}{0.57} & \se{66.88}{0.29} & 61.85 \\
    \quad \model-3B & & & & & & & \\
    \qquad DDPM $256$ steps/block ($w{=}3,\tau{=}0.1$) & \se{\B{85.57}}{0.29} & \se{\B{64.98}}{0.19} & \se{57.26}{0.82} & \se{\B{53.84}}{0.55} & \se{\B{58.00}}{0.43} & \se{73.62}{0.51} & \B{65.55} \\
    \qquad DDIM $32$ steps/block ($w{=}3,\tau{=}0.3$) & \se{84.48}{0.21} & \se{64.52}{0.21} & \se{55.37}{1.28} & \se{51.71}{1.23} & \se{53.00}{0.34} & \se{71.30}{0.45} & 63.40 \\
    \midrule
    \multicolumn{8}{l}{\textbf{8B models}} \\
    \quad AR initialization & 85.90 & 69.18 & 59.15 & 53.05 & 68.20 & 83.33 & 69.80 \\
    \cmidrule(lr){1-8}
    \multicolumn{8}{l}{\emph{AR mode}} \\
    \quad Nemotron-Labs-Diffusion-8B & \B{88.25} & 66.00 & \B{60.37} & \B{53.05} & \B{68.20} & \B{82.54} & \B{69.73} \\
    \quad \model\unreg-8B & 86.88 & 66.72 & 51.83 & 48.78 & 64.00 & 80.95 & 66.53 \\
    \quad \model\preg-8B & 86.66 & \B{66.92} & 53.66 & 50.00 & 63.80 & 79.63 & 66.78 \\
    \quad \model-8B & 84.84 & 65.18 & 56.10 & 52.44 & 63.80 & \B{82.54} & 67.48 \\
    \cmidrule(lr){1-8}
    \multicolumn{8}{l}{\emph{Diffusion mode}} \\
    \quad Nemotron-Labs-Diffusion-8B & 87.26 & 65.16 & 62.80 & 57.32 & \B{67.00} & \B{81.75} & 70.22 \\
    \quad \model\unreg-8B & & & & & & & \\
    \qquad DDPM $256$ steps/block ($w{=}2,\tau{=}0.1$) & \se{\B{89.07}}{0.28} & \se{\B{67.47}}{0.20} & \se{61.40}{1.73} & \se{57.44}{1.62} & \se{64.78}{0.54} & \se{77.83}{0.72} & 69.67 \\
    \qquad DDIM $32$ steps/block ($w{=}3,\tau{=}0.3$) & \se{87.78}{0.33} & \se{67.13}{0.15} & \se{58.54}{1.04} & \se{53.17}{1.27} & \se{59.54}{0.57} & \se{74.15}{0.43} & 66.72 \\
    \quad \model\preg-8B & & & & & & & \\
    \qquad DDPM $256$ steps/block ($w{=}3,\tau{=}0.2$) & \se{88.61}{0.31} & \se{66.40}{0.17} & \se{63.54}{1.39} & \se{58.23}{1.02} & \se{63.28}{0.57} & \se{77.49}{0.50} & 69.59 \\
    \qquad DDIM $32$ steps/block ($w{=}4,\tau{=}0.4$) & \se{87.60}{0.25} & \se{66.04}{0.21} & \se{61.04}{1.04} & \se{55.67}{0.88} & \se{60.34}{0.37} & \se{74.34}{0.46} & 67.51 \\
    \quad \model-8B & & & & & & & \\
    \qquad DDPM $256$ steps/block ($w{=}3,\tau{=}0.1$) & \se{88.57}{0.12} & \se{66.72}{0.16} & \se{65.12}{1.22} & \se{60.61}{1.15} & \se{65.74}{0.44} & \se{80.19}{0.76} & \B{71.16} \\
    \qquad DDIM $32$ steps/block ($w{=}3,\tau{=}0.3$) & \se{88.48}{0.14} & \se{65.91}{0.15} & \se{\B{65.37}}{0.57} & \se{\B{60.91}}{0.74} & \se{62.68}{0.30} & \se{77.72}{0.32} & 70.18 \\
    \bottomrule
    \end{tabular}%
    }
\end{table}

\subsection{Ablation Study: MDLM Initialization for Pretraining}
\label{app:pretrain-init}
[Back to~\cref{sec:pretrain} in the main paper.]

All \model models featured in our main results are warm-started from an AR checkpoint (``AR initialization'' in \cref{tab:base-3b}) and then trained for $300$B tokens under the joint diffusion + AR objective (\cref{sec:pretrain-from-scratch}). We ablate that choice against a masked diffusion language model (MDLM) warm start: identical $300$B-token recipe, run instead from the released Nemotron-Labs-Diffusion-3B checkpoint of \citet{fu2026nemotron}, which is our main baseline in~\cref{tab:base-3b}. All other knobs are held fixed.

\begin{figure*}[h]
    \centering
    \begin{subfigure}{0.4\linewidth}
        \centering
        \includegraphics[width=\linewidth]{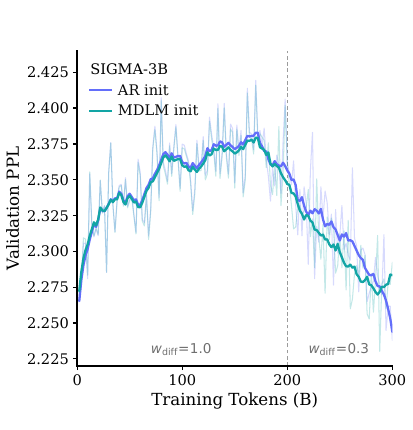}
        \caption{\centering Validation Loss (AR)}
        \label{fig:ar-vs-mdlm-init-ar}
    \end{subfigure}
    \begin{subfigure}{0.4\linewidth}
        \centering
        \includegraphics[width=\linewidth]{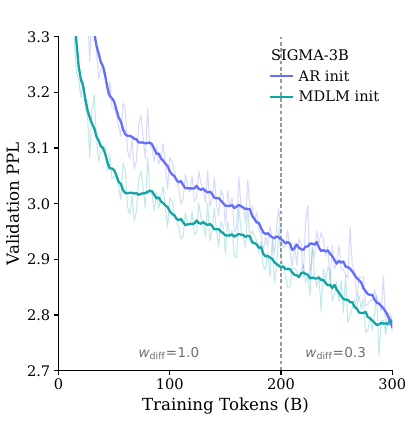}
        \caption{\centering Validation Loss (Diffusion)}
        \label{fig:ar-vs-mdlm-init-diff}
    \end{subfigure}
    \caption{ \textbf{Pretraining \model\unreg-3B} with block size 32 and AR regularization from AR initialization (AR init) and from MDLM initialization (MDLM init). PPL for AR stream (a) and diffusion stream (b) are computed using $\cL_{\text{ar}}(\x)$ \cref{eq:ar_loss} and $\cL_{\text{BD}}(\x)$~\cref{eq:block-nelbo}, respectively. MDLM initialization has an initial advantage that dies off during learning rate decay.}
    \label{fig:pretrain-init}
\end{figure*}

\paragraph{Evaluation setup.}
In \cref{tab:init-3b}, we evaluate the performance of the two trained \model\unreg-3B models at the two CFG--ST configurations of \cref{tab:base-3b}, both with SC enabled: DDPM ($256$ steps, EDM $\rho=7$ schedule, $w = 2$, $\tau = 0.1$) and DDIM ($32$ steps, uniform-$\gamma$ schedule, $w = 3$, $\tau = 0.3$). Numbers are mean $\pm$ two standard errors over $10$ sampling lanes, forming a $95\%$ confidence interval of the mean. %

\begin{table}[h]
    \centering
    \caption{ \textbf{Diffusion initialization ablation for \model\unreg-3B pretraining.} Same six generative reasoning tasks as \cref{tab:base-3b}. The \emph{AR init} rows are copied verbatim from \cref{tab:base-3b}. The \emph{MDLM init} rows use the same $300$B-token training recipe but warm-started from the Nemotron-Labs-Diffusion-3B checkpoint of \citet{fu2026nemotron}. $\Delta = $ MDLM $-$ AR. All metrics are mean $\pm$ two standard errors over $10$ sampling lanes. \textbf{Avg.} is the unweighted mean over the six tasks.}
    \label{tab:init-3b}
    \resizebox{\linewidth}{!}{%
    \begin{tabular}{lcccccccc}
    \toprule
    Sampler & Init & GSM8K & Minerva Math & HumanEval & HumanEval+ & MBPP & MBPP+ & Avg. \\
    \midrule
                                        & AR       & \se{84.50}{0.31} & \se{64.60}{0.28} & \se{59.45}{1.30} & \se{53.84}{1.29} & \se{52.32}{0.73} & \se{71.32}{0.91} & 64.34 \\
    DDPM ($256$ steps, $w{=}2,\tau{=}0.1$)  & MDLM     & \se{84.45}{0.35} & \se{65.37}{0.12} & \se{57.13}{1.05} & \se{52.87}{1.35} & \se{55.56}{0.65} & \se{72.62}{0.56} & 64.67 \\
                                        & $\Delta$ & -0.05          & +0.77          & -2.32          & -0.97          & +3.24          & +1.30          & +0.33 \\
    \midrule
                                        & AR       & \se{81.91}{0.32} & \se{62.76}{0.18} & \se{56.59}{1.31} & \se{50.67}{1.00} & \se{47.26}{0.46} & \se{63.49}{0.77} & 60.45 \\
    DDIM ($32$ steps, $w{=}3,\tau{=}0.3$)   & MDLM     & \se{83.71}{0.17} & \se{64.23}{0.28} & \se{52.99}{0.59} & \se{50.18}{0.68} & \se{51.24}{0.51} & \se{67.46}{0.58} & 61.64 \\
                                        & $\Delta$ & +1.80          & +1.47          & -3.60          & -0.49          & +3.98          & +3.97          & +1.19 \\
    \bottomrule
    \end{tabular}%
    }
\end{table}

\paragraph{Results.} MDLM initialization has a small but consistent win on average at both configurations, but the sign of the change is task-dependent rather than uniform. MBPP and MBPP+ strongly favor MDLM initialization, while HumanEval and HumanEval+ show opposite trends.

\clearpage

\section{Implementation Details and Additional Results for Likelihood-based Benchmarks}
\label{app:likelihood}
[Back to~\cref{sec:pretrain} in the main paper.]

\begin{table}[h!]
    \centering
    \caption{ \textbf{Likelihood-based multiple-choice benchmarks (3B).} Accuracy (\%) per task. \emph{AR mode} ranks candidates with the AR likelihood from the AR stream; \emph{diffusion mode} ranks them by the variational bound of the diffusion stream: \cref{eq:cond-nelbo} for \model, the count-based masked diffusion NELBO~\cref{eq:count-nelbo} for Nemotron-Labs-Diffusion. \underline{Underline}: the best within each mode. \textbf{Bold}: the best value across all methods.}
    \label{tab:likelihood-mc-3b}
    \resizebox{\linewidth}{!}{%
    \begin{tabular}{lccccccc}
    \toprule
    Method & ARC-e & ARC-c & HellaSwag & PIQA & WinoGrande & MMLU & Avg. \\
    \midrule
    \textit{Chance} & \textit{25.00} & \textit{25.00} & \textit{25.00} & \textit{50.00} & \textit{50.00} & \textit{25.00} & \textit{33.33} \\
    AR initialization & \textbf{80.39} & \textbf{57.42} & \textbf{71.74} & 78.18 & \textbf{70.64} & \textbf{70.74} & \textbf{71.52} \\
    \midrule
    \multicolumn{8}{l}{\emph{AR mode}} \\
    \quad Nemotron-Labs-Diffusion-3B & 78.70 & \underline{55.46} & \underline{70.39} & \underline{\textbf{78.24}} & \underline{70.32} & \underline{69.61} & \underline{70.46} \\
    \quad \textbf{\model\unreg-3B (Ours)} & \underline{79.29} & 55.29 & 69.98 & 77.64 & 69.46 & 68.50 & 70.03 \\
    \midrule
    \multicolumn{8}{l}{\emph{Diffusion mode}} \\
    \quad Nemotron-Labs-Diffusion-3B & 78.07 & 52.99 & 64.90 & \underline{73.78} & 68.75 & \underline{69.70} & 68.03 \\
    \quad \textbf{\model\unreg-3B (Ours)} & \underline{78.96} & \underline{54.61} & \underline{65.64} & 73.61 & \underline{69.93} & 67.66 & \underline{68.40} \\
    \bottomrule
    \end{tabular}%
    }
\end{table}

\begin{table}[h!]
    \centering
    \caption{ \textbf{Likelihood-based multiple-choice benchmarks (8B).} Accuracy (\%) per task. \emph{AR mode} ranks candidates with the AR likelihood from the AR stream; \emph{diffusion mode} ranks them by the variational bound of the diffusion stream: \cref{eq:cond-nelbo} for \model, the count-based masked diffusion NELBO~\cref{eq:count-nelbo} for Nemotron-Labs-Diffusion. \underline{Underline}: the best within each mode. \textbf{Bold}: the best value across all methods.}
    \label{tab:likelihood-mc-8b}
    \resizebox{\linewidth}{!}{%
    \begin{tabular}{lccccccc}
    \toprule
    Method & ARC-e & ARC-c & HellaSwag & PIQA & WinoGrande & MMLU & Avg. \\
    \midrule
    \textit{Chance} & \textit{25.00} & \textit{25.00} & \textit{25.00} & \textit{50.00} & \textit{50.00} & \textit{25.00} & \textit{33.33} \\
    AR initialization & \textbf{83.80} & \textbf{61.52} & \textbf{77.05} & 80.03 & \textbf{76.48} & \textbf{76.84} & \textbf{75.95} \\
    \midrule
    \multicolumn{8}{l}{\emph{AR mode}} \\
    \quad Nemotron-Labs-Diffusion-8B & \underline{82.32} & 59.73 & \underline{76.09} & 79.76 & 75.93 & \underline{74.88} & 74.78 \\
    \quad \textbf{\model\unreg-8B (Ours)} & 82.28 & \underline{60.67} & 75.70 & \underline{\textbf{80.41}} & \underline{\textbf{76.48}} & 74.58 & \underline{75.02} \\
    \midrule
    \multicolumn{8}{l}{\emph{Diffusion mode}} \\
    \quad Nemotron-Labs-Diffusion-8B & 80.81 & \underline{58.62} & \underline{71.47} & \underline{77.31} & 73.88 & \underline{76.01} & \underline{73.02} \\
    \quad \textbf{\model\unreg-8B (Ours)} & \underline{80.85} & 58.19 & 69.03 & 76.44 & \underline{75.22} & 73.96 & 72.28 \\
    \bottomrule
    \end{tabular}%
    }
\end{table}

\paragraph{Benchmarks.} We evaluate on ARC-e/c~\citep{clark2018think}, HellaSwag~\citep{zellers-etal-2019-hellaswag}, PIQA~\citep{bisk2020piqa}, WinoGrande~\citep{sakaguchi2020winogrande}, and MMLU~\citep{hendrycks2021measuring}.
These benchmarks score a candidate response $\r$ against a prompt $\p$ by comparing $\log p_\theta(\r \mid \p)$ across candidates. 
Per Open LLM Leaderboard convention, we report the plain likelihood on WinoGrande and MMLU, and the likelihood normalized by character length on the other four tasks which are length-varying.

\paragraph{Scoring our continuous method.} \model does not give likelihood in closed form, but a surrogate based on the NELBO: taking the prompt as the clean prefix and the candidate as the block to be denoised, and then applying \cref{eq:nelbo} verbatim with $\p$ as $\x^{(\bcolor{<b})}$ and $\r$ as $\x^{(\bcolor{b})}$:
\begin{align}\label{eq:cond-nelbo}
    - \log p_\theta(\r \mid \p) \leq 
    \cL_{\text{NELBO}}(\r; \p) &= \KL(q(\z_1 \mid \r) \,\|\, p(\z_1)) + \E_{q(\z_0 \mid \r)}\big[\tsum_{l=1}^{|\r|}
    - \log \langle \x_\theta^l(\z_0, 0; \p), \r^{l} \rangle\big] \nonumber\\
    & -\tfrac{1}{2}\ \E_{t\sim\unif[0,1]}\E_{q(\z_t \mid \r)}[\SNR'(t) \| \e_\theta(\z_t, t; \p) - \e \|^2].
\end{align}

In practice, we evaluate \cref{eq:cond-nelbo} with $128$ stratified samples
$t_k = \frac{k+u}{128}$, $k = 0, \dots, 127$, $u \sim \unif[0,1]$ and $8$ reconstruction samples.

\paragraph{Scoring the masked baseline via count-based NELBO.}
We first give a brief introduction of masked diffusion models~\citep{sahoo2024simple, shi2024simplified, ou2025your}. There, the latent $\z_t \in \cV^L$ is a partially masked token sequence: each position is independently kept clean with probability $1-t$ or replaced by the mask token $\mask$ with probability $t$. The denoiser $\x_\theta(\z_t) \in (\Delta^V)^L \subset \R^{L\times V}$ returns a categorical distribution over clean tokens at each position; one can prove that it needs not to take time argument as $t$ carries no information about $\x$ beyond what $\z_t$ already reveals~\citep{ou2025your}. The resulting NELBO is a $\frac{1}{t}$-weighted cross entropy over the masked positions only.

Let $\p$ be the prompt and $\r$ be a candidate response. We generate the masked states by the \emph{mask set}, instead of randomly masking each position with probability $t$: for $M \subseteq \{1, \dots, L_\r\}$, where $L_\r := |\r|$, we obtain $\z_M$ from masking indices in $M$ of $\r$, and denote
$C(M) := \sum_{l \in M} -\log \langle \x_\theta^l(\z_M; \p),\, \r^l \rangle$ their cross entropy. Conditioning that bound on $\p$ and integrating $t$ out yields an average over
the \emph{number} of masked tokens~\citep{ou2025your, nie2025large}:
\begin{align}
    - \log p_\theta(\r \mid \p) \;\leq\; \cL_{\text{MDLM}}(\r; \p)
    &\;=\; \E_{t \sim \unif[0,1]} \Big[\frac{1}{t}\,
    \E_{M \sim \mathrm{Ber}(t)^{\otimes L_\r}} C(M) \Big] \label{eq:cond-mdlm-nelbo} \\
    &\;=\; \E_{n \sim \unif\{1, \ldots, L_\r\}} \Big[ \frac{L_\r}{n}\,
    \E_{M \sim \unif\binom{[L_\r]}{n}} C(M) \Big], \label{eq:count-nelbo}
\end{align}
where $\binom{[L_\r]}{n}$ means the set of all $n$-element subsets of $[L_\r]:=\{1,2,...,L_\r\}$. \cref{eq:count-nelbo} is a Rao--Blackwellized low-variance estimator of \cref{eq:cond-mdlm-nelbo}, manifested in the following three major differences: \emph{(i)} The $\frac1t$ weight becomes $\frac{L_\r}{n} \leq L_\r$, so rare small-$t$ draws that leave one or two masked tokens would not cause the variance of the estimator to blow up. \emph{(ii)} Bernoulli masking leaves $\r$ wholly unmasked with
probability $(1-t)^{L_\r}$, which may be common for short responses and wastes a forward pass, while the $n \geq 1$ constraint avoids this. \emph{(iii)} As $n$ is finite, we can similarly use stratified sampling as the $t$-grid above, spreading counts evenly
over $\{1, \ldots, L_\r\}$ from one uniform offset, so every batch covers all masking rates.
In practice, we use \cref{eq:count-nelbo} with $128$ low-discrepancy draws $n_k$ and standard random sampling for $M_k$ for Nemotron-Labs-Diffusion~\citep{fu2026nemotron} in \cref{tab:likelihood-mc-3b,tab:likelihood-mc-8b}. On MMLU candidates are
single tokens, so
$L_\r = n = 1$ and \cref{eq:count-nelbo} is exact after one forward pass, while
\cref{eq:cond-mdlm-nelbo} would still be averaging $\frac1t$. We found \cref{eq:count-nelbo} to work better than  \cref{eq:cond-mdlm-nelbo} in early experiments.

\paragraph{Length generalization.} Diffusion training denoises blocks of exactly $L'$
tokens from packed sequences of length $L$, whereas candidates are
typically short and occasionally longer than $L'$. For each $(\p, \r)$ pair, we apply causal attention over $\p$ and bidirectional attention over $\r$, assuming that the bidirectional attention can generalize to response length other than $L'$. The attention map follows the conditional branch in \cref{fig:cfg-attention}.

\clearpage
\section{Implementation Details and Additional Results for Inference}
\label{app:inference}

[Back to~\cref{sec:inference,sec:few-step} in the main paper.]

 \begin{table}[h]
      \centering
      \caption{ \textbf{All-NFE pass@$1$ accuracy} of \model\unreg-3B, \model\preg-3B, and \model-3B.
      10 seed-runs with average and $\pm$ two standard errors reported. 
  \textbf{Bold} marks the best value per task among \model per-NFE rows and NLD;
  \underline{underline} marks the better of \model\unreg-3B, \model\preg-3B and \model-3B within each step budget.
  \small $\dagger$ we did not conduct distillation on \model\unreg-3B with the newest distillation recipe, which was historically developed on \model\preg-3B.
  }
      \label{tab:acc-nfe-3b}

            \resizebox{\linewidth}{!}{%
      \begin{tabular}{lcccccccc}
      \toprule
      & \multicolumn{2}{c}{\textbf{Math}} & \multicolumn{4}{c}{\textbf{Coding}} & & \\
      \cmidrule(lr){2-3} \cmidrule(lr){4-7}
      Method (Diffusion Mode) & GSM8K & Minerva Math & HumanEval & HumanEval+ & MBPP & MBPP+ & Avg. & $\Delta$ Avg. \\
      \midrule
      Nemotron-Labs-Diffusion-3B & 83.40 & 63.96 & 56.10 & 53.05 & \textbf{57.60} & \textbf{74.34} & 64.74 & \\
      \midrule
      \multicolumn{9}{l}{\textit{DDIM 256 steps / block}} \\
      \quad \model\unreg-3B ($w{=}3,\tau{=}0.2$) & \se{84.18}{0.18} & \se{64.00}{0.14} &
  \se{\textbf{\underline{60.24}}}{0.84} & \se{\underline{54.02}}{0.80} & \se{51.02}{0.42} & \se{69.79}{0.16} & 63.88 &
  \\
      \quad \model\preg-3B ($w{=}3,\tau{=}0.2$) & \se{84.12}{0.14} & \se{63.68}{0.18} & \se{57.32}{0.67} & \se{52.20}{0.74} &
  \se{56.08}{0.39} & \se{72.65}{0.31} & 64.34 & \cellcolor{gray!8}+0.46 \\
      \quad \model-3B ($w{=}3,\tau{=}0.1$) & \se{\underline{85.73}}{0.10} & \se{\underline{\textbf{64.74}}}{0.16} & \se{55.43}{0.53} & \se{52.87}{0.62} &
  \se{\underline{57.40}}{0.19} & \se{\underline{72.04}}{0.32} & \underline{64.70} & \cellcolor{gray!15}+0.82 \\
      \midrule
      \multicolumn{9}{l}{\textit{DDIM 128 steps / block}} \\
      \quad \model\unreg-3B ($w{=}3,\tau{=}0.2$) & \se{83.78}{0.11} & \se{64.01}{0.15} & \se{\underline{59.15}}{0.68} & \se{53.23}{0.91} &
  \se{50.92}{0.26} & \se{68.49}{0.37} & 63.26 & \\
      \quad \model\preg-3B ($w{=}3,\tau{=}0.1$) & \se{84.64}{0.16} & \se{63.50}{0.11} & \se{57.56}{0.39} & \se{51.77}{0.58} &
  \se{\underline{55.84}}{0.17} & \se{72.62}{0.38} & 64.32 & \cellcolor{gray!20}+1.06 \\
      \quad \model-3B ($w{=}3,\tau{=}0.2$) & \se{\underline{\textbf{85.84}}}{0.14} & \se{\underline{64.62}}{0.21} & \se{57.01}{0.50} & \se{\underline{54.15}}{0.69} &
  \se{\underline{56.64}}{0.32} & \se{\underline{72.94}}{0.27} & \underline{\textbf{65.20}} & \cellcolor{gray!25}+1.94 \\
      \midrule
      \multicolumn{9}{l}{\textit{DDIM 64 steps / block}} \\
      \quad \model\unreg-3B ($w{=}3,\tau{=}0.2$) & \se{83.09}{0.27} & \se{63.96}{0.16} & \se{56.83}{0.62} & \se{51.40}{0.60} &
  \se{49.38}{0.39} & \se{66.35}{0.35} & 61.84 & \\
      \quad \model\preg-3B ($w{=}4,\tau{=}0.2$) & \se{85.31}{0.15} & \se{63.68}{0.11} & \se{56.71}{0.55} & \se{51.95}{0.64} &
  \se{\underline{53.34}}{0.33} & \se{70.61}{0.43} & 63.60 & \cellcolor{gray!23}+1.76 \\
      \quad \model-3B ($w{=}3,\tau{=}0.2$) & \se{\underline{85.43}}{0.22} & \se{\underline{64.28}}{0.09} & \se{\underline{57.20}}{0.66} & \se{\underline{53.66}}{0.62} &
  \se{\underline{56.42}}{0.37} & \se{\underline{72.65}}{0.28} & \underline{64.94} & \cellcolor{gray!30}+3.10 \\
      \midrule
      \multicolumn{9}{l}{\textit{DDIM 32 steps / block}} \\
      \quad \model\unreg-3B ($w{=}3,\tau{=}0.3$) & \se{81.91}{0.32} & \se{62.76}{0.18} & \se{\underline{56.59}}{1.30} &
  \se{50.67}{1.00} & \se{47.06}{0.52} & \se{63.49}{0.77} & 60.41 & \\
      \quad \model\preg-3B ($w{=}4,\tau{=}0.4$) & \se{83.17}{0.30} & \se{63.28}{0.15} & \se{55.24}{1.36}
  & \se{50.91}{0.99} & \se{51.64}{0.57} & \se{66.88}{0.29} & 61.85 &
  \cellcolor{gray!20}+1.44 \\
      \quad \model-3B ($w{=}3,\tau{=}0.3$) & \se{\underline{84.48}}{0.21} & \se{\underline{64.52}}{0.21} & \se{55.37}{1.28}
  & \se{\underline{51.71}}{1.23} & \se{\underline{53.00}}{0.34} & \se{\underline{71.30}}{0.45} & \underline{63.40} & \cellcolor{gray!30}+2.99 \\
      \midrule
      \multicolumn{9}{l}{\textit{DDIM 16 steps / block}} \\
      \quad \model\unreg-3B ($w{=}3,\tau{=}0.5$) & \se{78.08}{0.34} & \se{60.16}{0.21} & \se{47.50}{1.14} &
  \se{43.84}{0.94} & \se{41.84}{1.00} & \se{59.58}{0.98} & 55.17 & \\
      \quad \model\preg-3B ($w{=}3,\tau{=}0.4$) & \se{80.85}{0.31} & \se{61.73}{0.24} &
  \se{\underline{53.60}}{0.80} & \se{48.84}{0.90} & \se{45.22}{0.57} & \se{60.79}{0.66} &
  58.50 & \cellcolor{gray!30}+3.33 \\
      \quad \model-3B ($w{=}3,\tau{=}0.4$) & \se{\underline{82.74}}{0.29} & \se{\underline{63.34}}{0.24} &
  \se{52.20}{1.45} & \se{\underline{49.57}}{1.41} & \se{\underline{48.28}}{0.30} & \se{\underline{66.98}}{0.74} & \underline{60.52} & \cellcolor{gray!40}+5.35 \\
\midrule
 \multicolumn{9}{l}{\textit{DDIM 8 steps / block}} \\
      \quad \model\unreg-3B ($w{=}3,\tau{=}0.8$) & \se{66.65}{0.40} & \se{45.13}{0.23} & \se{28.54}{1.53} &
    \se{25.67}{1.35} & \se{23.58}{0.65} & \se{32.80}{0.83} & 37.06 & \\
      \quad \model\preg-3B ($w{=}3,\tau{=}0.7$) & \se{70.88}{0.43} & \se{52.12}{0.40} &
    \se{34.45}{1.15} & \se{31.52}{1.29} & \se{27.98}{0.66} & \se{39.63}{0.99} &
    42.76 & \cellcolor{gray!40}+5.70 \\
      \quad \model-3B ($w{=}3,\tau{=}0.6$) & \se{\underline{76.33}}{0.45} & \se{\underline{57.95}}{0.18} &
    \se{\underline{43.96}}{1.20} & \se{\underline{40.55}}{1.12} & \se{\underline{34.54}}{0.64} & \se{\underline{51.53}}{0.86} & \underline{50.81} & \cellcolor{gray!60}+13.75 \\
\midrule
 \multicolumn{9}{l}{\textit{DDIM 4 steps / block}} \\
      \quad \model\unreg-3B ($w{=}2,\tau{=}0.9$) & \se{11.64}{0.57} & \se{4.94}{0.11} & \se{9.88}{0.69} &
    \se{9.15}{0.79} & \se{4.24}{0.53} & \se{6.59}{0.52} & 7.74 & \\
      \quad \model\preg-3B ($w{=}2,\tau{=}0.9$) & \se{26.04}{0.64} & \se{9.64}{0.20} &
    \se{11.59}{1.22} & \se{10.85}{1.09} & \se{4.32}{0.62} & \se{8.33}{0.84} &
    11.80 & \cellcolor{gray!35}+4.06 \\
      \quad \model-3B ($w{=}2,\tau{=}0.8$) & \se{\underline{38.52}}{0.40} & \se{\underline{20.67}}{0.21} &
    \se{\underline{21.89}}{1.27} & \se{\underline{20.24}}{1.25} & \se{\underline{11.32}}{0.53} & \se{\underline{19.39}}{0.81} & \underline{22.01} & \cellcolor{gray!65}+14.27 \\
    \midrule 
    
\multicolumn{9}{l}{\textit{Distilled, 16 steps / block}} \\
 \quad \model\distill\unreg-3B$^{\dagger}$
 & -- & -- & -- & -- & -- & -- & -- & \\
 \quad \model\distill\preg-3B
 & \se{83.12}{0.31} & \se{60.02}{0.24}
 & \se{57.32}{1.63} & \se{51.83}{1.57}
 & \se{49.40}{0.55} & \se{69.02}{0.87}
 & 61.78 &\\
 \quad \model\distill-3B
 & \se{\underline{83.42}}{0.27} & \se{\underline{61.70}}{0.21}
 & \se{\underline{58.54}}{1.15} & \se{\underline{\textbf{54.51}}}{1.48}
 & \se{\underline{52.08}}{0.87} & \se{\underline{69.95}}{0.69}
 & \underline{63.37} & \\
\midrule
\multicolumn{9}{l}{\textit{Distilled, 8 steps / block}} \\
 \quad \model\distill\unreg-3B$^{\dagger}$
 & -- & -- & -- & -- & -- & -- & -- & \\
 \quad \model\distill\preg-3B
  & \se{80.37}{0.33} & \se{58.22}{0.15}
 & \se{51.89}{1.49} & \se{47.20}{1.49}
 & \se{42.38}{0.84} & \se{59.79}{0.76}
 & 56.64 & \\
 \quad \model\distill-3B
 & \se{\underline{84.62}}{0.44} & \se{\underline{61.17}}{0.22}
 & \se{\underline{56.40}}{1.82} & \se{\underline{51.71}}{2.28}
 & \se{\underline{49.04}}{0.45} & \se{\underline{67.70}}{0.77}
 & \underline{61.77} & \\
\midrule
\multicolumn{9}{l}{\textit{Distilled, 4 steps / block}} \\
 \quad \model\distill\unreg-3B$^{\dagger}$
 & -- & -- & -- & -- & -- & -- & -- & \\
 \quad \model\distill\preg-3B
  & \se{76.94}{0.35} & \se{53.97}{0.27}
 & \se{36.77}{1.54} & \se{33.17}{1.70}
 & \se{23.70}{1.10} & \se{33.76}{0.69}
 & 43.05 &  \\
 \quad \model\distill-3B
 & \se{\underline{82.23}}{0.34} & \se{\underline{59.13}}{0.19}
 & \se{\underline{52.13}}{1.56} & \se{\underline{48.72}}{1.20}
 & \se{\underline{40.62}}{0.91} & \se{\underline{58.81}}{1.22}
 & \underline{56.94} & \\
 \midrule
\multicolumn{9}{l}{\textit{Distilled, 2 steps / block}} \\
\quad \model\distill\unreg-3B$^{\dagger}$
& -- & -- & -- & -- & -- & -- & -- & \\
\quad \model\distill\preg-3B 
& -- & -- & -- & -- & -- & -- & -- & \\
\quad \model\distill-3B
 & \se{\underline{74.42}}{0.46} & \se{\underline{49.71}}{0.30}
 & \se{\underline{32.38}}{1.34} & \se{\underline{30.55}}{0.86}
 & \se{\underline{20.78}}{1.00} & \se{\underline{30.87}}{0.79}
 & \underline{39.78} & \\
      \bottomrule

      \end{tabular}%
      }
  \end{table}

 \begin{table}[ht]
      \centering
      \caption{ \textbf{All-NFE pass@$1$ accuracy} of \model\unreg-3B, \model\preg-3B, and \model-3B.
      10 seed-runs with average and $\pm$ two standard errors reported. 
  \textbf{Bold} marks the best value per task among \model per-NFE rows and NLD;
  \underline{underline} marks the better of \model\unreg-3B, \model\preg-3B and \model-3B within each step budget.
  ;
  \underline{underline} marks the better of \model\unreg-8B, \model\preg-8B and \model-8B within each step budget, matching \cref{tab:acc-nfe-3b}.
  }
      \label{tab:acc-nfe-8b}
      \resizebox{\linewidth}{!}{%

\begin{tabular}{lcccccccc}
\toprule
 & \multicolumn{2}{c}{\textbf{Math}} & \multicolumn{4}{c}{\textbf{Coding}} & & \\
\cmidrule(lr){2-3} \cmidrule(lr){4-7}
Method (Diffusion Mode) & GSM8K & Minerva Math & HumanEval & HumanEval+ & MBPP & MBPP+ & Avg. & $\Delta$ Avg. \\
\midrule
Nemotron-Labs-Diffusion-8B
 & 87.26 & 65.16
 & 62.80 & 57.32
 & \textbf{67.00} & \textbf{81.75}
 & 70.22 &  \\
\midrule
\multicolumn{9}{l}{\textit{DDIM 256 steps / block}} \\
\quad \model\unreg-8B ($w{=}3,\tau{=}0.2$)
 & \se{88.41}{0.14} & \se{\textbf{\underline{67.73}}}{0.10}
 & \se{58.17}{0.80} & \se{52.87}{0.81}
 & \se{63.82}{0.26} & \se{76.85}{0.21}
 & 67.97 &  \\
\quad \model\preg-8B ($w{=}3,\tau{=}0.2$)
 & \se{88.38}{0.11} & \se{66.88}{0.11}
 & \se{64.63}{0.30} & \se{58.48}{0.40}
 & \se{62.58}{0.24} & \se{78.31}{0.24}
 & 69.88 & \cellcolor{gray!20}+1.91 \\
\quad \model-8B ($w{=}3,\tau{=}0.1$)
 & \se{\underline{89.12}}{0.12} & \se{67.06}{0.17}
 & \se{\underline{67.01}}{0.50} & \se{\underline{62.56}}{0.39}
 & \se{\underline{65.48}}{0.18} & \se{\underline{80.26}}{0.19}
 & \textbf{\underline{71.92}} & \cellcolor{gray!30}+3.95 \\
\midrule
\multicolumn{9}{l}{\textit{DDIM 128 steps / block}} \\
\quad \model\unreg-8B ($w{=}3,\tau{=}0.2$)
 & \se{\underline{88.70}}{0.14} & \se{\underline{67.28}}{0.09}
 & \se{58.48}{1.00} & \se{52.68}{1.00}
 & \se{63.22}{0.28} & \se{76.85}{0.31}
 & 67.87 &  \\
\quad \model\preg-8B ($w{=}3,\tau{=}0.1$)
 & \se{88.28}{0.12} & \se{66.36}{0.11}
 & \se{\underline{64.15}}{0.86} & \se{58.05}{0.82}
 & \se{62.22}{0.21} & \se{78.02}{0.22}
 & 69.51 & \cellcolor{gray!20}+1.64 \\
\quad \model-8B ($w{=}3,\tau{=}0.2$)
 & \se{88.27}{0.14} & \se{66.86}{0.08}
 & \se{63.72}{0.95} & \se{\underline{59.70}}{0.98}
 & \se{\underline{64.74}}{0.28} & \se{\underline{79.66}}{0.32}
 & \underline{70.49} & \cellcolor{gray!25}+2.62 \\
\midrule
\multicolumn{9}{l}{\textit{DDIM 64 steps / block}} \\
\quad \model\unreg-8B ($w{=}3,\tau{=}0.2$)
 & \se{\underline{88.68}}{0.19} & \se{\underline{67.30}}{0.17}
 & \se{58.90}{0.71} & \se{52.56}{0.54}
 & \se{61.96}{0.36} & \se{76.35}{0.28}
 & 67.63 &  \\
\quad \model\preg-8B ($w{=}4,\tau{=}0.2$)
 & \se{88.66}{0.14} & \se{66.40}{0.16}
 & \se{63.54}{0.93} & \se{56.65}{0.63}
 & \se{63.02}{0.25} & \se{76.46}{0.24}
 & 69.12 & \cellcolor{gray!20}+1.49 \\
\quad \model-8B ($w{=}3,\tau{=}0.2$)
 & \se{88.12}{0.15} & \se{66.35}{0.12}
 & \se{\underline{63.60}}{0.60} & \se{\underline{59.94}}{0.52}
 & \se{\underline{64.50}}{0.28} & \se{\underline{80.08}}{0.24}
 & \underline{70.43} & \cellcolor{gray!25}+2.80 \\
\midrule
\multicolumn{9}{l}{\textit{DDIM 32 steps / block}} \\
\quad \model\unreg-8B ($w{=}3,\tau{=}0.3$)
 & \se{87.78}{0.33} & \se{\underline{67.13}}{0.15}
 & \se{58.54}{1.04} & \se{53.17}{1.27}
 & \se{59.54}{0.57} & \se{74.15}{0.43}
 & 66.72 &  \\
\quad \model\preg-8B ($w{=}4,\tau{=}0.4$)
 & \se{87.60}{0.25} & \se{66.04}{0.21}
 & \se{61.04}{1.04} & \se{55.67}{0.88}
 & \se{60.34}{0.37} & \se{74.34}{0.46}
 & 67.51 & \cellcolor{gray!10}+0.79 \\
\quad \model-8B ($w{=}3,\tau{=}0.3$)
 & \se{\underline{88.48}}{0.14} & \se{65.91}{0.15}
 & \se{\underline{65.37}}{0.57} & \se{\underline{60.91}}{0.74}
 & \se{\underline{62.68}}{0.30} & \se{\underline{77.72}}{0.32}
 & \underline{70.18} & \cellcolor{gray!30}+3.46 \\
\midrule
\multicolumn{9}{l}{\textit{DDIM 16 steps / block}} \\
\quad \model\unreg-8B ($w{=}3,\tau{=}0.5$)
 & \se{85.22}{0.33} & \se{\underline{65.90}}{0.20}
 & \se{54.27}{0.96} & \se{48.96}{1.12}
 & \se{55.06}{0.79} & \se{69.47}{0.69}
 & 63.15 &  \\
\quad \model\preg-8B ($w{=}3,\tau{=}0.4$)
 & \se{86.57}{0.28} & \se{65.40}{0.18}
 & \se{55.55}{1.41} & \se{50.00}{1.29}
 & \se{56.68}{0.57} & \se{69.42}{0.52}
 & 63.94 & \cellcolor{gray!10}+0.79 \\
\quad \model-8B ($w{=}3,\tau{=}0.4$)
 & \se{\underline{87.03}}{0.20} & \se{65.66}{0.21}
 & \se{\underline{63.90}}{1.36} & \se{\underline{59.33}}{1.28}
 & \se{\underline{60.16}}{0.41} & \se{\underline{75.61}}{0.53}
 & \underline{68.61} & \cellcolor{gray!40}+5.46 \\
\midrule
\multicolumn{9}{l}{\textit{DDIM 8 steps / block}} \\
\quad \model\unreg-8B ($w{=}3,\tau{=}0.8$)
 & \se{76.49}{0.42} & \se{56.79}{0.23}
 & \se{31.65}{1.74} & \se{29.02}{1.50}
 & \se{35.80}{0.94} & \se{46.53}{0.93}
 & 46.05 &  \\
\quad \model\preg-8B ($w{=}3,\tau{=}0.7$)
 & \se{80.40}{0.50} & \se{59.90}{0.24}
 & \se{40.98}{0.94} & \se{38.05}{1.26}
 & \se{40.24}{0.78} & \se{52.51}{1.08}
 & 52.01 & \cellcolor{gray!40}+5.96 \\
\quad \model-8B ($w{=}3,\tau{=}0.6$)
 & \se{\underline{83.95}}{0.36} & \se{\underline{62.94}}{0.22}
 & \se{\underline{53.48}}{1.23} & \se{\underline{49.63}}{1.41}
 & \se{\underline{52.58}}{0.88} & \se{\underline{66.61}}{0.79}
 & \underline{61.53} & \cellcolor{gray!60}+15.48 \\
\midrule
\multicolumn{9}{l}{\textit{DDIM 4 steps / block}} \\
\quad \model\unreg-8B ($w{=}2,\tau{=}0.9$)
 & \se{18.24}{0.38} & \se{10.65}{0.19}
 & \se{11.16}{0.48} & \se{11.04}{0.76}
 & \se{7.46}{0.38} & \se{11.46}{0.40}
 & 11.67 &  \\
\quad \model\preg-8B ($w{=}2,\tau{=}0.9$)
 & \se{37.85}{0.47} & \se{17.06}{0.24}
 & \se{14.94}{1.15} & \se{14.39}{1.04}
 & \se{8.88}{0.62} & \se{12.99}{0.57}
 & 17.69 & \cellcolor{gray!50}+6.02 \\
\quad \model-8B ($w{=}2,\tau{=}0.8$)
 & \se{\underline{56.29}}{0.39} & \se{\underline{32.43}}{0.24}
 & \se{\underline{31.40}}{1.13} & \se{\underline{28.41}}{0.83}
 & \se{\underline{26.84}}{0.87} & \se{\underline{38.57}}{0.69}
 & \underline{35.66} & \cellcolor{gray!70}+23.99 \\
\midrule
\multicolumn{9}{l}{\textit{Distilled, 16 steps / block}} \\
\quad \model\distill\unreg-8B
 & \se{87.21}{0.21} & \se{\underline{66.46}}{0.21}
 & \se{62.74}{0.80} & \se{59.39}{0.73}
 & \se{61.26}{0.53} & \se{75.29}{1.00}
 & 68.73 & \\
\quad \model\distill\preg-8B
 & \se{85.78}{0.48} & \se{63.80}{0.18}
 & \se{62.93}{0.77} & \se{58.72}{0.66}
 & \se{61.02}{0.65} & \se{77.28}{0.46}
 & 68.25 & \\
\quad \model\distill-8B
 & \se{\textbf{\underline{89.15}}}{0.28} & \se{65.51}{0.23}
 & \se{\textbf{\underline{67.32}}}{1.14} & \se{\textbf{\underline{63.35}}}{1.18}
 & \se{\underline{61.96}}{0.81} & \se{\underline{77.70}}{0.96}
 & \underline{70.83} & \\
\midrule
\multicolumn{9}{l}{\textit{Distilled, 8 steps / block}} \\
\quad \model\distill\unreg-8B
 & \se{87.37}{0.22} & \se{64.98}{0.24}
 & \se{\underline{63.84}}{1.16} & \se{\underline{60.49}}{1.37}
 & \se{\underline{60.80}}{0.67} & \se{74.81}{0.41}
 & \underline{68.71} & \\
\quad \model\distill\preg-8B
 & \se{86.62}{0.27} & \se{63.51}{0.29}
 & \se{60.91}{0.97} & \se{57.20}{1.56}
 & \se{59.34}{0.78} & \se{\underline{74.95}}{0.66}
 & 67.09 & \\
\quad \model\distill-8B
 & \se{\underline{87.85}}{0.18} & \se{\underline{65.36}}{0.25}
 & \se{63.60}{1.74} & \se{59.63}{2.09}
 & \se{58.18}{0.91} & \se{72.75}{1.36}
 & 67.89 & \\
\midrule
\multicolumn{9}{l}{\textit{Distilled, 4 steps / block}} \\
\quad \model\distill\unreg-8B
 & \se{85.82}{0.28} & \se{62.53}{0.27}
 & \se{51.89}{1.32} & \se{49.15}{1.06}
 & \se{47.92}{0.78} & \se{62.78}{0.94}
 & 60.01 & \\
\quad \model\distill\preg-8B
 & \se{86.90}{0.36} & \se{63.20}{0.19}
 & \se{51.40}{0.93} & \se{48.48}{0.84}
 & \se{46.66}{0.87} & \se{59.10}{0.67}
 & 59.29 & \\
\quad \model\distill-8B
 & \se{\underline{87.15}}{0.24} & \se{\underline{65.05}}{0.29}
 & \se{\underline{60.12}}{0.86} & \se{\underline{57.01}}{0.71}
 & \se{\underline{55.06}}{0.97} & \se{\underline{72.62}}{0.78}
 & \underline{66.17} & \\
 \midrule
\multicolumn{9}{l}{\textit{Distilled, 2 steps / block}} \\
\quad \model\distill\unreg-8B 
& -- & -- & -- & -- & -- & -- & -- & \\
\quad \model\distill\preg-8B 
& -- & -- & -- & -- & -- & -- & -- & \\
\quad \model\distill-8B
 & \se{\underline{80.02}}{0.39} & \se{\underline{56.34}}{0.33}
 & \se{\underline{44.15}}{1.40} & \se{\underline{41.52}}{1.57}
 & \se{\underline{33.18}}{1.24} & \se{\underline{47.35}}{1.54}
 & \underline{50.43} & \\
 
\bottomrule
\end{tabular}
      }
  \end{table}

\clearpage

\subsection{Sampler Updates}\label{sec:app-samplers}
Let $1=t_0>\cdots>t_N=0$, with
$\alpha_i=\alpha_{t_i}$, $\sigma_i=\sigma_{t_i}$,
$\rho_i=\sigma_i/\alpha_i$, and
$\lambda_i=\log(\alpha_i/\sigma_i)=-\gamma(t_i)/2$.
At step $i$, the clean prediction
$\hat\e_i=\e_\theta(\z_i,t_i;\c)$ implies
$\hat\beps_i=(\z_i-\alpha_i\hat\e_i)/\sigma_i$.
Predictions incorporate CFG and ST as in \cref{eq:cfg,eq:st}.

\paragraph{DDPM.}
The ancestral VDM update~\citep[App.~A.4]{kingma2021variational}
uses the Gaussian posterior with $\hat\e_i$ replacing the clean
embedding:
\begin{equation}\label{eq:ddpm}
\z_{i+1}
=(1-c_i)\frac{\alpha_{i+1}}{\alpha_i}\z_i
+c_i\alpha_{i+1}\hat\e_i
+\sigma_{i+1}\sqrt{c_i}\,\beps,
\quad
c_i=1-\ee^{\gamma(t_{i+1})-\gamma(t_i)},
\quad
\beps\sim\mathcal N(\zero,\I).
\end{equation}

\paragraph{DDIM.}
DDIM~\citep{song2021denoising} deterministically combines the
current clean and noise predictions:
\begin{equation}\label{eq:ddim}
\z_{i+1}^{\mathrm{DDIM}}
=\alpha_{i+1}\hat\e_i+\sigma_{i+1}\hat\beps_i.
\end{equation}
It is the first-order update obtained by freezing either prediction
in the corresponding integrating-factor formulation of the PFODE.

\subsection{Ablation Study: Time Discretization}
\label{app:inference-time-disc}

[Back to~\cref{sec:few-step-ddim} in the main paper.]

\begin{figure*}[h]
    \centering
    \includegraphics[width=0.35\textwidth]{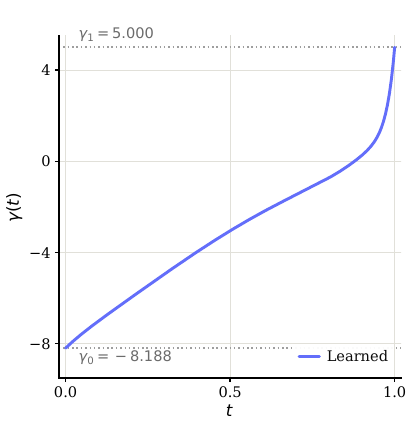}
    \includegraphics[width=0.35\textwidth]{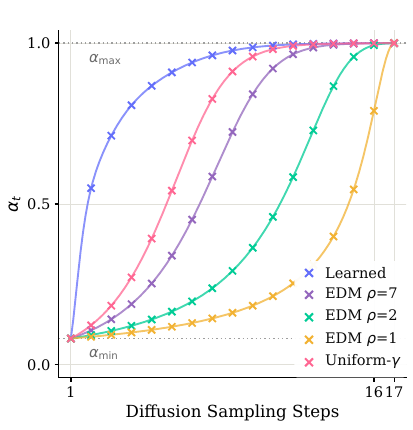}
    \caption{Left: the \textbf{learned noise schedule} $\gamma(t)$ for \model\unreg-3B. Right: Different \textbf{time discretization schemes} for 16-step generation. Note that an extra decoding step is needed at $t=0$.}
    \label{fig:few-step-schedule}
\end{figure*}

\begin{figure*}[h]
    \centering
    \includegraphics[width=0.75\linewidth]{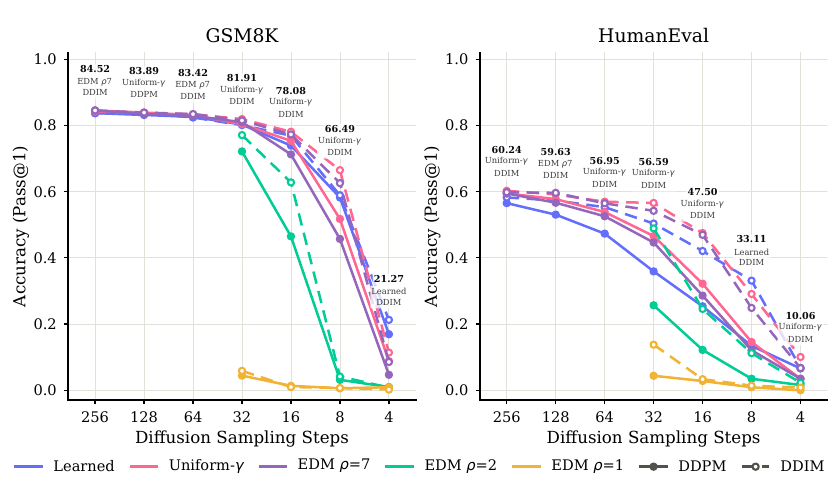}
    \caption{ \textbf{DDIM substantially outperforms DDPM at low NFEs} for \model\unreg-3B, while matching DDPM at high NFEs.
    }
    \label{fig:few-step-sweep}
\end{figure*}

We mainly consider the following discretization schemes. Let $N$ be the total number of steps and assume the $i$-th step ($i=1,...,N$) goes from $t_{i-1}$ to $t_i<t_{i-1}$.
\begin{itemize}
    \item Learned: arithmetic progression in time: $t_i = 1-\frac{i}{N}$.
    \item Uniform-$\gamma$: arithmetic progression in $\gamma$, the minus log-SNR: $\gamma(t_i) = \gamma_{\max} + \frac{i}{N}(\gamma_{\min} - \gamma_{\max})$.
    \item EDM $\rho$ \citep{karras2022elucidating}: first convert the VP schedule to VE by defining $\tilde\sigma_t := \frac{\sigma_t}{\alpha_t}$ as the essential standard deviation at time $t$. The grid points are chosen so that $\tilde\sigma_{t_i}^{1/\rho}$ is an arithmetic progression, i.e., $\tilde\sigma_{t_i} = \left(\tilde\sigma_{\max}^{1/\rho} + \frac{i}{N} (\tilde\sigma_{\min}^{1/\rho} - \tilde\sigma_{\max}^{1/\rho})\right)^{\rho}$.
    Note that as $\rho\to\infty$, this converges to the uniform-$\gamma$ scheme.
\end{itemize}
We visualize the three schemes in \cref{fig:few-step-schedule}. Empirically, we find that for high NFE sampling, both the uniform-$\gamma$ and EDM $\rho=7$ schemes perform similarly, while uniform-$\gamma$ generally performs the best for low NFE sampling. See \cref{fig:few-step-sweep} for the few-step reasoning accuracy of DDPM and DDIM over GSM8K and HumanEval.

\subsection{Reasoning and Coding Benchmark Setup}
\label{app:inference-eval-setup}

[Back to~\cref{sec:inference,sec:few-step} in the main paper.]

We describe the harness, scoring pipeline, and sampler that produce \cref{tab:base-3b,tab:base-8b,tab:acc-nfe-3b,tab:acc-nfe-8b}. The scoring pipeline is consistent for all base models and generation modes.

\paragraph{Benchmarks and few-shot budgets.} We use six tasks from \texttt{lm-evaluation-harness} following \citet{fu2026nemotron}:
GSM8K (8-shot, 1319 questions) \citep{cobbe2021gsm8k}, Minerva Math (4-shot, 5000 questions\footnote{Number of questions in each sub-domain: algebra 1187, counting/prob 474, geometry 479, intermediate algebra 903, number theory 540, prealgebra 871, precalc 546.}) \citep{lewkowycz2022solving}, HumanEval and HumanEval+ (0-shot, 164 questions) \citep{chen2021evaluating}, and MBPP and MBPP+ (3-shot, 500 and 378 questions, respectively) \citep{austin2021program}.
The per-task generation budget is 256 tokens for GSM8K and 512 tokens for the other five tasks.

\paragraph{Scoring.}
GSM8K is scored by exact-match on the numerical answer, extracted from the model's response by a permissive regex that keeps the last number appearing in the completion (stripping separators and trailing punctuation).
Minerva Math is scored by symbolic equivalence after canonicalizing boxed and inline expressions.
HumanEval and HumanEval+ report post-processed pass@$1$: the completion is stripped out of the Markdown code fence, an AST-based longest-valid-code extraction is applied, and the resulting program is executed against the task test cases in the container.
MBPP and MBPP+ report the harness pass@$1$ directly without this post-processing step, matching the convention of prior work.

\paragraph{Decoding configuration.}
All \model diffusion-mode runs use block size $L' = 32$, \texttt{BF16} model weights at inference time, and operation of SC follows \cref{fig:cfg-steering}. The DDPM-256 row uses the ancestral sampler of~\cref{eq:ddpm} on the EDM $\rho = 7$ time discretization (\cref{app:inference-time-disc}); the DDIM-32 row uses the deterministic sampler of~\cref{eq:ddim} on the uniform-$\gamma$ time discretization.

\paragraph{Seed error bars.} Each \model diffusion row aggregates $10$ independent sampling lanes as follows. For question $i$, we independently draw $10$ responses under the same model with $10$ different seeds. Let $s_{i,j}$ be the score of the $j$-th response to question $i$. We first average over questions to obtain $10$ per-lane scores $s_j = \operatorname{mean}_i(s_{i,j})$ for $j = 1, \dots, 10$, and then obtain $\bar s = \operatorname{mean}_j(s_j)$ together with the across-lane standard deviation $\hat\sigma = \operatorname{std}_j(s_j)$. The half-widths in \cref{tab:base-3b} are $\frac{2}{\sqrt{10}} \hat\sigma$, an approximate $95\%$ confidence interval for the mean.

\paragraph{Baseline decoders.}
The AR-mode for NLD and \model row uses greedy argmax at each step; NLD's diffusion-mode row uses the confidence-threshold greedy sampler of \citet{fu2026nemotron} (block size $L' = 32$; $\argmax$ readout at every position whose top-1 probability exceeds a threshold; we report the better of thresholds $0.8$ and $0.9$ per task). For LLaDA-8B and Dream-7B we quote each model's best diffusion-mode number without the confidence-threshold rule: at each denoising step the single position with the highest top-1 confidence is unmasked.
Both rules are instances of confidence-based decoding~\citep{chang2022maskgit}, and both inherit the absorbing property of masked diffusion: once a position is unmasked, its token is frozen for the rest of the trajectory.

\subsection{Sweeping Classifier-Free Guidance and Score Temperature}

[Back to~\cref{sec:inference-cfg-st} or~\cref{subsec:3b-base-math-code} in the main paper.]

\begin{figure}[p]
    \centering
    \includegraphics[width=0.86\linewidth]{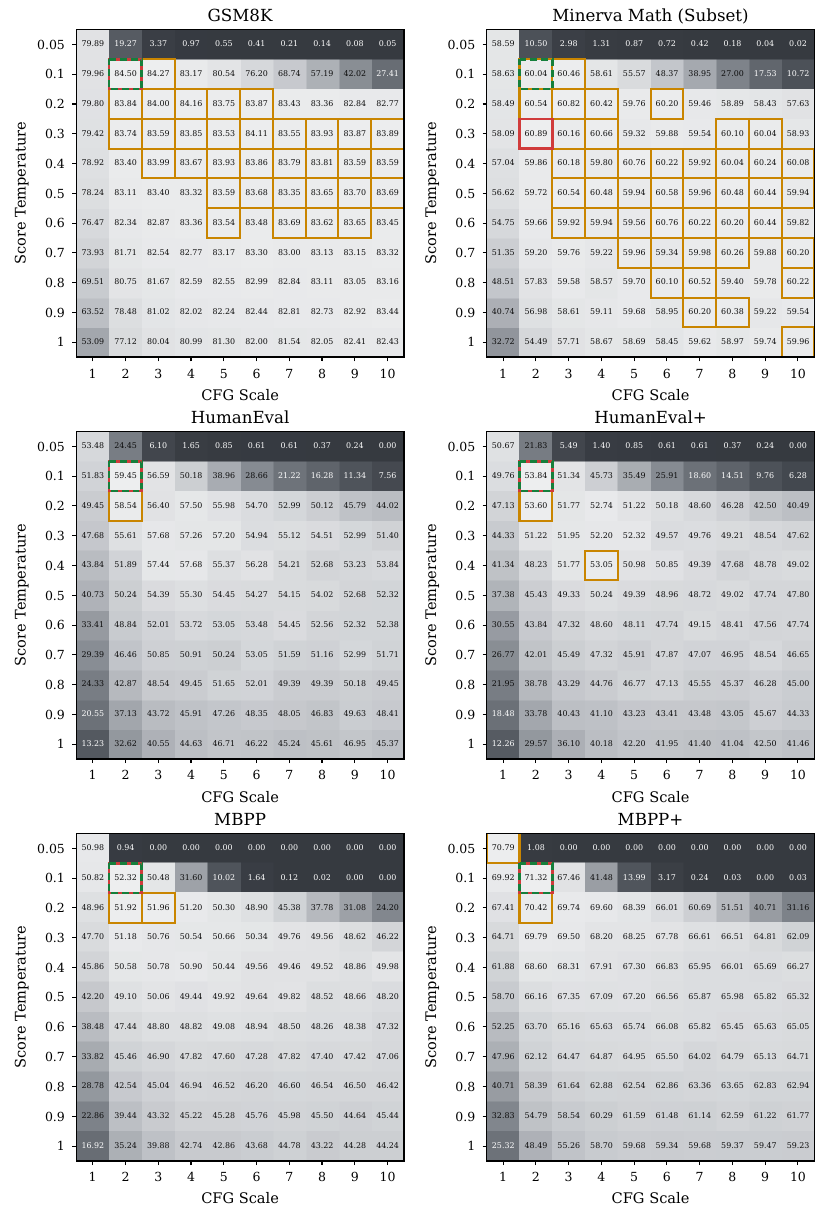}
    \caption{Sweeping CFG and ST for \model\unreg-3B trained with AR regularization. Accuracies are averaged over 10 random seeds. {\color{red} Red}: per-task best; {\color{orange}orange}: within 1 point of per-task best; {\color{green!50!black}green}: best average across 6 tasks. \textbf{For this model, per-task best and best average align on 5 out of 6 tasks.} In the exception of Minerva Math (subset; random $10\%$ from each subtask), per-task best is still within 1 point from best average. We report best average ($w=2$, $\tau=0.1$) in~\cref{fig:cfg-st-tuning} and~\cref{tab:base-3b}.}
    \label{fig:ddpm-256-sweep}
\end{figure}

\begin{figure}[p]
    \centering
    \includegraphics[width=0.9\linewidth]{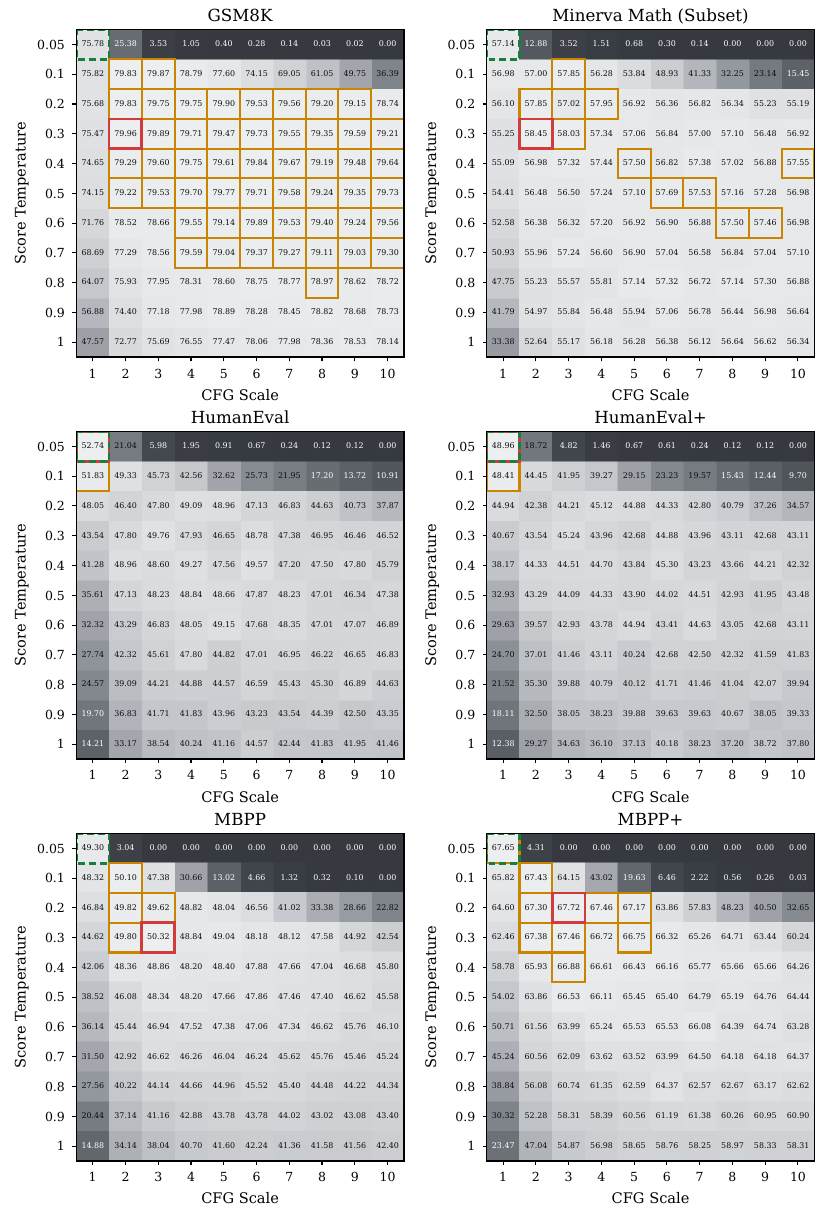}
    \caption{Sweeping CFG and ST for \model\unreg-3B trained \textit{without} AR regularization. Accuracies are averaged over 10 random seeds. {\color{red} Red}: per-task best; {\color{orange}orange}: within 1 point of per-task best; {\color{green!50!black}green}: best average across 6 tasks. Both per-task best and best average are worse than \model\unreg-3B trained AR regularization (\cref{fig:ddpm-256-sweep}).}
    \label{fig:ddpm-256-sweep-no-ar-loss}
\end{figure}

In \cref{fig:ddpm-256-sweep,fig:ddpm-256-sweep-no-ar-loss}, we present the full tables for CFG and ST sweep for our \model\unreg-3B models trained with and without AR regularization, respectively.

\subsection{Ablation Study: Auto-Guidance with Self-Conditioning}
\label{app:inference-cfg-bad-pred}
[Back to~\cref{sec:inference-cfg-st} in the main paper.]

Standard CFG~\citep{ho2022classifierfreediffusionguidance}, also following the idea of auto-guidance \citep{karras2024guiding}, extrapolates a conditional prediction away from a ``bad'' prediction to sharpen the model's prediction. Self-conditioning introduces a second source of conditioning, as is used in existing continuous dLMs (e.g., \citet{lee2026flow,potaptchik2026discrete,hu2026elf}; see \cref{app:related-work} for detailed discussion).
We define the following notations in this subsection:
\begin{align*}
    \e^{+}_\theta
    := \e_\theta(\z_t^{(\bcolor{b})},t;\c,\e_{\selfcond}),\qquad
    \e^{\c,\zero}_\theta
    := \e_\theta(\z_t^{(\bcolor{b})},t;\c,\zero),\qquad
    \e^{\varnothing,\zero}_\theta
    := \e_\theta(\z_t^{(\bcolor{b})},t;\varnothing,\zero),
\end{align*}
where the good prediction $\e^{+}_\theta$ is the conditional prediction that uses both the clean context and SC; for either bad prediction $\e^{-}_\theta \in \{\e^{\c,\zero}_\theta, \e^{\varnothing,\zero}_\theta\}$, the guided mean embedding prediction is
\begin{align*}
    \e^w_\theta
    = \e^{+}_\theta + (w-1)(\e^{+}_\theta-\e^{-}_\theta).
\end{align*}
The \textit{context-preserving} baseline keeps the prompt and previously generated blocks in both branches and sets $\e^{-}_\theta\gets\e^{\c,\zero}_\theta$; it therefore extrapolates only away from the prediction made without SC.
Our proposed method in \cref{sec:inference-cfg-st} instead uses the fully unconditional branch $\e^{-}_\theta\gets\e^{\varnothing,\zero}_\theta$, removing both clean context and SC.

\paragraph{Experimental setup.}
We compare the two choices using the same \model\unreg-3B checkpoint with block length $32$, full GSM8K and HumanEval evaluation sets, and 10 samples per problem. The DDIM sweep uses $64$ steps per block, uniform-$\gamma$ schedule, and ST $\tau=0.3$; the DDPM sweep uses $256$ steps per block, EDM $\rho=7$ schedule, and ST $\tau=1$. We remark that the guidance is inactive at $w=1$ and the two choices coincide.

\begin{table*}[h!]
    \centering
    \caption{ GSM8K accuracy (\%) for self-conditioning auto-guidance.
    ``Context/no-SC'' retains clean context but removes SC (i.e., $\e^{-}_\theta\gets\e^{\c,\zero}_\theta$); ours removes both (i.e., $\e^{-}_\theta\gets\e^{\varnothing,\zero}_\theta$). Score entries are mean $\pm$ two standard errors over the 10 sampling lanes, an approximate $95\%$ confidence interval for the mean; $\Delta$ is ours $-$ context/no-SC.}
    \label{tab:cfg-bad-pred-gsm8k}
    \setlength{\tabcolsep}{4.0pt}
    DDIM, 64 steps/block, uniform-$\gamma$ schedule, $\tau=0.3$\\
    \resizebox{\linewidth}{!}{%
    \begin{tabular}{cccccccccc}
        \toprule
        Guidance $w$ & 1 & 2 & 3 & 4 & 5 & 6 & 8 & 12 & 16 \\
        \midrule
        Context/no-SC & \se{68.13}{0.38} & \se{72.48}{0.27} & \se{73.99}{0.47} & \se{74.37}{0.35} & \se{75.12}{0.51} & \se{76.13}{0.41} & \se{\textbf{76.19}}{0.44} & \se{75.60}{0.56} & \se{74.09}{0.52} \\
        Ours    & \se{68.13}{0.38} & \se{82.52}{0.21} & \se{82.84}{0.28} & \se{83.04}{0.20} & \se{\textbf{83.19}}{0.18} & \se{83.01}{0.30} & \se{82.78}{0.40} & \se{81.26}{0.40} & \se{78.00}{0.32} \\
        $\Delta$ & 0.00 & +10.05 & +8.85 & +8.67 & +8.07 & +6.88 & +6.59 & +5.66 & +3.91 \\
        \bottomrule
    \end{tabular}
    }

    \vspace{0.5em}
    DDPM, 256 steps/block, EDM $\rho=7$ schedule, $\tau=1$\\
    \resizebox{\linewidth}{!}{%
    \begin{tabular}{ccccccccccc}
        \toprule
        Guidance $w$ & 1 & 2 & 3 & 4 & 5 & 6 & 7 & 8 & 9 & 10 \\
        \midrule
        Context/no-SC & \se{53.09}{0.85} & \se{62.43}{0.97} & \se{65.44}{0.77} & \se{67.53}{0.58} & \se{\textbf{68.45}}{0.77} & \se{\textbf{68.45}}{0.56} & \se{67.66}{0.61} & \se{67.82}{1.17} & \se{67.09}{0.48} & \se{66.76}{0.56} \\
        Ours    & \se{53.09}{0.85} & \se{77.12}{0.66} & \se{80.04}{0.39} & \se{80.99}{0.48} & \se{81.30}{0.41} & \se{82.00}{0.47} & \se{81.54}{0.64} & \se{82.05}{0.29} & \se{82.41}{0.51} & \se{\textbf{82.43}}{0.50} \\
        $\Delta$ & 0.00 & +14.69 & +14.60 & +13.46 & +12.85 & +13.55 & +13.88 & +14.23 & +15.32 & +15.67 \\
        \bottomrule
    \end{tabular}
    }
\end{table*}

\begin{table*}[h!]
    \centering
    \caption{ HumanEval pass@$1$ accuracy (\%)  for self-conditioning auto-guidance.
    ``Context/no-SC'' retains clean context but removes SC (i.e., $\e^{-}_\theta\gets\e^{\c,\zero}_\theta$); ours removes both (i.e., $\e^{-}_\theta\gets\e^{\varnothing,\zero}_\theta$). Score entries are mean $\pm$ two standard errors over the 10 sampling lanes, an approximate $95\%$ confidence interval for the mean; $\Delta$ is ours $-$ context/no-SC.}
    \label{tab:cfg-bad-pred-humaneval}
    \setlength{\tabcolsep}{4.0pt}
    DDIM, 64 steps/block, uniform-$\gamma$ schedule, $\tau=0.3$\\
    \resizebox{\linewidth}{!}{%
    \begin{tabular}{cccccccccc}
        \toprule
        Guidance $w$ & 1 & 2 & 3 & 4 & 5 & 6 & 8 & 12 & 16 \\
        \midrule
        Context/no-SC & \se{42.93}{0.66} & \se{46.65}{1.08} & \se{50.00}{0.81} & \se{50.85}{0.61} & \se{\textbf{50.91}}{0.86} & \se{50.43}{0.66} & \se{49.39}{1.30} & \se{43.23}{1.20} & \se{40.85}{2.07} \\
        Ours    & \se{42.93}{0.66} & \se{53.96}{0.95} & \se{56.04}{1.16} & \se{\textbf{56.22}}{0.83} & \se{54.09}{1.22} & \se{54.51}{0.89} & \se{54.02}{0.78} & \se{43.66}{1.52} & \se{31.22}{1.62} \\
        $\Delta$ & 0.00 & +7.31 & +6.04 & +5.37 & +3.18 & +4.08 & +4.63 & +0.43 & -9.63 \\
        \bottomrule\vspace{1em}
    \end{tabular}
    }
    DDPM, 256 steps/block, EDM $\rho=7$ schedule, $\tau=1$\\
    \resizebox{\linewidth}{!}{%
    \begin{tabular}{ccccccccccc}
        \toprule
        Guidance $w$ & 1 & 2 & 3 & 4 & 5 & 6 & 7 & 8 & 9 & 10 \\
        \midrule
        Context/no-SC & \se{13.23}{1.50} & \se{20.85}{2.29} & \se{25.73}{1.51} & \se{27.93}{1.66} & \se{29.57}{1.03} & \se{29.76}{1.75} & \se{31.40}{2.07} & \se{32.38}{1.88} & \se{\textbf{32.50}}{2.26} & \se{30.00}{1.35} \\
        Ours    & \se{13.23}{1.50} & \se{32.62}{1.98} & \se{40.55}{1.79} & \se{44.63}{1.07} & \se{46.71}{1.82} & \se{46.22}{2.12} & \se{45.24}{1.49} & \se{45.61}{1.03} & \se{\textbf{46.95}}{1.73} & \se{45.37}{1.63} \\
        $\Delta$ & 0.00 & +11.77 & +14.82 & +16.70 & +17.14 & +16.46 & +13.84 & +13.23 & +14.45 & +15.37 \\
        \bottomrule
    \end{tabular}
    }
\end{table*}

\paragraph{Results.}
As shown in \cref{tab:cfg-bad-pred-gsm8k,tab:cfg-bad-pred-humaneval}, the fully unconditional bad prediction (ours) is consistently stronger in the useful guidance range:
it almost wins every matched CFG weight in sweep on both tasks and both samplers, and, tuned over the sweep, reaches a higher ceiling in each task.
Thus, we conclude that the advantage shown here is not an artifact of using different guidance search ranges.

\subsection{Ablation Study: Steering in Logit Space instead of Embedding Space}
\label{app:inference-logit-space}
[Back to~\cref{sec:inference-cfg-st} in the main paper.]

Both steering knobs introduced in \cref{sec:inference-cfg-st} act on the predicted clean embedding $\e_\theta \in \R^{L'\times d_e}$, equivalently on the score or velocity. Recall that our denoiser leverages \textit{categorical reparameterization}, i.e., $\e_\theta=\softmax(\bell_\theta)\Emb=\x_\theta\Emb$, where $\bell_\theta \in \R^{L' \times V}$ are the predicted logits, and $\softmax$ operates in the vocabulary dimension. Thus, a natural alternative is to apply the same operations to $\bell_\theta$ before the $\softmax$. This is the form of guidance used by discrete dLMs, which does not have embedding-space prediction to steer \citep{nie2025large,li2025adaptive,schiff2025simple} (\cref{app:related-work}).

\paragraph{Logit-space CFG (LCFG).} Reusing the notations in \cref{app:inference-cfg-bad-pred}, with the same fully unconditional bad prediction, we do softmax on the extrapolated logits:
\begin{align*}
    \e^{w,\text{logit}}_\theta
    := \softmax(\bell^{\varnothing,\zero}_\theta + w\,[\bell^{+}_\theta-\bell^{\varnothing,\zero}_\theta])\Emb,
    \qquad w\ge 1,
\end{align*}
against our default $\e^{w}_\theta=\e^{\varnothing,\zero}_\theta+w[\e^{+}_\theta-\e^{\varnothing,\zero}_\theta]$ of \cref{eq:cfg}, which extrapolates after the softmax. The two coincide at $w=1$ (i.e., no guidance).

\paragraph{Logit temperature (LT).} This is the analogue of ST in logit space, which rescales the conditional logits themselves:
\begin{align*}
    \e^{T}_\theta := \softmax\Big(\frac{\bell^{+}_\theta}{T}\Big)\Emb,
    \qquad T\in[0,1],
\end{align*}
where $T=1$ recovers the conditional mean embedding prediction (i.e., no guidance), and smaller $T$ sharpens. The $T\to 0$ limit is the maximum a posteriori (MAP) denoiser $\e^{0}_\theta=\Emb[\argmax\bell^{+}_\theta]$, which snaps the prediction onto a single vocabulary embedding.

We remark that both operations change the trajectory only; tokens are still read out as the $\argmax$ of the conditional logits at the final step $t=0$, exactly as in \cref{sec:inference-solver,alg:sample}. Also, LCFG does not cost extra compute compared with the embedding-space counterpart, since we need to get both conditional and unconditional logit predictions.

\paragraph{Experimental setup.} We use the setting identical to the default in \cref{app:inference-cfg-bad-pred}. We use the same \model\unreg-3B checkpoint with block length $32$, the full GSM8K and HumanEval evaluation sets, and generate 10 responses per problem. DDIM uses $64$ steps/block on a uniform-$\gamma$ time discretization with $\tau=0.3$, and DDPM uses $256$ steps/block on the EDM $\rho=7$ time discretization and $\tau=1$.

\begin{table*}[h!]
    \centering
    \caption{ Logit-space versus embedding-space CFG. GSM8K is accuracy (\%) and HumanEval is execution-based pass@$1$ (\%). Entries are mean $\pm$ two standard errors over the 10 generations per prompt, an approximate $95\%$ confidence interval for the mean, and \textbf{bold} marks the best entry in each row. The last column is the best value over the \emph{full} sweep of \cref{tab:cfg-bad-pred-gsm8k,tab:cfg-bad-pred-humaneval}, which continues past $w=5$ for the embedding rows. The logit rows were swept only to $w=5$ as they already peak at $w=2$ and fall monotonically after that.
    }
    \label{tab:logit-vs-embedding-cfg}
    \setlength{\tabcolsep}{4.0pt}
    \resizebox{\linewidth}{!}{%
    \begin{tabular}{cccccccc}
        \toprule
        Setting & Guidance space & $w=1$ & $w=2$ & $w=3$ & $w=4$ & $w=5$ & Tuned best \\
        \midrule
        \multirow{2}{*}{GSM8K, DDIM $64$ steps, $\tau=0.3$}
          & Logit             & \se{68.13}{0.38} & \se{\textbf{83.49}}{0.22} & \se{81.55}{0.41} & \se{77.08}{0.41} & \se{67.78}{0.53} & \se{\textbf{83.49}}{0.22} ($w=2$) \\
          & Embedding (ours)  & \se{68.13}{0.38} & \se{82.52}{0.21} & \se{82.84}{0.28} & \se{83.04}{0.20} & \se{\textbf{83.19}}{0.18} & \se{83.19}{0.18} ($w=5$) \\
        \midrule
        \multirow{2}{*}{GSM8K, DDPM $256$ steps, $\tau=1$}
          & Logit             & \se{53.09}{0.85} & \se{\textbf{76.76}}{0.50} & \se{75.16}{0.79} & \se{69.48}{0.59} & \se{60.77}{0.99} & \se{76.76}{0.50} ($w=2$) \\
          & Embedding (ours)  & \se{53.09}{0.85} & \se{77.12}{0.66} & \se{80.04}{0.39} & \se{80.99}{0.48} & \se{\textbf{81.30}}{0.41} & \se{\textbf{82.43}}{0.50} ($w=10$) \\
        \midrule
        \multirow{2}{*}{HumanEval, DDIM $64$ steps, $\tau=0.3$}
          & Logit             & \se{42.93}{0.66} & \se{\textbf{53.78}}{1.09} & \se{39.15}{1.13} & \se{20.67}{0.84} & \se{12.68}{0.47} & \se{53.78}{1.09} ($w=2$) \\
          & Embedding (ours)  & \se{42.93}{0.66} & \se{53.96}{0.95} & \se{56.04}{1.16} & \se{\textbf{56.22}}{0.83} & \se{54.09}{1.22} & \se{\textbf{56.22}}{0.83} ($w=4$) \\
        \midrule
        \multirow{2}{*}{HumanEval, DDPM $256$ steps, $\tau=1$}
          & Logit             & \se{13.23}{1.50} & \se{\textbf{29.70}}{2.69} & \se{22.68}{1.36} & \se{14.51}{2.14} & \se{9.76}{0.48} & \se{29.70}{2.69} ($w=2$) \\
          & Embedding (ours)  & \se{13.23}{1.50} & \se{32.62}{1.98} & \se{40.55}{1.79} & \se{44.63}{1.07} & \se{\textbf{46.71}}{1.82} & \se{\textbf{46.95}}{1.73} ($w=9$) \\
        \bottomrule
    \end{tabular}
    }
\end{table*}

\begin{table*}[h!]
    \centering
    \caption{ Logit temperature, GSM8K accuracy (\%) with DDPM, $256$ steps/block, EDM $\rho=7$ schedule, no CFG. Score entries are mean $\pm$ two standard errors over the 10 generations per prompt, an approximate $95\%$ confidence interval for the mean. $T=1$ is the unsharpened MMSE denoiser and is the same configuration as the $w=1$ logit-space entry of \cref{tab:logit-vs-embedding-cfg}. The bottom row is LCFG at the matched weight $w=\frac1T$, repeated from \cref{tab:logit-vs-embedding-cfg}; the two are the same operation if the unconditional logits are uniform over the vocabulary, so the gap isolates what logit-space guidance adds over plain sharpening.}
    \label{tab:logit-temperature}
    \setlength{\tabcolsep}{4.0pt}
    \resizebox{\linewidth}{!}{%
    \begin{tabular}{cccccccc}
        \toprule
        $T$ & $1$ & $\frac12$ & $\frac13$ & $\frac14$ & $\frac15$ & $\frac1{10}$ & $0$ (MAP) \\
        \midrule
        LT at $T$ & \se{52.44}{1.00} & \se{76.88}{0.62} & \se{79.07}{0.38} & \se{78.97}{0.50} & \se{\textbf{79.11}}{0.32} & \se{78.74}{0.23} & \se{78.61}{0.33} \\
        \midrule
        LCFG at $w=\frac1T$ & \se{52.44}{1.00} & \se{76.76}{0.50} & \se{75.16}{0.79} & \se{69.48}{0.59} & \se{60.77}{0.99} & --- & --- \\
        $\Delta$ (LT $-$ LCFG) & 0.00 & +0.12 & +3.91 & +9.49 & +18.34 & --- & --- \\
        \bottomrule
    \end{tabular}
    }
\end{table*}

\begin{table*}[h!]
    \centering
    \caption{ Composing logit temperature with the other two knobs: GSM8K accuracy (\%) with DDIM, $64$ steps/block, uniform-$\gamma$ schedule, ST $\tau=0.3$. Entries are mean $\pm$ two standard errors over the 10 generations per prompt, an approximate $95\%$ confidence interval for the mean.
    \textbf{Bold} marks the best entry in each row.}
    \label{tab:logit-temp-ddim}
    \setlength{\tabcolsep}{4.0pt}
    \begin{tabular}{cccccc}
        \toprule
        $T$ & $1$ & $\frac12$ & $\frac13$ & $\frac15$ & $0$ (MAP) \\
        \midrule
        No guidance ($w=1$) & \se{68.18}{0.33} & \se{\textbf{77.60}}{0.23} & \se{77.42}{0.35} & \se{76.13}{0.31} & \se{71.83}{0.24} \\
        Embedding CFG at $w=5$ & \se{\textbf{83.19}}{0.18} & \se{81.44}{0.19} & \se{81.49}{0.25} & \se{79.96}{0.31} & \se{76.84}{0.44} \\
        \bottomrule
    \end{tabular}
\end{table*}

\paragraph{Results.}
Logit-space CFG (LCFG) peaks at $w=2$ in all four settings, whereas embedding-space CFG improves up to $w=4\sim10$; at their respective optima, LCFG gains only $0.30$ points on GSM8K/DDIM and loses $2.44\sim17.26$ points elsewhere (\cref{tab:logit-vs-embedding-cfg}). Logit temperature (LT) provides effective sharpening without a second branch: on DDPM, lowering $T$ from $1$ to $1/5$ raises GSM8K from $52.44$ to $79.11$, while the MAP endpoint $T=0$ retains $78.61$ (\cref{tab:logit-temperature}). Relative to matched sharpening ($T=1/w$), the unconditional branch in LCFG contributes negligibly at $w=2$ and hurts at larger weights. LT also complements ST, adding $9.42$ points on DDIM with ST $=0.3$, but reduces accuracy when combined with both CFG and ST (\cref{tab:logit-temp-ddim}). We therefore retain embedding-space CFG and ST as the default choice.
\subsection{Ablation Study: Commit-Stride Sampling}
\label{app:inference-stride-sampling}
[Back to~\cref{sec:inference} in the main paper.]

Beyond blockwise decoding of $L'$ tokens per denoising window, we further introduce a \emph{commit stride} $S \in \{1, \dots, L'\}$.
Each denoising window still runs the same blockwise reverse process over a block of fresh Gaussian noise with length $L'=32$, but only the leading $S \in \{8, 16, 32\}$ tokens are committed back into the causal KV cache; the remaining $L' - S$ tail positions are discarded and re-drawn as fresh noise in the next window (which is shifted forward by $S$ absolute positions). Setting $S\gets L'$ recovers the default block-decoding method.
The pseudocode for commit stride can be found in \cref{alg:sample}.
This is a pure inference-time technique that further improves the output quality even under matched compute budgets, as we will see below.

\paragraph{Setup.} We evaluate on the base models of \cref{tab:base-3b,tab:base-8b}: \model\unreg-3B/8B, both with block size $L' = 32$. Decoding invariants match the diffusion-mode rows of \cref{tab:base-3b}. For each stride, CFG $w$ and ST $\tau$ are chosen based on the number of DDIM steps per window (\cref{tab:acc-nfe-3b}): $(w,\tau) = (3,0.2)$ at $\geq 64$ steps, $(3,0.3)$ at $32$ steps, $(3,0.5)$ at $16$ steps, and $(3,0.8)$ at $8$ steps. All cells generate $10$ samples for each prompt and report mean $\pm$ two standard errors (\cref{app:inference-eval-setup}).

\paragraph{Nominal vs.\ true NFE.} For simplicity, we match compute cost by \emph{nominal} NFE, defined as the number of denoiser forward passes required to commit $L'=32$ tokens under the given (steps, stride) pair:
$$
\text{Nominal~NFE} \;:=\; \frac{NL'}{S},
$$
where $N$ is the number of diffusion steps per denoising window.
Since every denoising window operates on the \emph{same} $L'$-wide noisy block appended to the same-length clean prefix, one forward pass costs the same FLOPs regardless of $S$.

\begin{table}[h]
    \centering
    \caption{ \textbf{Commit-stride ablation on \model\unreg-3B.} Same six generative reasoning tasks as \cref{tab:base-3b}. \emph{Nom. NFE} is the nominal number of forward passes per $32$ committed tokens (i.e., $\text{steps}/\text{stride} \times 32$). All DDIM rows use the uniform-$\gamma$ schedule; the DDPM row uses the EDM $\rho = 7$ schedule. Rows within each nominal-NFE bucket are directly compute-comparable. Mean $\pm$ two standard errors across $10$ sampling lanes. \textbf{Bold} marks the best value per column across the stride sweep.}
    \label{tab:stride-3b}
    \resizebox{\linewidth}{!}{%
    \begin{tabular}{ccccccccccc}
    \toprule
    Sampler & Nom. NFE & Steps & Stride & GSM8K & Minerva Math & HumanEval & HumanEval+ & MBPP & MBPP+ & Avg. \\
    \midrule
    DDPM ($w{=}2,\tau{=}0.1$) & 256 & 256 & 32 & \se{84.50}{0.31} & \se{64.60}{0.28} & \se{59.45}{1.30} & \se{53.84}{1.29} & \se{52.32}{0.73} & \se{71.32}{0.91} & 64.34 \\
    DDIM ($w{=}3,\tau{=}0.2$) & 256 & 128 & 16 & \se{85.70}{0.15} & \se{64.68}{0.20} & \se{60.67}{0.80} & \se{54.88}{0.84} & \se{55.52}{0.36} & \se{72.14}{0.34} & 65.60 \\
    DDIM ($w{=}3,\tau{=}0.2$) & 256 & 64 & 8 & \se{84.94}{0.17} & \se{\textbf{64.73}}{0.18} & \se{61.52}{0.76} & \se{56.10}{0.62} & \se{\textbf{58.86}}{0.35} & \se{\textbf{75.87}}{0.40} & \textbf{67.00} \\
    \midrule
    DDIM ($w{=}3,\tau{=}0.2$) & 128 & 128 & 32 & \se{83.78}{0.11} & \se{63.89}{0.13} & \se{59.15}{0.68} & \se{53.17}{0.87} & \se{50.48}{0.21} & \se{68.12}{0.51} & 63.10 \\
    DDIM ($w{=}3,\tau{=}0.2$) & 128 & 64 & 16 & \se{\textbf{85.82}}{0.09} & \se{64.24}{0.14} & \se{60.73}{0.83} & \se{54.63}{0.85} & \se{54.06}{0.30} & \se{71.56}{0.44} & 65.17 \\
    DDIM ($w{=}3,\tau{=}0.3$) & 128 & 32 & 8 & \se{84.17}{0.29} & \se{64.44}{0.22} & \se{\textbf{62.13}}{0.63} & \se{\textbf{56.34}}{0.49} & \se{57.56}{0.61} & \se{74.68}{0.50} & 66.55 \\
    \midrule
    DDIM ($w{=}3,\tau{=}0.2$) & 64 & 64 & 32 & \se{82.93}{0.15} & \se{63.93}{0.13} & \se{56.89}{0.73} & \se{50.91}{0.66} & \se{49.26}{0.27} & \se{66.35}{0.35} & 61.71 \\
    DDIM ($w{=}3,\tau{=}0.3$) & 64 & 32 & 16 & \se{85.21}{0.25} & \se{63.44}{0.16} & \se{59.82}{0.92} & \se{54.27}{1.16} & \se{51.88}{0.62} & \se{69.87}{0.49} & 64.08 \\
    DDIM ($w{=}3,\tau{=}0.5$) & 64 & 16 & 8 & \se{83.95}{0.28} & \se{62.97}{0.23} & \se{59.63}{1.06} & \se{53.66}{0.94} & \se{54.90}{0.50} & \se{73.10}{0.57} & 64.70 \\
    \midrule
    DDIM ($w{=}3,\tau{=}0.3$) & 32 & 32 & 32 & \se{81.91}{0.32} & \se{62.76}{0.18} & \se{56.59}{1.31} & \se{50.67}{1.00} & \se{47.26}{0.46} & \se{63.49}{0.77} & 60.45 \\
    DDIM ($w{=}3,\tau{=}0.5$) & 32 & 16 & 16 & \se{82.40}{0.38} & \se{62.37}{0.21} & \se{56.77}{0.82} & \se{51.52}{0.90} & \se{50.12}{0.71} & \se{67.72}{0.46} & 61.82 \\
    DDIM ($w{=}3,\tau{=}0.8$) & 32 & 8 & 8 & \se{80.60}{0.47} & \se{59.48}{0.22} & \se{52.56}{1.94} & \se{46.95}{1.86} & \se{46.50}{0.80} & \se{60.50}{1.02} & 57.76 \\
    \bottomrule
    \end{tabular}%
    }
\end{table}

\begin{table}[h]
    \centering
    \caption{ \textbf{Commit-stride ablation on \model\unreg-8B.} Same six tasks as \cref{tab:base-8b}; conventions match \cref{tab:stride-3b}. \textbf{Bold} marks the best value per column across the stride sweep.}
    \label{tab:stride-8b}
    \resizebox{\linewidth}{!}{%
    \begin{tabular}{ccccccccccc}
    \toprule
    Sampler & Nom. NFE & Steps & Stride & GSM8K & Minerva Math & HumanEval & HumanEval+ & MBPP & MBPP+ & Avg. \\
    \midrule
    DDPM ($w{=}2,\tau{=}0.1$) & 256 & 256 & 32 & \se{89.07}{0.28} & \se{67.47}{0.20} & \se{\textbf{61.40}}{1.73} & \se{\textbf{57.44}}{1.62} & \se{64.78}{0.54} & \se{77.83}{0.72} & 69.66 \\
    DDIM ($w{=}3,\tau{=}0.2$) & 256 & 128 & 16 & \se{88.64}{0.11} & \se{67.59}{0.10} & \se{60.91}{0.64} & \se{55.67}{0.60} & \se{67.44}{0.35} & \se{\textbf{81.75}}{0.33} & 70.33 \\
    DDIM ($w{=}3,\tau{=}0.2$) & 256 & 64 & 8 & \se{\textbf{90.17}}{0.15} & \se{\textbf{68.82}}{0.20} & \se{60.49}{0.98} & \se{53.96}{0.93} & \se{\textbf{67.72}}{0.21} & \se{81.35}{0.28} & \textbf{70.42} \\
    \midrule
    DDIM ($w{=}3,\tau{=}0.2$) & 128 & 128 & 32 & \se{88.70}{0.14} & \se{67.28}{0.09} & \se{58.48}{1.00} & \se{52.68}{1.00} & \se{63.22}{0.28} & \se{76.85}{0.31} & 67.87 \\
    DDIM ($w{=}3,\tau{=}0.2$) & 128 & 64 & 16 & \se{88.76}{0.17} & \se{67.63}{0.17} & \se{59.33}{1.06} & \se{53.54}{0.94} & \se{66.62}{0.42} & \se{80.90}{0.31} & 69.46 \\
    DDIM ($w{=}3,\tau{=}0.3$) & 128 & 32 & 8 & \se{89.61}{0.24} & \se{68.11}{0.15} & \se{61.22}{0.55} & \se{55.18}{0.69} & \se{66.22}{0.49} & \se{80.19}{0.34} & 70.09 \\
    \midrule
    DDIM ($w{=}3,\tau{=}0.2$) & 64 & 64 & 32 & \se{88.68}{0.19} & \se{67.30}{0.18} & \se{58.90}{0.71} & \se{52.56}{0.54} & \se{61.96}{0.36} & \se{76.35}{0.28} & 67.62 \\
    DDIM ($w{=}3,\tau{=}0.3$) & 64 & 32 & 16 & \se{88.15}{0.33} & \se{67.58}{0.21} & \se{58.05}{1.09} & \se{52.80}{0.98} & \se{65.32}{0.41} & \se{78.10}{0.59} & 68.33 \\
    DDIM ($w{=}3,\tau{=}0.5$) & 64 & 16 & 8 & \se{88.45}{0.31} & \se{67.21}{0.26} & \se{61.40}{1.73} & \se{55.67}{1.31} & \se{61.92}{0.77} & \se{75.21}{0.73} & 68.31 \\
    \midrule
    DDIM ($w{=}3,\tau{=}0.3$) & 32 & 32 & 32 & \se{87.78}{0.33} & \se{67.13}{0.15} & \se{58.54}{1.04} & \se{53.17}{1.27} & \se{59.54}{0.57} & \se{74.15}{0.43} & 66.72 \\
    DDIM ($w{=}3,\tau{=}0.5$) & 32 & 16 & 16 & \se{87.04}{0.35} & \se{67.14}{0.27} & \se{58.41}{1.22} & \se{53.29}{1.19} & \se{60.46}{0.39} & \se{73.41}{0.51} & 66.63 \\
    DDIM ($w{=}3,\tau{=}0.8$) & 32 & 8 & 8 & \se{85.91}{0.29} & \se{64.57}{0.22} & \se{55.98}{1.46} & \se{50.61}{1.75} & \se{52.74}{0.85} & \se{64.44}{0.87} & 62.38 \\
    \bottomrule
    \end{tabular}%
    }
\end{table}

\paragraph{Results.}
We present the results for commit-stride sampling for \model\unreg-3B/8B in \cref{tab:stride-3b,tab:stride-8b}, respectively. The tables sweep all three strides $S \in \{8, 16, 32\}$ within each nominal-NFE bucket $\{256, 128, 64, 32\}$. At the top NFE (nominal $256$), the DDIM $64$ steps / stride $8$ row is the best on both models: it beats the tuned DDPM-256 baseline of \cref{tab:base-3b,tab:base-8b} by $+2.66$ average points on 3B and $+0.76$ on 8B, with the largest per-task gains on the coding benchmarks such as MBPP. The same ordering holds down to around nominal $64$: committing fewer tokens per window (i.e., smaller stride $S$) improves the average monotonically. At nominal $128$ NFE, DDIM $32$ steps / stride $8$ reaches within a point of the top budget on both sizes, and still above the full-stride DDPM-256 baselines. The benefit of a smaller stride disappears once it forces the per-window step count below roughly $16$: at nominal $32$, stride $8$ leaves only $8$ steps per window, and its coarse time grid collapses quality due to larger discretization error, falling below the stride-$16$ and stride-$32$ rows at the same budget.

\subsection[Pass@$k$ Evaluation]{Pass@$k$ Evaluation
}
\label{app:inference-passk}
[Back to~\cref{sec:inference-result} in the main paper.]

We study inference-time scaling in two distinct evaluation regimes. 
First, we compare the pretrained 3B models on MBPP;
second, we compare the instruct 3B models on AIME 2025.

\paragraph{MBPP for base models.}

\begin{figure*}[h]
    \centering
    \includegraphics[width=0.90\linewidth]{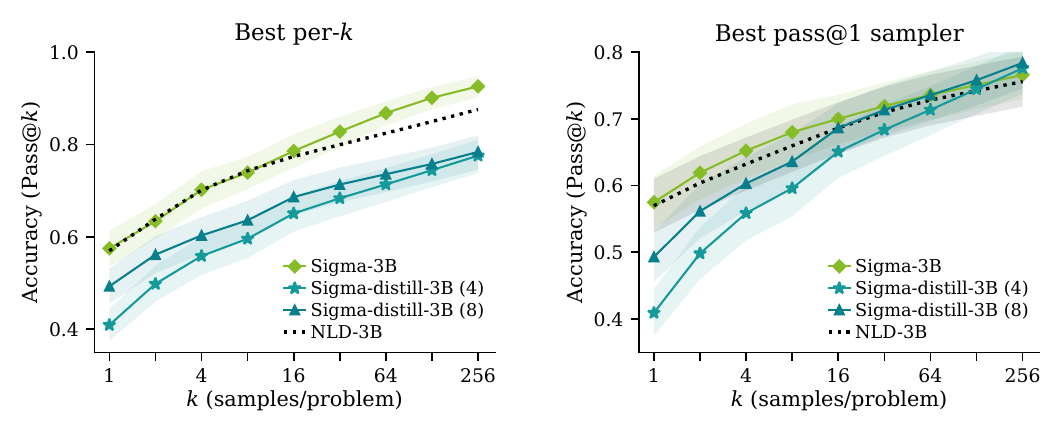}
    \caption{
    \textbf{Pass@$k$ on MBPP of \model-3B, \model\distill-3B and NLD-3B.}
    \emph{Left:} At each $k$, each family uses the best configuration in its tested sampler grid; 
    \emph{Right:} Fixed sampler configuration picked by highest pass@$1$ accuracy: NLD-3B $(\vartheta=0.9,T=0.3)$, \model-3B $(DDPM, w=2,\tau=0.1)$, \model\distill-3B have a fixed sampler (no CFG/ST). 
    }
    \label{fig:passk-mbpp}
\end{figure*}

We compare \model-3B, \model\distill-3B and NLD-3B \citep{fu2026nemotron} on MBPP, drawing 256 generations per problem for every configuration. Both models use block size 32. \model-3B uses DDPM with 256 steps/block and the EDM $\rho=7$ schedule, sweeping CFG $w\in\{1,2,3\}$ and ST $\tau\in\{0.1,0.4,0.7,1.0\}$; NLD-3B uses 32 transfer steps per block, sweeping confidence threshold $\vartheta\in\{0.8,0.9\}$ and temperature $T\in\{0.3,0.6,1.0\}$; \model\distill-3B use 4 and 8 NFEs and dedicated sampler without any post-hoc guidance. Candidate draws use different random seeds while preserving identical prompts and few-shot examples across configurations, as in \cref{app:inference-eval-setup}.

The left panel of \cref{fig:passk-mbpp} shows the best performance available in each tested sampler grid at every generation budget. As $k$ grows, the \model-3B setting carrying this curve moves from $\tau=0.1$ to $0.4$, $0.7$, and finally $1.0$, shifting from single-sample quality toward broader coverage. 
the right panel provides the complementary fixed-configuration result tuned on pass@1 (i.e. that in \cref{tab:base-3b}). 
\model-3B consistently beats NLD=3B by a small, almost constant margin. 
However, most interesting is the behavior of \model\distill-3B which with only 8-NFEs matches the \model-3B teacher from $k=32$ onward, and the 4-NFE version closes the gap as $k$ increases, finally matching \model-3B at $k=256$.
This demonstrates that our distilled models exercise \textit{useful} diversity.

\begin{wrapfigure}{r}{0.4\linewidth}
    \centering
    \includegraphics[width=\linewidth]{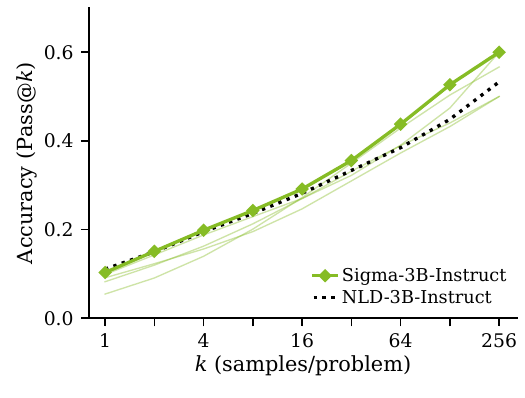}
    \caption{\textbf{Pass@$k$ on AIME 2025 of \model-3B-Instruct and NLD-3B-Instruct.}
    At each $k$, each family uses the best configuration in its tested sampler grid; Best overall \model-3B-Instruct configuration is highlighted, the other thin lines correspond to the other configs.
    }
    \label{fig:passk-aime25}
    \vspace{-2em}
\end{wrapfigure}

\paragraph{AIME 2025 for instruct models.}
We repeat the comparison on the \emph{instruct} models, under the same protocol as the benchmark table in~\cref{tab:sft-3b} (harness described in~\cref{app:sft-eval}), on AIME 2025.
\model-3B-\instruct uses DDIM with 32 steps and the uniform-$\gamma$ schedule, sweeping the five settings $(w,\tau)\in\{(3,0.3),\,(4,0.3),\,(4,0.7),\,(5,0.7),\,(3,1.0)\}$; NLD-3B-Instruct uses confidence-threshold masked decoding with $\vartheta=0.9$ and $T\in\{0.2,0.6,1.0\}$.

The response budgets are 4096 tokens for \model-3B-\instruct and 16384 for NLD-3B-Instruct, but the shorter \model cap is not binding since the reasonable generations (i.e., the samples that are not degenerate, such as repeating the same pattern without emitting EOS token) are all within this length range for both models.
The results are in \cref{fig:passk-aime25}.

\subsection{Wall-Clock Cost of the Packed CFG Forward}
\label{app:inference-cfg-fwd-cost}
[Back to~\cref{sec:inference-cfg-st} in the main paper.]

\begin{table}[h]
    \centering
    \caption{
    Wall-clock cost per denoiser evaluation (NFE) of packing the conditional and unconditional CFG branches into one forward pass, at batch size $1$. Times are in milliseconds on H100. 
    }
    \label{tab:cfg-fwd-cost}
    \resizebox{\linewidth}{!}{%
    \begin{tabular}{cccccc}
    \toprule
    Block Size & Stage & No CFG (ms) & Packed CFG (ms) & $\Delta$ & $\Delta\ (\%)$ \\
    \midrule
    \multirow{3}{*}{$L'=32$}
      & Transformer forward         & $18.06$ & $18.38$ & $+0.322 \pm 0.181$ & $+1.8$ \\
      & Logit head + cat. reparam.  & $1.02$  & $1.50$  & $+0.481 \pm 0.008$ & $+47.2$ \\
      & Total per NFE               & $19.08$ & $19.89$ & $+0.804 \pm 0.186$ & $+4.2$ \\
    \midrule
    \multirow{3}{*}{$L'=64$}
      & Transformer forward         & $18.23$ & $18.21$ & $-0.021 \pm 0.159$ & $-0.1$ \\
      & Logit head + cat. reparam.  & $1.08$  & $1.60$  & $+0.519 \pm 0.006$ & $+48.3$ \\
      & Total per NFE               & $19.31$ & $19.80$ & $+0.498 \pm 0.162$ & $+2.6$ \\
    \bottomrule
    \end{tabular}
    }
\end{table}

We measure the wall-clock cost of the packed CFG forward with attention mask (\cref{fig:cfg-attention}) at batch size $1$ compared with the unguided forward. We decode the same set of $16$ GSM8K questions ($8$-shot, prompt length $729\sim814$) with the DDIM sampler at $32$ steps per block, on the first block only.
Each question therefore requires $32+1$ forward passes (\cref{alg:sample}). 
The very first denoiser query of each process involves a one-time CUDA initialization which takes a longer time, so is excluded to avoid biasing the measurement. All runs are performed on one NVIDIA H100 GPU, using \model\unreg-3B with precision \texttt{BF16}.

Results in \cref{tab:cfg-fwd-cost} show that at $L'=32$ the packed transformer forward costs $+1.8\%$ over the unguided one, and at $L'=64$ it is free to within measurement noise. The per-NFE overhead is dominated by computing the second set of logits, doing softmax, and multiplying by the embedding matrix rather than by attention.

\subsection{Reasoning Trajectory Analysis}
\label{app:inference-reason-traj}
[Back to~\cref{subsec:3b-base-math-code} in the main paper.]

\begin{figure*}[p]
    \centering
    \includegraphics[width=0.7\linewidth]{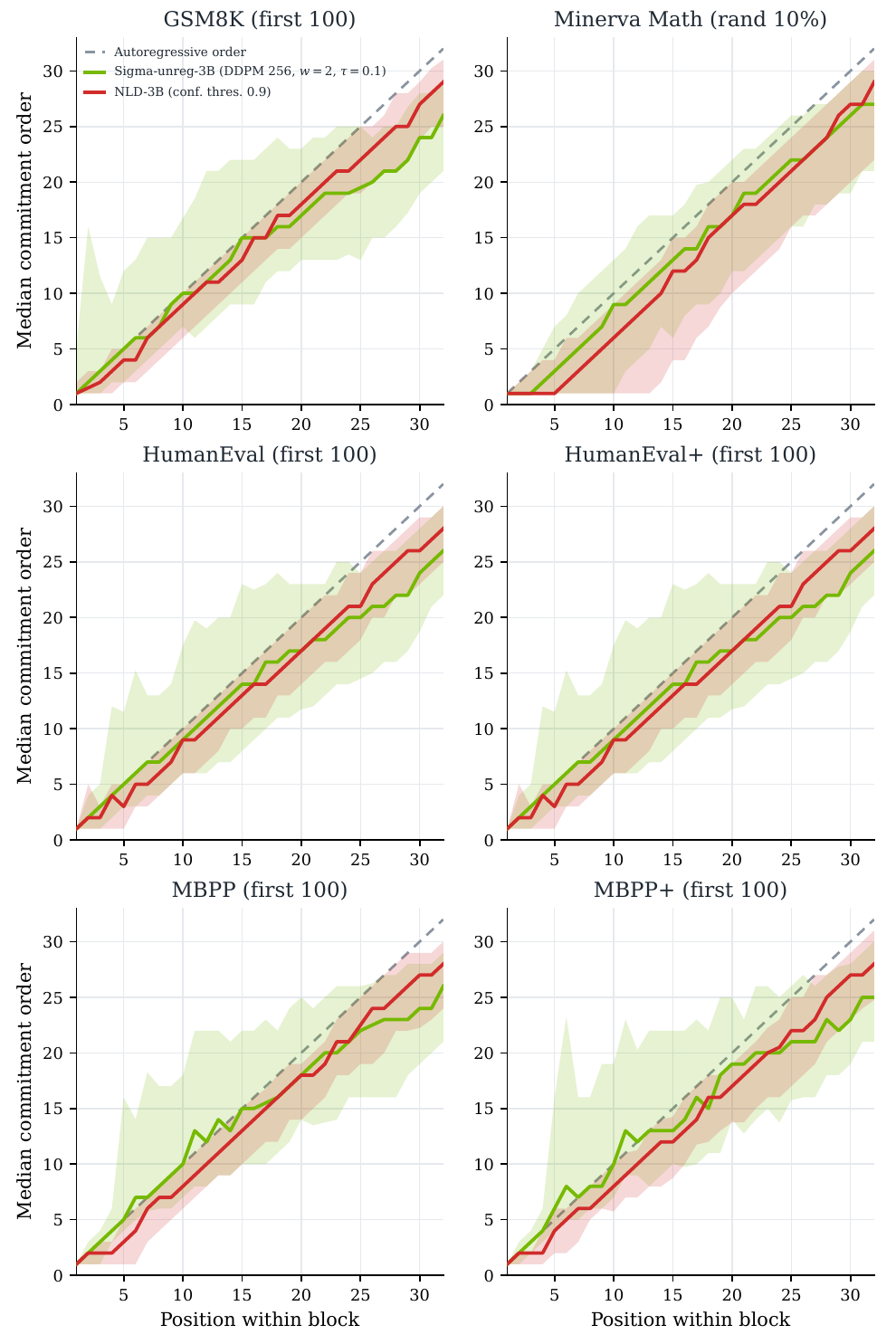}
    \caption{ \textbf{Token commitment order vs token position for \model\unreg-3B}
    \model uses 256-step DDPM ($w=2,\tau=0.1$) as in \cref{tab:base-3b}. Generation order averaged across multiple documents and blocks. For \model\unreg-3B, the commitment step $t_i$ of position $i$ is defined as the smallest $t$ such that for every step $t' \ge t$ up to the final readout, (i) the argmax of $\mathbf{x}_\theta$ equals the finally decoded token, and (ii) that argmax is assigned probability at least $0.9$. The commitment order of position $i$ is the rank of $t_i$ among ${t_1,\dots,t_{32}}$ within its block, in ascending order. Positions sharing a commitment step share the lowest rank of the tie.}
    \label{fig:commitment-order}
\end{figure*}

As noted in \cref{sec:inference-result}, we empirically observe less left-to-right bias within a block for \model compared with the confidence-based masked dLM sampling. We demonstrate this in~\cref{fig:commitment-order}, where we show a commit order analysis between continuous and discrete dLMs.

\clearpage 

\section{Implementation Details and Additional Results for Few-step Distillation}\label{app:distillation-details}

\subsection{Method and Training Objective}
We initialize a student with parameters $\phi$ from the teacher parameters $\theta$, and devise a continuous relaxation of Distribution Matching Distillation (DMD, \citet{yin2024one}) for language models, inspired by \citet{van2026distribution} and \citet{fan2026phased}, to train the student to estimate a clean-context-conditional teacher-generated data distribution $p_\text{teacher}(\cdot \mid \c )$. In the following we omit the context conditioning for readability.
As detailed in \cref{app:dmd-details}, this distribution is obtained using the strongest teacher sampler for optimal distributional quality (high NFE and guidance).
We choose $p_\text{teacher}$ as the target distribution instead of using the pretraining dataset, as we observed that distillation performance was better for the former, likely due to the inductive bias existing in self-distillation since the student is initialized from the teacher. 
We ablate this choice along with other design decisions in \cref{app:dmd-additional-results}.

We train the student on-policy for a specific number of NFEs $M$, i.e., the student should produce its own multi-step mapping from the Gaussian prior to the target data distribution by matching intermediate noisy marginals, following the principle in \citet{fan2026phased}.
The training procedure is detailed in \cref{alg:dmd-training}.

At each training iteration, 
we pick a clean target sample $\e \sim p_\text{teacher}$ and independently initialize the rollout with a near-standard-Gaussian state $\e_1 \sim q_1(\zero, \gamma_1)$.
The student is rolled out for $M_s \sim \unif\{0,\dots,M-1\}$ steps from $\e_1$ until a \textit{source state} $\e_s$ with noise level $\gamma_s = \gamma_1 - \frac{M_s}{M} ( \gamma_1 - \gamma_0 )$ using self-conditioning. 
The multi-step sampler we use for rollout as well as during inference is the consistency-model-style procedure used in \citet{yin2024improved,song2023consistency}:
\begin{equation}\label{eq:dmd-update}
    \e_{t_{n-1}} = \alpha_{t_{n+1}} \e_0^{\phi,t_{n}} + \sigma_{t_{n+1}} \beps
    \qquad
    \e_0^{\phi,t_{n}} = \e_\phi(\e_{t_{n}},\gamma_{t_{n}})
    \qquad
    \beps \sim \mathcal{N}(\zero,\I)\,,
\end{equation}
This means the intermediate states have the exact same noise marginals as the forward variance-preserving process, but the state is centered around the current clean estimate.
We detach this prefix rollout and treat its source-state distribution as fixed during the selected student update: the student objective is not differentiated through the earlier rollout steps.

\begin{algorithm}[t]
\caption{On-policy distribution-matching distillation}
\label{alg:dmd-training}
\begin{algorithmic}[1]
\Require Teacher $\theta$, teacher-generated data $p_{\mathrm{teacher}}$, $M$ NFE count
\State Initialize independent parameter copies
       $\phi,\eta,\psi \gets \theta$
\State $\Delta\gamma \gets (\gamma_1-\gamma_0)/M$;
       $\bar\gamma_j \gets \gamma_1-j\Delta\gamma$
       for $j=0,\ldots,M$ \Comment{Uniform logNSR discretization}

        \vspace{1em}
    \algsection{yellow}{Reference pre-fitting}

       \For{$n = 1, \dots, N_\text{ref-fitting}$}
        \State $\gamma_u\sim\unif[\gamma_0,\gamma_1]$;
               $\e_u\sim q_u(\e\mid\gamma_u)$
        \State $\cL^\psi_{\mathrm{ref}}\gets
          \mathrm{CE}\big(
          \x, \x_\psi(\e_u,\gamma_u)
          \big)$ 
          \Comment{Reference loss \eqref{eq:dmd-reference}}
        \State $\psi \gets \text{AdamW}(\cL^\psi_{\mathrm{ref}})$
            \Comment{Reference update}
        \EndFor
        \vspace{1em}

\While{not converged}
    \State $\x\sim p_{\mathrm{teacher}}$;
           $\e\gets\x\Emb$ \Comment{Sample target data}

           \algsection{orange}{Student rollout}
    \State $M_s\sim\unif\{0,\ldots,M-1\}$ \Comment{Sample rollout depth}

    \State $\e_s\gets\sigma(\gamma_1)\beps$,
           $\beps\sim\mathcal{N}(\zero,\I)$;
           $\e_{\mathrm{SC}}\gets\varnothing$
        \Comment{Initialize rollout states}
    \For{$j=0,\ldots,M_s-1$}
        \State $\widehat\e_0\gets
          \e_\phi(\e_s,\bar\gamma_j;\e_{\mathrm{SC}})$
          \Comment{Rollout transition with carried self-conditioning}
        \State $\e_s\gets
          \alpha(\bar\gamma_{j+1})\widehat\e_0
          +\sigma(\bar\gamma_{j+1})\beps$,
          $\beps\sim\mathcal{N}(\zero,\I)$
          \Comment{Consistency-style transition}
        \State $\e_{\mathrm{SC}}\gets\widehat\e_0$ \Comment{Update self-conditioning}
    \EndFor

    \State $\gamma_s\gets\bar\gamma_{M_s}$;
           $\gamma_r\gets\bar\gamma_{M_s+1}$ \Comment{Segment start and end points}
    \State $\e_0^{\phi,s}\gets
    \e_\phi(\sg[\e_s],
                     \gamma_s)$ 
                     \Comment{Student prediction}

       \algsection{green}{Auxiliary updates}
    \For{$k=1,\ldots,4$}
        \State $\gamma_t\sim\unif[\gamma_r,\gamma_s]$;
          $\e_t^{\phi,s}\sim
          q_t(\sg[\e_0^{\phi,s}]\mid\gamma_t)$ \Comment{Re-noise student sample to query point}
        \State $\cL^\eta_{\mathrm{aux}}\gets
          h_t\,\mathrm{CE}\big(
          \sg[\x^{\phi,s}],
          \e_\eta(\e_t^{\phi,s},\gamma_t) \big)$ \Comment{Auxiliary loss \eqref{eq:dmd-auxiliary}}
        \State $\eta \gets \text{AdamW}(\cL^\eta_\text{aux})$ 
            \Comment{Auxiliary update}

    \EndFor

    \algsection{blue}{Student update}
    \State $\gamma_t\sim\unif[\gamma_r,\gamma_s]$; 
      $\e_t^{\phi,s}\sim
      q_t(\sg[\e_0^{\phi,s}]\mid\gamma_t)$ \Comment{Re-noise student sample to query point}
    \State $\widehat\e_\eta\gets
      \e_\eta(\e_t^{\phi,s},\gamma_t)$;
      $\widehat\e_\psi\gets
      \e_\psi(\e_t^{\phi,s},\gamma_t)$
        \Comment{Auxiliary and reference denoising}
    \State $\cL^\phi_{\mathrm{DMD2\text{-}grad}}\gets
      \frac{1}{R} \displaystyle\sum_{l=1}^{L'}
      \big\langle
      \e_0^{\phi,s,l}, \sg\big[ \widehat\e_\eta - \widehat\e_\psi \big]
      \big\rangle$ 
      \Comment{Student loss \eqref{eq:dmd2-grad}}
        \State $\phi \gets \text{AdamW}(\cL^\phi_{\mathrm{DMD2\text{-}grad}})$ 
            \Comment{Student update}
\EndWhile
\State \Return $\phi$

\end{algorithmic}
\end{algorithm}

From the given source state $\e_s$, the student is called to produce a clean embedding estimate $\e_0^{\phi,s} := \e_\phi(\e_s, \gamma_s)$, which is re-noised to a \textit{query noise level} sampled uniformly within the segment starting from the source noise level $\gamma_t \sim \unif\Big[ \gamma_1 - \frac{M_s+1}{M} ( \gamma_1 - \gamma_0 ) , \gamma_1 - \frac{M_s}{M} ( \gamma_1 - \gamma_0 )  \Big]$. 
This together yields the re-noised student sample $\e^{\phi, s}_t \sim q_t(\e_0^{\phi, s} \mid \gamma_t)$
and we denote the resulting marginal as $q^{\phi, s}_t$.
Similarly, the target embedding $\e$ is re-noised to the same query level, yielding a sample $\e_t \sim q_t(\e, \gamma_t)$.

These choices of source and query distribution correspond to a specific choice of components described in \citet{fan2026phased}. 
Like \citet{fan2026phased}, we train our students in an on-policy fashion; however, we use fully stochastic transitions as in \citet{yin2024improved,van2026distribution} whereas \citet{fan2026phased} used deterministic DDIM transitions. Furthermore, we do not use phase experts but only one set of student weights.
As we ablate it later, this specific choice of source and query noise level distributions has a large influence on the overall student quality.

We match the marginals $q^{\phi, s}_t$ and $q_t$ using reverse-KL, which smooths out the training objective compared to directly trying to match clean embedding distributions \citep{yin2024one}:
\begin{equation}\label{eq:dmd-objective}
    \mathcal{K}_{s,t}(\phi)
    =
    \KL\big( q^{\phi, s}_t \,\|\, q_t \big) \,,
    \qquad
    \cL^\phi_\text{DMD}
    =
    \E_{\gamma_s, \gamma_t} \big[ \mathcal{K}_{s,t}(\phi) \big] \,.
\end{equation}
Deriving the gradient of $\cL^\phi_\text{DMD}$ \citep{van2026distribution} yields the following gradient surrogate loss:
\begin{equation}\label{eq:dmd-grad}
    \cL^\phi_\text{DMD-grad} = 
    \E_{\gamma_t, \gamma_s, \e^{\phi,s}_t \sim q_t( \e^{\phi,s}_0 \mid \gamma_t )}
    \Big[
    \; \ee^{- \gamma_t} \sum_{l=1}^{L'} \big\langle \e_0^{\phi, s, l}, \sg\big[ \e_\eta^l(\e^{\phi, s}_t, \gamma_t) - \e_\psi^l(\e^{\phi, s}_t, \gamma_t) \big] \big\rangle \Big]\,,
\end{equation}
with \textit{reference denoiser} $\e_\psi$ (linked to the so-called ``real score estimator'' in the literature) and the \textit{auxiliary denoiser} $\e_\eta$ (linked to the ``fake score estimator'' in the literature).
The full derivation is presented in \cref{app:dmd-grad-proof}.
The reference denoiser $\e_\psi$ denoises the re-noised student's samples toward the target distribution $p_\text{teacher}$, while the auxiliary denoiser $\e_\eta$ denoises the re-noised student's samples toward the student distribution. Steering the student with the detached difference of these two denoisers in~\cref{eq:dmd-grad} matches the student and target distributions.
Following \citet{van2026distribution}, we train the categorical auxiliary denoiser with a soft-target cross-entropy against the detached student-produced probabilities:
 \begin{equation}\label{eq:dmd-auxiliary}
     \cL^\eta_\text{aux} = 
     \E_{\gamma_t, \gamma_s, \e^{\phi,s}_t \sim q_t( \e^{\phi,s}_0 \mid \gamma_t )}
\Big[-\sum_{l=1}^{L'}
\big\langle
\sg[\x^{\phi,s,l}]\,
, \, \log \x^l_{\eta}(\sg[\e^{\phi,s}_t],\gamma_t) 
\big\rangle\Big]\,,
\end{equation}
with $\e_\eta = \x_\eta \Emb$\,.
Usually the teacher diffusion model serves directly as $\e_\psi$ since it is trained as a denoiser of the pretraining data distribution $p_\mathrm{data}$.
However, since our target distribution is $p_\text{teacher}$, i.e., teacher-generated samples rather than the pretraining distribution, we re-fit the reference denoiser (initialized from the teacher) to $p_\text{teacher}$ using cross-entropy:
\begin{equation}\label{eq:dmd-reference}
    \cL^\psi_\text{ref} = 
    \E_{\gamma_u \sim \unif[\gamma_0, \gamma_1], \e \sim p_\text{teacher}, \e_u \sim q_t(\e \mid \gamma_u)} 
     \Big[-\sum_{l=1}^{L'}
     \big\langle \x^l ,\, \log\x^l_{\psi}(\e_u,\gamma_u) \big\rangle\Big]\,,
\end{equation}
with $\e = \x \Emb$ and $\e_\psi = \x_\psi \Emb$.

The auxiliary denoiser, the reference denoiser and the main student all use same-state two-call training for self-conditioning \citep{chen2023analog} with probability $p_\text{SC} = 0.25$ while inference carries the previous raw clean prediction, thereby mimicking the original teacher's training and inference behaviors.
The complete training procedure is summarized in \cref{alg:dmd-training}.

\subsection{Implementation Details}
\label{app:dmd-details}

\paragraph{Target distribution.}
The target distribution is a teacher-generated bank
containing 65,536 clean prefixes of 256 tokens, taken from the pretraining dataset,
and four teacher-generated
32-token continuation blocks per prefix. 
This bank is generated with the empirically optimal teacher sampler, i.e.,
256-step DDPM sampling with self-conditioning enabled and optimal guidance: for \model-3B this is CFG $w=2$ and ST $\tau=0.1$, whereas for \model-8B this is CFG $w=3$ and ST $\tau=0.1$.

\paragraph{Parameterization and optimization.}
All three neural networks are initialized from the same respective teacher checkpoint,
with separate trainable parameter copies and optimizers. We copy the embedding matrix, noise schedule and noise endpoints from the teacher and retain them throughout
distillation. 
We detach the source rollout and the SC preliminary
prediction, so student gradients pass only through the selected
denoising transition, similar to \citet{fan2026phased}. 
We prefit the reference denoiser $\e_\psi$ for 50 optimizer updates before starting student/auxiliary updates. Each subsequent student update is preceded by four auxiliary updates
and four reference updates. 
We optimize all networks with AdamW with a global batch size of 256, $\beta_1=0.9$, $\beta_2=0.999$, zero
weight decay, and unit-norm gradient clipping. The learning rates for each network are selected through gradient balancing and are $3.6\times10^{-7}$ for the student,
$1.188\times10^{-5}$ for the auxiliary denoiser, and $10^{-6}$ for the reference denoiser.

In practice, we use the residual-normalization strategy in \citet{yin2024improved} and remove the SNR weight $\ee^{-\gamma_t}$, thereby replacing \cref{eq:dmd-grad} with
\begin{equation}\label{eq:dmd2-grad}
    \cL^\phi_\text{DMD2-grad} = 
    \E_{\gamma_t, \gamma_s, \e^{\phi,s}_t \sim q_t( \e^{\phi,s}_0 \mid \gamma_t )}
    \Big[
    \; \frac{1}{R} \sum_{l=1}^{L'} \big\langle \e_0^{\phi, s, l}, \sg\big[ \e_\eta^l(\e^{\phi, s}_t, \gamma_t) - \e_\psi^l(\e^{\phi, s}_t, \gamma_t) \big] \big\rangle \Big]\,,
\end{equation}
with the student-to-reference denoiser residual
\begin{equation}
R = 10^{-6} + \frac{1}{L' d_e}\sum_{l=1}^{L'}
\sg\big[\big\|\e_0^{\phi,s,l}
-\e^l_\psi(\e^{\phi,s}_t,\gamma_t)\big\|_1\big] \,.
\end{equation}

We further stabilize the auxiliary denoiser training by multiplying its objective in \cref{eq:dmd-auxiliary} with
\begin{equation}
 h_t=\min(1, \, 10 \, \sigma^2(\gamma_t)
    (1-\ee^{\gamma_r-\gamma_t})) \,,
\end{equation}
where $\gamma_r$ is the cleaner endpoint of the
selected segment.

\subsection{Additional Results}
\label{app:dmd-additional-results}

\paragraph{Diversity and pass@$k$.}
We demonstrate here that our distribution-matching distillation procedure preserves the diversity of the target teacher distribution, i.e., we do not fall into the pitfall often encountered in DMD methods.
We use 512 prompts and sample 8 trajectories per prompt, on 3 benchmarks (GSM8K, MBPP and Novelty-Bench).
GSM8K measures math reasoning, MBPP measures coding ability, and Novelty-Bench \citep{zhang2025noveltybenchevaluatinglanguagemodels} is an open-ended creative writing benchmark which helps us measure actual text diversity and creativity.

\begin{figure}[!t]
    \centering
    \includegraphics[width=0.8\linewidth]{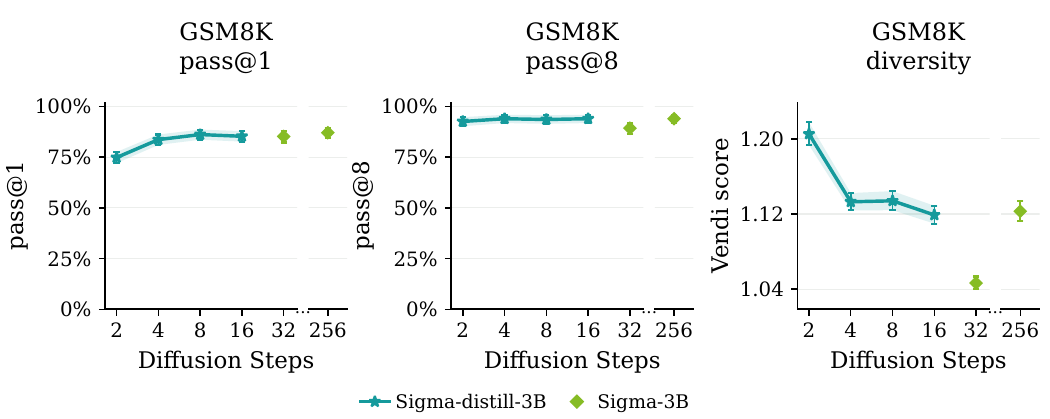}
    \caption{Diversity and pass@$k$ results on GSM8K 512-prompt subset.}
    \label{fig:diversity-gsm8k}
\end{figure}

\begin{figure}[!t]
    \centering
    \includegraphics[width=0.8\linewidth]{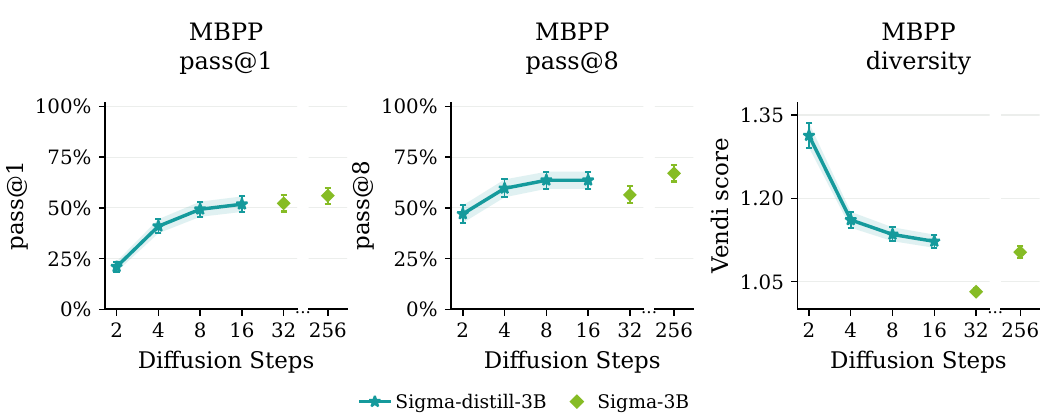}
    \caption{Diversity and pass@$k$ results on MBPP 500-prompt test set.}
    \label{fig:diversity-mbpp}
\end{figure}

\begin{figure}[!t]
    \centering
    \includegraphics[width=0.8\linewidth]{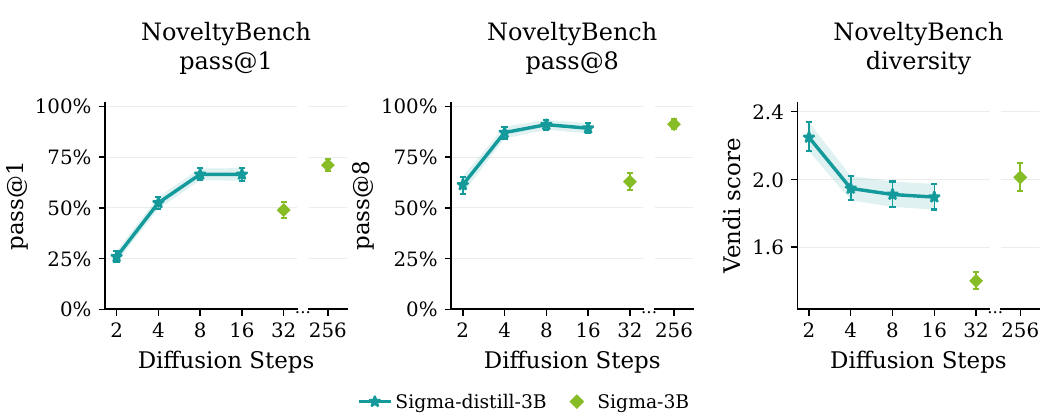}
    \caption{Diversity and pass@$k$ results on NoveltyBench and NoveltyBench-WildCat 512-prompt test set.}
    \label{fig:diversity-noveltybench}
\end{figure}

We measure pass@$1$ and pass@$8$ accuracies as well as semantic diversity using the Vendi score \citep{friedman2023vendi}. We compute the Vendi score as the exponentiated Jensen-Shannon entropy of the pairwise similarity matrix between SentenceBERT embeddings of the produced outputs.
Results displayed in
\cref{fig:diversity-gsm8k,fig:diversity-mbpp,fig:diversity-noveltybench}
clearly demonstrate that our DMD student recovers the DDPM-256 Vendi score as well as its pass@$8$ accuracy, with as few as 4 diffusion steps. Furthermore, the DDIM-32 \model-3B (with self-conditioning) teacher has lower diversity and pass@$8$ accuracy than both our \model\distill-3B students and the DDPM-256 \model-3B, although its pass@$1$ is somewhat comparable to our 4-NFE \model\distill-3B.
This suggests that our distilled students balance quality and diversity for only a fraction of the inference cost, which is very promising for, e.g., RL-based post-training.


\paragraph{Ablation 1: Teacher.} 

\begin{figure}[!t]
    \centering
    \includegraphics[width=0.8\linewidth]{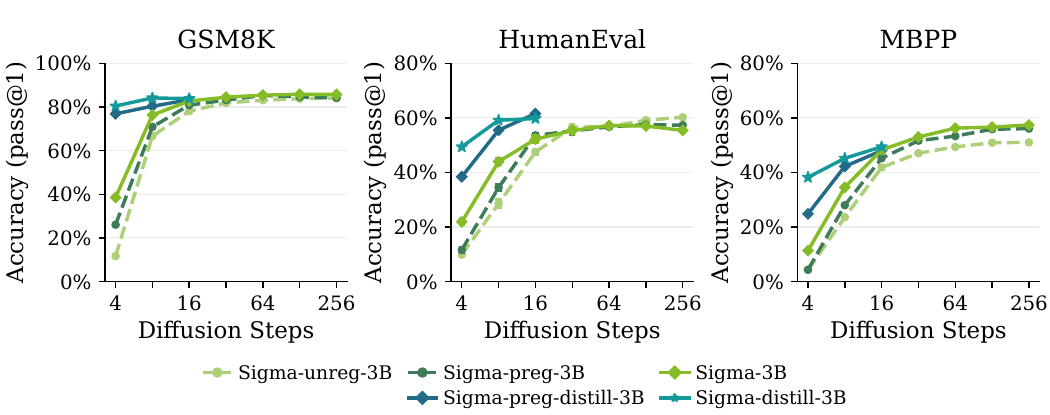}
    \caption{Teacher ablation for DMD distillation at 3B scale. Global batch size 64.}
    \label{fig:distill-teacher}
\end{figure}

We investigate the influence of the teacher's geometry on the distillation performance. We observed earlier that improving the regularity of the geometry is key for low-NFE performance. Indeed, the unregularized \model\unreg-3B teacher exhibits local embedding collapse, lowering the global rank participation and creating uncertainty during decoding, which particularly affects coding tasks (see \cref{sec:few-step-susreg,tab:acc-nfe-3b}).
This consequence transfers naturally to the distillation quality, as we notice in \cref{fig:distill-teacher} that \model-3B teaches much stronger students at 4- and 8-NFE in particular.

\paragraph{Ablation 2: Global Batch Size.}

We ablate the effect of global batch size on distillation quality. Increasing the global batch size from 64 to 256 yields large and consistent improvements across NFEs. Increasing batch size above 256 did not further improve results.

\begin{figure}[!t]
    \centering
    \includegraphics[width=0.8\linewidth]{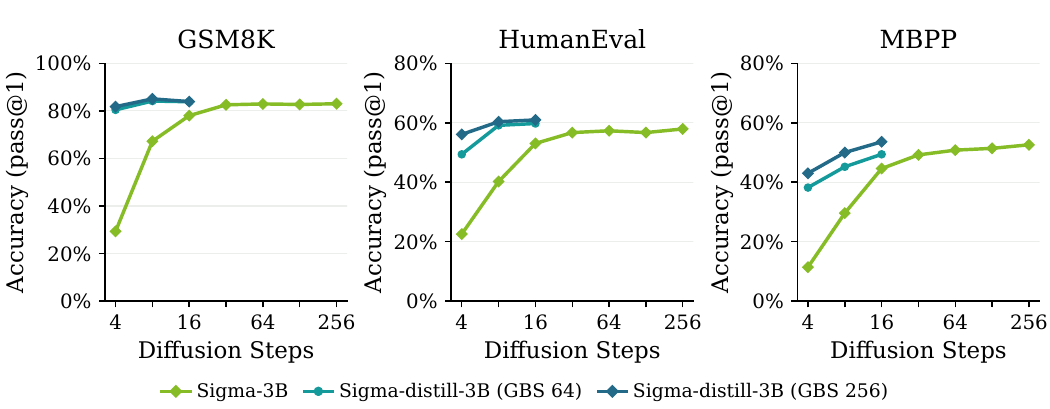}
    \caption{Global batch size ablation for DMD distillation from \model-3B teacher.}
    \label{fig:distill-gbs}
\end{figure}

\paragraph{Ablation 3: Query Distributions.}

\begin{figure}[!t]
    \centering
    \includegraphics[width=0.8\linewidth]{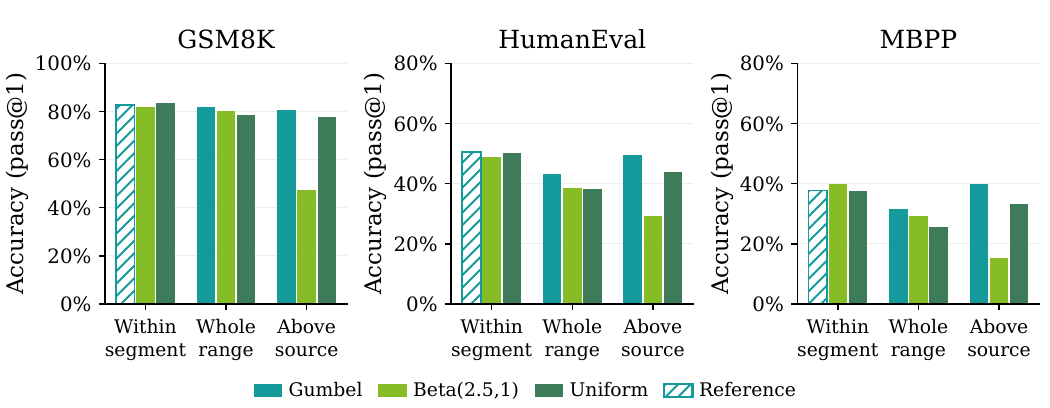}
    \caption{Query distribution ablation for DMD distillation from \model-3B teacher. Global batch size 64. $M=4$ NFE distillation.}
    \label{fig:distill-query}
\end{figure}

Choosing the query distribution is a key design choice in DMD distillation. After obtaining very good early results with Phased DMD \citep{fan2026phased}, we ablated the discrepancies between Phased DMD and plain DMD, and noticed that the source states, the transitions and the source and query distributions were the most discriminative factors. As we kept the consistency-style transitions with fresh Gaussian noise and the student rollout for obtaining our input states, the main remaining difference was the query distribution. 
We choose three candidate distributions: Uniform, Gumbel fitted to the teacher's empirical denoising cross-entropy distribution (skewed toward intermediate noise levels), and $\beta(2.5, 1)$ (skewed toward high noise levels); as well as three ranges for the distribution support: the whole range $[\gamma_0, \gamma_1]$, within segment $[\gamma_r, \gamma_t]$, and above segment $[\gamma_t, \gamma_1]$.
We conduct this ablation specifically on the 4-NFE distillation setup.
In \cref{fig:distill-query}, we observe that using the current segment as the support consistently obtained the best results, and that the exact distribution did not matter so much there. The teacher-CE-fit Gumbel above segment obtained excellent results too, but we noticed that the resulting distillation procedure was very unstable.

\paragraph{Ablation 4: Self-conditioning and student input.}

\begin{figure}[!t]
    \centering
    \includegraphics[width=0.8\linewidth]{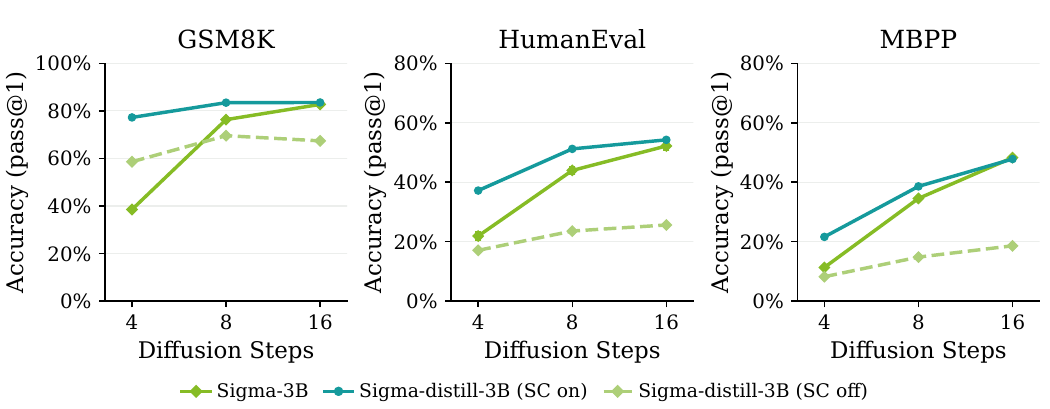}
    \caption{Self-conditioning ablation for DMD distillation from \model-3B teacher. Global batch size 64.}
    \label{fig:distill-sc}
\end{figure}

Using self-conditioning in the student is absolutely key to its generation quality.
The self-conditioning probability used during distillation hardly influences the result: for simplicity we retain $p_\text{SC}=0.25$.
Using $p_\text{SC}=0$ degraded results slightly, but far worse is deactivating self-conditioning at inference, as shown in \cref{fig:distill-sc}.

\paragraph{Ablation 5: Training data.}

\begin{figure}[!t]
    \centering
    \includegraphics[width=0.8\linewidth]{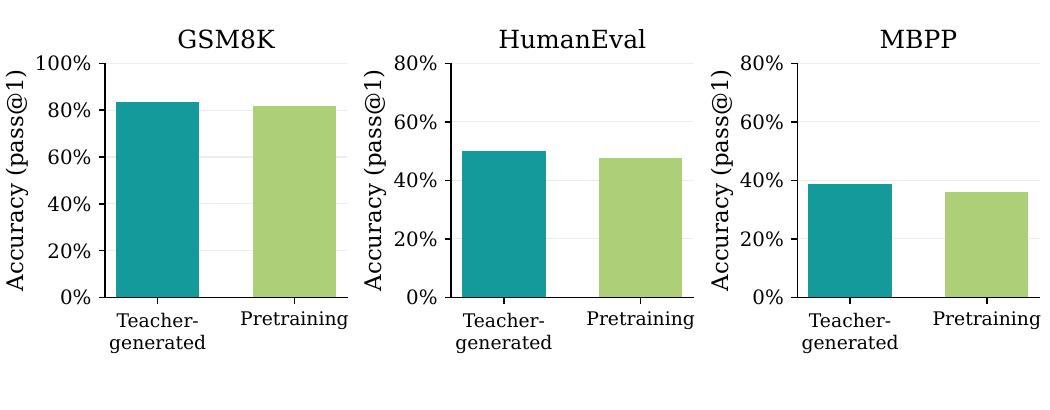}
    \caption{Data distribution ablation for DMD distillation from \model-3B teacher. Global batch size 64. $M=4$ NFE distillation.}
    \label{fig:distill-data}
\end{figure}

We observe in \cref{fig:distill-data} that teacher-generated data led to better results than directly using the pretraining dataset. We posit this is due to the inductive bias carried by the student about the target distribution, since it is initialized from the teacher that produced this distribution.
We conduct this ablation specifically on the 4-NFE distillation setup.

\paragraph{Ablation 6: Student and auxiliary objectives.}

\begin{figure}[!t]
    \centering
    \includegraphics[width=0.8\linewidth]{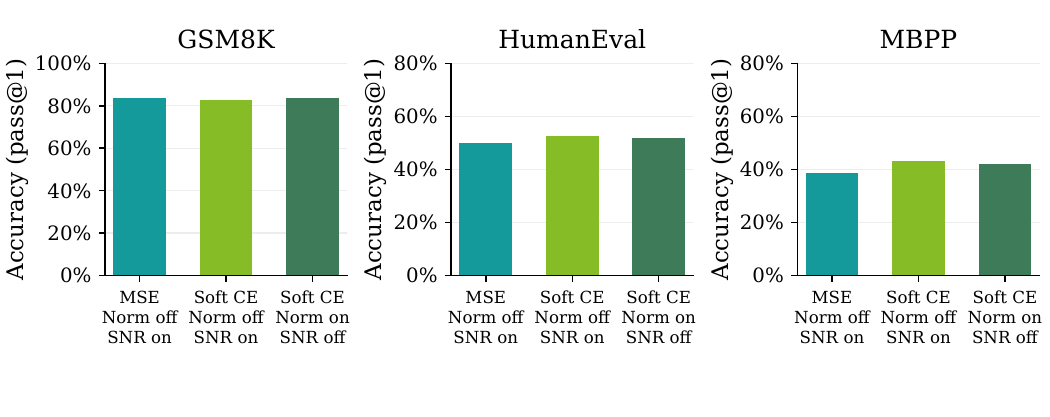}
    \caption{Student objective ablation for DMD distillation from \model-3B teacher. Global batch size 64. $M=4$ NFE distillation.}
    \label{fig:distill-student-objective}
\end{figure}

We ablate the choice of student loss weighting (``Norm off SNR on'' \cref{eq:dmd-grad} vs. ``Norm on SNR off'' \cref{eq:dmd2-grad}) and the auxiliary denoiser parameterization and objective.
``Soft CE'' indicates our baseline, i.e., a categorical denoiser (similar to the teacher) trained with cross-entropy on soft probabilities. ``MSE'' indicates a denoiser augmented with an MLP-parameterized residual connection from Transformer final hidden state to the re-embedded categorical prediction, trained with MSE in the embedding space directly.
We conduct this ablation specifically on the 4-NFE distillation setup.
The results highlighted in \cref{fig:distill-student-objective} favor the Soft CE auxiliary objective introduced in \citet{van2026distribution}; the DMD2-type normalization \citep{yin2024improved} does not play a major role, although we informally observed it led to faster and more stable convergence, therefore being favored in our final setup.

\subsection{Proof: DMD Gradient Surrogate Loss}
\label{app:dmd-grad-proof}

In this section, we prove the form of the DMD gradient surrogate loss in \cref{eq:dmd-grad}.
\begin{proof}
We first fix the source and query noise levels $(\gamma_s,\gamma_t)$.
All distributions below are implicitly conditioned on the clean prefix.
For this derivation, we hold the source-state sampling distribution
fixed, consistently with detaching the on-policy source rollout.

Write $\alpha_t=\alpha(\gamma_t)$ and
$\sigma_t=\sigma(\gamma_t)$.
The Gaussian corruption admits the reparameterization
\begin{equation}
    \e_t^{\phi,s}
    =
    \alpha_t \e_0^{\phi,s} + \sigma_t \beps,
    \qquad
    \e_0^{\phi,s} = \e_\phi(\e_s,\gamma_s),
    \qquad
    \beps \sim \mathcal{N}(\zero,\I).
\end{equation}
Our log-noise-to-signal parameterization satisfies
\begin{equation}
    \gamma_t
    =
    \log \frac{\sigma_t^2}{\alpha_t^2},
    \qquad
    \frac{\alpha_t^2}{\sigma_t^2}
    =
    \ee^{-\gamma_t}.
\end{equation}

For fixed $(\gamma_s,\gamma_t)$, the per-level objective of \cref{eq:dmd-objective} reads
\begin{equation}
    \mathcal{K}_{s,t}(\phi)
    =
    \KL\big(q_t^{\phi,s}\,\|\,q_t\big)
    =
    \E_{\e_s,\beps}
    \Big[
        \log q_t^{\phi,s}(\e_t^{\phi,s})
        -
        \log q_t(\e_t^{\phi,s})
    \Big].
\end{equation}
There are two contributions when differentiating the student
log-density: its explicit dependence on $\phi$ at a fixed input,
and its dependence through the sampled input $\e_t^{\phi,s}$.
Consequently,
\begin{align}
    \nabla_\phi \mathcal{K}_{s,t}
    &=
    \E_{\e_s,\beps}
    \Big[
        \nabla_\phi \log q_t^{\phi,s}(\e)
        \Big|_{\e=\e_t^{\phi,s}}
        +
        \Big(
            \frac{\partial \e_t^{\phi,s}}{\partial\phi}
        \Big)^{\!\top}
        \Big(
            \nabla_{\e}\log q_t^{\phi,s}(\e_t^{\phi,s})
            -
            \nabla_{\e}\log q_t(\e_t^{\phi,s})
        \Big)
    \Big].
\end{align}
The expectation of the explicit density derivative vanishes:
\begin{align}
    \E_{\e\sim q_t^{\phi,s}}
    \Big[
        \nabla_\phi\log q_t^{\phi,s}(\e)\Big|_{\e\ \mathrm{fixed}}
    \Big]
    &=
    \int \nabla_\phi q_t^{\phi,s}(\e)\,d\e
    =
    \nabla_\phi \int q_t^{\phi,s}(\e)\,d\e
    =
    0.
\end{align}
Thus, introducing the student and target scores
\begin{equation}
    \mathbf{s}_{\mathrm{stu}}^{\phi,s}(\e,\gamma_t)
    :=
    \nabla_{\e}\log q_t^{\phi,s}(\e),
    \qquad
    \mathbf{s}_{\mathrm{ref}}(\e,\gamma_t)
    :=
    \nabla_{\e}\log q_t(\e),
\end{equation}
we obtain
\begin{equation}
    \nabla_\phi \mathcal{K}_{s,t}
    =
    \E_{\e_s,\beps}
    \Big[
        \Big(
            \frac{\partial \e_t^{\phi,s}}{\partial\phi}
        \Big)^{\!\top}
        \Big(
            \mathbf{s}_{\mathrm{stu}}^{\phi,s}
                (\e_t^{\phi,s},\gamma_t)
            -
            \mathbf{s}_{\mathrm{ref}}
                (\e_t^{\phi,s},\gamma_t)
        \Big)
    \Big].
\end{equation}

For Gaussian corruption, the score is related to the posterior
mean of the clean embedding by
\begin{equation}
    \nabla_{\e_t}\log q_t(\e_t)
    =
    \frac{
        \alpha_t\,\E[\e_0\mid\e_t]
        -
        \e_t
    }{\sigma_t^2}.
\end{equation}
Let the ideal auxiliary and reference denoisers be
\begin{equation}
    \e_\eta^\star(\e_t,\gamma_t)
    :=
    \E_{\mathrm{stu}}[
        \e_0^{\phi,s}\mid\e_t,\gamma_s,\gamma_t
    ],
    \qquad
    \e_\psi^\star(\e_t,\gamma_t)
    :=
    \E_{\mathrm{ref}}[
        \e_0\mid\e_t,\gamma_t
    ],
\end{equation}
where source conditioning of the ideal auxiliary is suppressed
in its arguments.
The corresponding score difference is therefore
\begin{align}
    \mathbf{s}_{\mathrm{stu}}^{\phi,s}(\e_t,\gamma_t)
    -
    \mathbf{s}_{\mathrm{ref}}(\e_t,\gamma_t)
    &=
    \frac{
        \alpha_t\e_\eta^\star(\e_t,\gamma_t)-\e_t
    }{\sigma_t^2}
    -
    \frac{
        \alpha_t\e_\psi^\star(\e_t,\gamma_t)-\e_t
    }{\sigma_t^2}
    \nonumber\\
    &=
    \frac{\alpha_t}{\sigma_t^2}
    \Big[
        \e_\eta^\star(\e_t,\gamma_t)
        -
        \e_\psi^\star(\e_t,\gamma_t)
    \Big].
\end{align}
Meanwhile, because the corruption noise and detached source are
held fixed under differentiation,
\begin{equation}
    \frac{\partial\e_t^{\phi,s}}{\partial\phi}
    =
    \alpha_t
    \frac{\partial\e_0^{\phi,s}}{\partial\phi}.
\end{equation}
Combining these identities gives
\begin{align}
    \nabla_\phi \mathcal{K}_{s,t}
    &=
    \E_{\e_s,\beps}
    \Big[
        \frac{\alpha_t^2}{\sigma_t^2}
        \Big(
            \frac{\partial\e_0^{\phi,s}}{\partial\phi}
        \Big)^{\!\top}
        \Big(
            \e_\eta^\star(\e_t^{\phi,s},\gamma_t)
            -
            \e_\psi^\star(\e_t^{\phi,s},\gamma_t)
        \Big)
    \Big]
    \nonumber\\
    &=
    \E_{\e_s,\beps}
    \Big[
        \ee^{-\gamma_t}
        \Big(
            \frac{\partial\e_0^{\phi,s}}{\partial\phi}
        \Big)^{\!\top}
        \Big(
            \e_\eta^\star(\e_t^{\phi,s},\gamma_t)
            -
            \e_\psi^\star(\e_t^{\phi,s},\gamma_t)
        \Big)
    \Big].
\end{align}

Replacing the ideal posterior means with learned denoisers and
averaging over source and query levels as in \cref{eq:dmd-objective} yields
\begin{equation}
    \nabla_\phi\cL^\phi_\text{DMD}
    =
    \E_{\gamma_s,\gamma_t,\e_s,\beps}
    \Big[
        \ee^{-\gamma_t}
        \Big(
            \frac{\partial\e_0^{\phi,s}}{\partial\phi}
        \Big)^{\!\top}
        \Big(
            \e_\eta(\e_t^{\phi,s},\gamma_t)
            -
            \e_\psi(\e_t^{\phi,s},\gamma_t)
        \Big)
    \Big].
\end{equation}
The Jacobian--vector product above contracts over both the $L'$ token
positions of the block and the $d_e$ embedding coordinates, so writing it
as a sum of per-token inner products shows that this gradient is obtained
by differentiating the scalar surrogate
\begin{equation}\label{eq:dmd-grad-proof}
    \cL^\phi_\text{DMD-grad}
    =
    \E_{\gamma_s,\gamma_t,\e_s,\beps}
    \Big[
        \ee^{-\gamma_t}
        \sum_{l=1}^{L'}
        \Big\langle
            \e_0^{\phi,s,l},
            \sg\big[
                \e_\eta^l(\e_t^{\phi,s},\gamma_t)
                -
                \e_\psi^l(\e_t^{\phi,s},\gamma_t)
            \big]
        \Big\rangle
    \Big] \,,
\end{equation}
which is \cref{eq:dmd-grad} and concludes the proof.
\end{proof}

\clearpage

\section{Implementation Details and Additional Results for Supervised Fine-Tuning}
\label{app:sft}

[Back to~\cref{sec:sft} in the main paper.]

\subsection{Supervised Fine-Tuning Setup}
\label{app:sft-setup}

Our SFT pipeline uses NeMo-RL with joint diffusion and AR supervision on assistant responses, following Nemotron-Labs-Diffusion (NLD)~\citep{fu2026nemotron}. Prompts remain clean, and training and evaluation share the non-thinking chat boundary described below.
We freeze the diffusion embedding $\Emb$, noise-schedule network $\tilde\gamma_\psi(t)$, and endpoints $\gamma_0,\gamma_1$. The clean-stream input embedding, transformer, conditioning projections, and both readout heads remain trainable. Therefore, SUSReg is used only during pretraining.

\paragraph{Initialization and objective.}
We initialize from the 150k-step pretrained \model\unreg-3B/8B, \model\preg-3B/8B, and \model-3B/8B checkpoints.
Before SFT of \model\preg-8B and \model-8B, we reinitialize the AR embedding rows with $\ell_2$ norm below $10^{-6}$ to the mean direction of the remaining rows, scaled to their median norm. This includes chat-boundary IDs 10 and 11 and prevents the initial gradient-norm transient caused by their near-zero embeddings. The repair is applied once; its benchmark effect is not separately ablated.
Let $\mathcal{A}$ denote the supervised answer positions, including the terminal EOS padding described below. The SFT objective is
\begin{align}
    \cL_{\text{sft}}
    = \alpha\,\frac{\sum_{i\in\mathcal{A}}\cL_{\text{BD},i}}{|\mathcal{A}|}
    + \frac{\sum_{i\in\mathcal{A}}\cL_{\text{ar},i}}{|\mathcal{A}|},
    \label{eq:sft-objective}
\end{align}
where $\cL_{\text{BD}}$ is the continuous block NELBO of \cref{eq:block-nelbo,eq:nelbo}, including its diffusion, reconstruction, and prior terms, and $\cL_{\text{ar}}$ is next-token cross-entropy from the clean stream \cref{eq:ar_loss}. The ELBO weight is $\alpha=0.3$ for \model\unreg-3B/8B and \model\preg-3B/8B, and $\alpha=1$ for \model-3B/8B, following pretraining (\cref{app:pretrain}). Numerators and token counts are reduced globally across data-parallel ranks. The AR and diffusion denominators coincide under the answer mask because position zero is unsupervised, so $\alpha$ has the same meaning as the pretraining diffusion-loss weight.

As in pretraining (\cref{sec:pretrain}), we estimate the reconstruction and diffusion terms with the Horvitz--Thompson estimator in~\cref{eq:ht}: independently for each block, we select reconstruction at $t=0$ with probability $p_{\mathrm{rec}}=0.1$ and otherwise draw $t\sim\unif[0,1]$.  Gaussian noise is applied only in embedding space; the token IDs always remain clean. Prompt and ordinary batch-padding positions have zero loss and are pinned to their clean embeddings, whereas the synthetic EOS padding is supervised. Answer blocks see the clean context $\c = (\p, \r^{(\bcolor{<b})})$ of~\cref{sec:inference-solver}.

\paragraph{Data and preprocessing.}
We use the same non-thinking, single-turn SFT corpus as NLD~\citep{chandiramani2026nemotron}: $848{,}087$ training and $1{,}200$ validation conversations, each with an empty system message, a user prompt, and an assistant response.
We use the same Mistral-NeMo~\citep{mistral_nemo_2024} tokenizer ($V=131{,}072$) and the Nemotron Nano chat template \citep{blakeman2025nemotron}, explicitly setting \texttt{enable\_thinking=false}. Thus, the supervised response follows the intended prompt suffix \texttt{<think></think>}; training and evaluation use the same answer boundary.
During training, we left-pad prompts with at most $31$ unsupervised tokens so answers begin on the $32$-token block grid; no padding is inserted between prompt and answer. Inference uses unpadded, fully causal prompts. Supervised EOS tokens complete the final answer block, with one full EOS block appended when already aligned. Unsupervised right-padding then rounds the batch length to a multiple of $1024$ to limit FlexAttention compilation shapes. Samples exceeding $8192$ tokens after padding are dropped.

\paragraph{Optimization.}
Both model sizes train for $8000$ steps ($\approx2.4$ epochs) at sequence length and maximum position $8192$, block length $32$, and global/micro batch sizes $256/1$. We use AdamW with $(\beta_1,\beta_2)=(0.9,0.98)$, $\epsilon=10^{-5}$, weight decay $0.1$, and learning rate $2.5\times10^{-6}$ after $50$ steps of linear warmup, held constant thereafter. Training uses \texttt{BF16}, gradient clipping at $1.0$, and FSDP2 with TP${}={}$CP${}={}$1, activation checkpointing, and no sequence packing. Self-conditioning is enabled for $25\%$ of microbatches on every optimizer step and all validation sequences. The chat template enables the generation prompt; clean-prompt, answer-only supervision and EOS padding follow the preprocessing above.

Setting the maximum position to $8192$ also extends the pretrained rotary frequencies: the effective YaRN factor is $8192/16384=0.5$, compared with $0.25$ at the pretraining length of $4096$.

\subsection{Evaluation Setup, CFG--ST Sweep, and Full Results}
\label{app:sft-eval}

The instruct evaluations in~\cref{tab:sft-3b,tab:sft-8b} use the base-model harness (\cref{app:inference-eval-setup}), with the differences below.

\paragraph{Benchmarks.}
We evaluate GSM8K (1319 problems), MATH-500 (500), AIME 2024/2025 (30 each), HumanEval+ (164), MBPP (378), LiveCodeBench-CPP (454), MMLU (14042), GPQA (198), and IFEval (541), using the SFT chat template and diffusion block length $L'=32$.

\paragraph{Scoring.}
Math tasks use symbolic equivalence on the final \texttt{\textbackslash boxed\{\}} answer; GSM8K also accepts the last number. HumanEval and MBPP report executed pass@$1$ on EvalPlus’s base tests (\texttt{passing\_base\_tests}), and LiveCodeBench-CPP compiles and tests extracted C++ programs. GPQA and MMLU request reasoning ending in \texttt{Answer: \textbackslash boxed\{A/B/C/D\}}; grading extracts the last boxed letter, with a regex fallback. IFEval reports instruction-level loose accuracy, matching the reference models in~\cref{tab:sft-3b,tab:sft-8b}.

\paragraph{Baselines.}
Baseline scores come from~\citet[Tabs.~5, 7]{fu2026nemotron}, except NLD-8B diffusion, which we reproduce under the aligned protocol without format injection. We use SDAR-8B Chat at TPF~$=1.00$ rather than $1.75$: it scores higher on eight tasks, with gaps $\leq0.6$ points on the remaining two (MMLU and HumanEval).

\paragraph{Decoding algorithms.}
As in the evaluation for base models, we always enable SC; DDPM-256 uses EDM $\rho=7$ schedule, and DDIM-32 uses uniform-$\gamma$ schedule. Their respective $(w,\tau)$ settings follow the recommended values for base models \cref{tab:acc-nfe-3b}.

\paragraph{Generation budget.}
The generation budget is $1024$ tokens for GSM8K and HumanEval at 3B, rising to $1536$ and $2560$, respectively, at 8B; $2560$ for MBPP, LiveCodeBench-CPP, and IFEval; and $4096$ for AIME 2024/2025, MATH-500, and GPQA. MMLU uses $2048$ for 3B DDPM and $2560$ otherwise. AR uses $16384$ except on MMLU ($2560$). Caps follow per-lane graded-answer lengths: no correct answer in the smoke runs required more than $4096$ tokens, and the longer tail consists mainly of degenerate repetition.

\paragraph{Seed error bars.}
Each diffusion row averages $10$ independent sampling lanes and reports half-width $\frac{2}{\sqrt{10}}\hat\sigma$, where $\hat\sigma$ is the standard deviation of per-lane benchmark scores. AR greedily decodes the clean stream through its AR head.

\paragraph{Post-SFT CFG--ST sweep.}
On GSM8K and HumanEval, the pre-SFT guidance and temperature optima remain within sampling uncertainty of the best post-SFT settings for both 3B checkpoints and samplers (\cref{tab:sft-cfg-st-sweep}). The 8B SUSReg checkpoint uses the corresponding 3B settings without a separate sweep.

\begin{table}[h]
    \centering
    \caption{\textbf{Post-SFT CFG--ST sweep and transfer of pre-SFT optima.} Entries are the unweighted GSM8K--HumanEval average (\%), reported as mean $\pm$ two standard errors over $10$ seeds. DDPM-256 uses EDM $\rho=7$; DDIM-32 uses uniform-$\gamma$. $\dagger$ marks the corresponding pre-SFT optimum; \textbf{bold} marks the maximum in each model--sampler grid. The pre-SFT settings remain within sampling uncertainty of the post-SFT maxima.}
    \label{tab:sft-cfg-st-sweep}
    \setlength{\tabcolsep}{4.5pt}
    \resizebox{\linewidth}{!}{%
    \begin{tabular}{ccccccc}
    \toprule
    Model and sampler & $w$ & $\tau=0.1$ & $\tau=0.2$ & $\tau=0.3$ & $\tau=0.4$ & $\tau=0.5$ \\
    \midrule
    \model\unreg-3B, DDPM-256 & 1 & \se{81.75}{0.56} & --- & --- & --- & --- \\
    \model\unreg-3B, DDPM-256 & 2 & \se{\textbf{83.32}$^\dagger$}{0.52} & --- & --- & --- & --- \\
    \model\unreg-3B, DDPM-256 & 3 & \se{80.63}{0.42} & --- & --- & --- & --- \\
    \midrule
    \model\unreg-3B, DDIM-32 & 2 & --- & \se{80.93}{0.44} & \se{79.83}{0.76} & \se{78.84}{0.60} & \se{78.65}{0.85} \\
    \model\unreg-3B, DDIM-32 & 3 & --- & \se{60.89}{0.40} & \se{80.85$^\dagger$}{0.60} & \se{81.17}{0.57} & \se{81.32}{0.70} \\
    \model\unreg-3B, DDIM-32 & 4 & --- & \se{38.88}{0.59} & \se{77.70}{0.99} & \se{\textbf{81.68}}{0.62} & \se{81.21}{0.84} \\
    \model\unreg-3B, DDIM-32 & 5 & --- & \se{21.21}{0.26} & \se{69.86}{0.81} & \se{80.94}{0.72} & \se{81.05}{0.78} \\
    \midrule
    \model\preg-3B, DDPM-256 & 1 & --- & \se{80.10}{0.56} & --- & --- & --- \\
    \model\preg-3B, DDPM-256 & 2 & --- & \se{80.75}{0.50} & --- & --- & --- \\
    \model\preg-3B, DDPM-256 & 3 & --- & \se{\textbf{80.80}$^\dagger$}{0.57} & --- & --- & --- \\
    \model\preg-3B, DDPM-256 & 4 & --- & \se{79.75}{0.69} & --- & --- & --- \\
    \midrule
    \model\preg-3B, DDIM-32 & 2 & --- & \se{\textbf{79.17}}{0.55} & \se{78.76}{0.90} & \se{78.03}{0.56} & \se{77.63}{0.84} \\
    \model\preg-3B, DDIM-32 & 3 & --- & \se{73.81}{0.66} & \se{78.29}{0.56} & \se{78.70}{1.08} & \se{78.59}{0.42} \\
    \model\preg-3B, DDIM-32 & 4 & --- & \se{54.50}{0.65} & \se{77.93}{0.65} & \se{78.65$^\dagger$}{0.55} & \se{78.32}{0.60} \\
    \model\preg-3B, DDIM-32 & 5 & --- & \se{43.24}{0.32} & \se{74.94}{0.81} & \se{79.09}{0.89} & \se{78.97}{0.51} \\
    \bottomrule
    \end{tabular}%
    }
\end{table}

\paragraph{Full instruct results for all three variants.}
\Cref{tab:sft-variants} reports the ten instruct benchmarks for all three variants at 3B and 8B.

\begin{table}[h]
    \centering
    \caption{\textbf{Pass@$1$ accuracy of all three \model\instruct variants at 3B and 8B.} Same ten tasks, harness, and operating points as \cref{tab:sft-3b,tab:sft-8b}. \model diffusion rows are mean (over $10$ sampling lanes) $\pm$ two standard errors. \textbf{Bold} marks the best value per task within each mode and model size.}
    \label{tab:sft-variants}
    \resizebox{\linewidth}{!}{%
    \begin{tabular}{lccccccccccc}
    \toprule
    & \multicolumn{4}{c}{\textbf{Math}} & \multicolumn{3}{c}{\textbf{Coding}} & \multicolumn{2}{c}{\textbf{Science \& Knowledge}} & \textbf{IF} & \\
    \cmidrule(lr){2-5} \cmidrule(lr){6-8} \cmidrule(lr){9-10} \cmidrule(lr){11-11}
    Method & GSM8K & Math500 & AIME24 & AIME25 & HumanEval & MBPP & LCB-CPP & \makebox[4.2em]{MMLU} & \makebox[4.2em]{GPQA} & IFEval & Avg. \\
    \midrule
    \multicolumn{12}{l}{\textbf{3B models}} \\
    \multicolumn{12}{l}{\emph{AR mode}} \\
    \quad Nemotron-Labs-Diffusion-3B-Instruct & 87.87 & \B{77.60} & \B{23.33} & \B{16.67} & 76.22 & 71.16 & \B{21.37} & 71.70 & 39.39 & 69.39 & \B{55.47} \\
    \quad \model\unreg-3B\instruct & \B{88.17} & 71.60 & 20.00 & 10.00 & 71.34 & 77.51 & 13.00 & 71.36 & 38.38 & \B{73.62} & 53.50 \\
    \quad \model\preg-3B\instruct & 86.81 & 74.60 & 10.00 & 13.33 & \B{79.88} & 77.78 & 15.86 & \B{72.87} & \B{39.90} & \B{73.62} & 54.46 \\
    \quad \model-3B\instruct & 85.29 & 71.60 & 3.33 & 3.33 & 73.17 & \B{78.04} & 14.76 & 69.98 & 34.34 & 70.74 & 50.46 \\
    \cmidrule(lr){1-12}
    \multicolumn{12}{l}{\emph{Diffusion mode}} \\
    \quad Nemotron-Labs-Diffusion-3B-Instruct & 88.40 & \B{74.80} & \B{16.67} & 10.00 & 74.39 & 73.54 & \B{16.30} & 72.06 & 33.84 & 68.93 & 52.89 \\
    \quad \model\unreg-3B\instruct & & & & & & & & & & & \\
    \qquad DDPM ($256$ steps, $w{=}2,\tau{=}0.1$) & \sev{89.32}{0.24} & \sev{73.20}{0.45} & \sev{15.67}{2.84} & \sev{11.00}{1.90} & \sev{\B{77.32}}{1.05} & \sev{71.75}{0.72} & \sev{12.97}{0.62} & \sev{\B{72.58}}{0.08} & \sev{\B{37.68}}{1.70} & \sev{70.11}{0.64} & 53.16 \\
    \qquad DDIM ($32$ steps, $w{=}3,\tau{=}0.3$) & \sev{88.10}{0.24} & \sev{71.42}{0.88} & \sev{14.00}{2.46} & \sev{8.67}{1.93} & \sev{73.60}{1.17} & \sev{68.76}{0.69} & \sev{11.56}{0.61} & \sev{71.37}{0.08} & \sev{34.24}{1.42} & \sev{68.18}{0.32} & 50.99 \\
    \quad \model\preg-3B\instruct & & & & & & & & & & & \\
    \qquad DDPM ($256$ steps, $w{=}3,\tau{=}0.2$) & \sev{89.46}{0.16} & \sev{72.96}{0.99} & \sev{13.67}{1.48} & \sev{8.67}{1.40} & \sev{72.13}{1.09} & \sev{71.80}{0.52} & \sev{12.58}{0.43} & \sev{72.20}{0.07} & \sev{37.17}{1.49} & \sev{72.49}{0.45} & 52.31 \\
    \qquad DDIM ($32$ steps, $w{=}4,\tau{=}0.4$) & \sev{88.95}{0.19} & \sev{70.88}{0.47} & \sev{13.00}{1.48} & \sev{\B{14.33}}{2.32} & \sev{68.35}{1.10} & \sev{68.54}{0.63} & \sev{11.30}{0.37} & \sev{71.14}{0.09} & \sev{36.31}{1.68} & \sev{70.96}{0.77} & 51.38 \\
    \quad \model-3B\instruct & & & & & & & & & & & \\
    \qquad DDPM ($256$ steps, $w{=}3,\tau{=}0.1$) & \sev{\B{89.64}}{0.10} & \sev{73.08}{0.62} & \sev{13.67}{1.99} & \sev{11.33}{1.03} & \sev{77.01}{1.24} & \sev{\B{74.97}}{0.59} & \sev{13.50}{0.42} & \sev{70.64}{0.12} & \sev{36.01}{1.46} & \sev{\B{73.26}}{0.59} & \B{53.31} \\
    \qquad DDIM ($32$ steps, $w{=}3,\tau{=}0.3$) & \sev{88.82}{0.15} & \sev{72.10}{0.74} & \sev{15.33}{2.70} & \sev{13.33}{2.11} & \sev{73.96}{0.91} & \sev{73.31}{0.82} & \sev{12.16}{0.56} & \sev{70.20}{0.07} & \sev{34.80}{1.99} & \sev{69.00}{0.49} & 52.30 \\
    \midrule
    \multicolumn{12}{l}{\textbf{8B models}} \\
    \multicolumn{12}{l}{\emph{AR mode}} \\
    \quad Nemotron-Labs-Diffusion-8B-Instruct & \B{94.01} & \B{88.00} & \B{33.33} & \B{33.33} & 80.49 & \B{85.19} & \B{28.85} & \B{79.85} & 44.44 & 68.65 & \B{63.61} \\
    \quad \model\unreg-8B\instruct & 91.74 & 80.80 & 16.67 & 20.00 & 78.05 & 84.92 & 21.59 & 78.22 & \B{46.97} & 79.50 & 59.85 \\
    \quad \model\preg-8B\instruct & 92.57 & 78.60 & 16.67 & 6.67 & 79.88 & 83.07 & 15.20 & 78.32 & 44.44 & \B{80.70} & 57.61 \\
    \quad \model-8B\instruct & 91.58 & 79.40 & 26.67 & 16.67 & \B{81.10} & 80.69 & 12.33 & 76.27 & 43.94 & 78.06 & 58.67 \\
    \cmidrule(lr){1-12}
    \multicolumn{12}{l}{\emph{Diffusion mode}} \\
    \quad Nemotron-Labs-Diffusion-8B-Instruct & 91.58 & \B{84.60} & \B{30.00} & \B{23.33} & \B{83.54} & 82.01 & \B{21.15} & 76.65 & 33.33 & 73.02 & \B{59.92} \\
    \quad \model\unreg-8B\instruct & & & & & & & & & & & \\
    \qquad DDPM ($256$ steps, $w{=}2,\tau{=}0.1$) & \sev{\B{93.24}}{0.20} & \sev{79.80}{0.55} & \sev{23.67}{3.19} & \sev{18.00}{2.15} & \sev{77.99}{1.06} & \sev{80.40}{0.80} & \sev{20.79}{0.52} & \sev{\B{77.91}}{0.11} & \sev{42.37}{1.80} & \sev{79.83}{0.42} & 59.40 \\
    \qquad DDIM ($32$ steps, $w{=}3,\tau{=}0.3$) & \sev{92.90}{0.18} & \sev{78.20}{0.33} & \sev{19.00}{0.97} & \sev{19.00}{2.50} & \sev{78.35}{1.62} & \sev{79.74}{0.98} & \sev{18.33}{0.61} & \sev{77.60}{0.09} & \sev{41.87}{1.74} & \sev{76.69}{0.38} & 58.17 \\
    \quad \model\preg-8B\instruct & & & & & & & & & & & \\
    \qquad DDPM ($256$ steps, $w{=}3,\tau{=}0.2$) & \sev{92.81}{0.30} & \sev{80.78}{0.69} & \sev{20.33}{2.74} & \sev{18.00}{2.15} & \sev{80.30}{0.93} & \sev{81.72}{0.87} & \sev{19.93}{0.34} & \sev{77.31}{0.08} & \sev{\B{44.80}}{1.55} & \sev{\B{80.82}}{0.28} & 59.68 \\
    \qquad DDIM ($32$ steps, $w{=}4,\tau{=}0.4$) & \sev{92.37}{0.15} & \sev{80.52}{0.69} & \sev{22.33}{3.27} & \sev{17.67}{3.41} & \sev{80.91}{1.61} & \sev{80.24}{0.59} & \sev{18.00}{0.66} & \sev{76.12}{0.07} & \sev{42.58}{1.18} & \sev{76.74}{0.64} & 58.75 \\
    \quad \model-8B\instruct & & & & & & & & & & & \\
    \qquad DDPM ($256$ steps, $w{=}3,\tau{=}0.1$) & \sev{91.96}{0.21} & \sev{78.40}{0.67} & \sev{23.33}{2.67} & \sev{17.00}{1.75} & \sev{82.50}{1.17} & \sev{\B{82.04}}{0.67} & \sev{20.09}{0.63} & \sev{76.57}{0.08} & \sev{42.37}{1.17} & \sev{77.18}{0.38} & 59.14 \\
    \qquad DDIM ($32$ steps, $w{=}3,\tau{=}0.3$) & \sev{91.90}{0.17} & \sev{76.70}{0.67} & \sev{21.67}{3.02} & \sev{14.33}{3.53} & \sev{80.55}{0.58} & \sev{79.47}{0.55} & \sev{19.38}{0.48} & \sev{76.15}{0.08} & \sev{42.93}{1.75} & \sev{75.96}{0.32} & 57.90 \\
    \bottomrule
    \end{tabular}%
    }
\end{table}

\clearpage

\section{Implementation Details and Additional Results for Self-Speculative Decoding}\label{app:cdlm3b-specdec}

\definecolor{specNLD}{HTML}{C2620E}       %
\definecolor{specDrafter}{HTML}{1263B8}   %
\definecolor{specDistilled}{HTML}{B5102F} %
\definecolor{specAccept}{HTML}{1B6E1B}    %

This section gives the protocol (\cref{app:specdec-setup}) and all results:
the models of \cref{sec:specdec} in full (\cref{app:specdec-full}),
the undistilled \model with trained drafters (\cref{app:specdec-undistilled}),
and acceptance training (\cref{app:specdec-lk}).

\subsection{Setup}\label{app:specdec-setup}

\paragraph{Drafting and verification.}
Following the two-pass framework of NLD~\citep[Sec.~3.3]{fu2026nemotron}, each round initializes one block of $L'=32$ draft positions with Gaussian noise in the 16-dimensional latent space and reads out categorical draft tokens from a single denoiser call at $t=1$.
The AR head then scores all drafts causally in one forward pass over the shared backbone and cached prefix, keeps the longest prefix that matches its own greedy tokens, and appends one token of its own.
Draft-only parameters are disabled during verification, so the committed text is the model's greedy AR output, up to \texttt{BF16} near-ties between the decoding and verification kernels.
More denoiser calls per draft raised AL but not tokens per forward pass, so we use one.

\paragraph{Trained drafters.}
Optionally, we train draft-only parameters on Tulu-3 prompts~\citep{lambert2024tulu3} with NLD's LK-hybrid and verifier cross-entropy objective~\citep[Eqs.~6--9]{fu2026nemotron}, keeping the backbone and the AR head frozen.
We train a rank-128 LoRA on the query and value projections together with the diffusion head (\emph{Q/V}, 440M parameters at 3B and 595M at 8B).
At 8B, the 6 of 1{,}000 training steps whose gradient norm is not finite make no update.

\paragraph{Benchmarks.}
GSM8K (1,319 problems, 8-shot chain of thought), HumanEval (164), and MBPP (500) use the prompts of our base-model evaluation.
The end-of-text token is ignored, and an answer ends at the first of the task's stop strings, within 256, 512, and 512 new tokens.

\paragraph{Serving.}
We serve with SGLang~\citep{cdlmapp_zheng2024sglang} in \texttt{BF16}, with greedy decoding, TP1, and one request at a time on one NVIDIA GB200.
AL and accuracy use the numerics of the training code (FP32 diffusion-stream parameters and eager kernels).
Throughput uses FlashAttention-4 and CUDA graphs.

\paragraph{NLD reference.}
NLD-3B-Base and NLD-8B-Base ship without a drafter adapter, so they draft with their own diffusion head.
Every NLD model runs at its released block size: 32 positions holding the last committed token and 31 drafts, one draft fewer per round than \model.
NLD's speculative server searched only the end of each committed block for stop strings and missed earlier ones; we let it search the whole block, as SGLang's own speculative path does, so that it stops in the block that contains the first stop.
NLD's runner times AR and speculative decoding once each, concurrently on two GPUs of one node.

\paragraph{Metrics.}
AL counts the draft tokens that a round adds to the returned answer, i.e., its committed answer tokens minus the verifier's own token, pooled over rounds; drafts accepted past the end of the answer are discarded and not counted.
Repetition loops are trivially predictable and would inflate AL.
On GSM8K, HumanEval, and MBPP, we therefore drop, per model, every prompt whose AR, speculative, or timing output loops (at least half of its 16-grams repeat) or reaches the length cap; accuracy uses all prompts.
Throughput is answer tokens per second of client time, and speedup divides the speculative throughput by the same model's AR throughput on the same GPU and prompts, averaged over two runs on different GPUs.
Paired gains use the prompts that both models keep, with 95\% percentile bootstrap intervals over prompts.

\subsection{Main Results in Full}\label{app:specdec-full}

\Cref{tab:specdec-full} lists throughput, speedup, and AL for the models of \cref{sec:specdec}, and \cref{tab:specdec-acc} lists their accuracy and kept prompts.
For \model\distill, a draft--verify round costs 3.0--3.5 AR decoding steps rather than the ideal two.
Speedups are relative to each model's own AR decoding, which NLD's serving implementation runs about 20\% faster at 3B and 12\% faster at 8B.

\begin{table}[t]
    \centering
    \caption{
    \textbf{Self-speculative decoding in full.}
    Tokens/s of AR and speculative (Spec) decoding, speedup, and AL.
    }
    \label{tab:specdec-full}
    \footnotesize
    \setlength{\tabcolsep}{3.3pt}
    \begin{tabular}{@{}llrrccrrccrrcc@{}}
    \toprule
    & & \multicolumn{4}{c}{\textbf{GSM8K}} & \multicolumn{4}{c}{\textbf{HumanEval}} & \multicolumn{4}{c}{\textbf{MBPP}} \\
    \cmidrule(lr){3-6} \cmidrule(lr){7-10} \cmidrule(lr){11-14}
    & Model & AR & Spec & Speedup & AL & AR & Spec & Speedup & AL & AR & Spec & Speedup & AL \\
    \midrule
    \multirow{3}{*}{3B}
    & \textcolor{specNLD}{NLD-3B-Base} & 453 & 1{,}197 & 2.64$\times$ & 7.63 & 471 & 906 & 1.92$\times$ & 5.01 & 454 & 1{,}078 & 2.37$\times$ & 7.52 \\
    & \textcolor{specDistilled}{\model\distill-3B} & 375 & 1{,}168 & 3.12$\times$ & 9.21 & 399 & 880 & 2.21$\times$ & 5.90 & 374 & 1{,}011 & 2.70$\times$ & 8.36 \\
    & \textcolor{specAccept}{\model\distill-\textit{spec}-3B} & 379 & 1{,}204 & 3.18$\times$ & 9.68 & 399 & 889 & 2.23$\times$ & 6.25 & 375 & 1{,}027 & 2.74$\times$ & 8.85 \\
    \midrule
    \multirow{3}{*}{8B}
    & \textcolor{specNLD}{NLD-8B-Base} & 247 & 865 & 3.50$\times$ & 8.78 & 259 & 652 & 2.51$\times$ & 6.08 & 241 & 750 & 3.11$\times$ & 8.40 \\
    & \textcolor{specDistilled}{\model\distill-8B} & 221 & 864 & 3.92$\times$ & 10.75 & 229 & 543 & 2.37$\times$ & 7.12 & 217 & 707 & 3.26$\times$ & 9.56 \\
    & \textcolor{specAccept}{\model\distill-\textit{spec}-8B} & 222 & 931 & 4.20$\times$ & 11.68 & 228 & 620 & 2.72$\times$ & 7.33 & 220 & 762 & 3.47$\times$ & 10.33 \\
    \bottomrule
    \end{tabular}
\end{table}

\begin{table}[t]
    \centering
    \caption{
    \textbf{Accuracy (\%) of AR / speculative decoding, and kept prompts.}
    GSM8K: flexible / strict extract; code: pass@$1$.
    Kept prompts are out of 1,319, 164, and 500.
    }
    \label{tab:specdec-acc}
    \footnotesize
    \setlength{\tabcolsep}{4pt}
    \begin{tabular}{@{}llccccc@{}}
    \toprule
    & Model & GSM8K flexible & GSM8K strict & HumanEval & MBPP & Kept prompts \\
    \midrule
    \multirow{3}{*}{3B}
    & \textcolor{specNLD}{NLD-3B-Base} & 83.8 / 84.0 & 75.4 / 77.0 & 52.4 / 51.8 & 61.6 / 61.8 & 1{,}306 / 151 / 497 \\
    & \textcolor{specDistilled}{\model\distill-3B} & 82.9 / 82.9 & 79.5 / 79.0 & 56.1 / 56.1 & 58.2 / 59.2 & 1{,}296 / 157 / 498 \\
    & \textcolor{specAccept}{\model\distill-\textit{spec}-3B} & 82.2 / 82.5 & 78.5 / 78.5 & 55.5 / 55.5 & 58.8 / 58.6 & 1{,}298 / 160 / 499 \\
    \midrule
    \multirow{3}{*}{8B}
    & \textcolor{specNLD}{NLD-8B-Base} & 88.8 / 88.1 & 87.3 / 87.0 & 61.6 / 62.2 & 66.2 / 66.8 & 1{,}307 / 149 / 497 \\
    & \textcolor{specDistilled}{\model\distill-8B} & 86.4 / 86.8 & 85.2 / 85.7 & 56.1 / 57.9 & 64.0 / 63.8 & 1{,}311 / 158 / 497 \\
    & \textcolor{specAccept}{\model\distill-\textit{spec}-8B} & 86.4 / 86.1 & 85.1 / 84.8 & 57.3 / 56.7 & 63.8 / 64.4 & 1{,}310 / 159 / 497 \\
    \bottomrule
    \end{tabular}
\end{table}

\subsection{Undistilled Model and Trained Drafters}\label{app:specdec-undistilled}

The undistilled \model-3B and \model-8B can draft with their own diffusion head or with a trained Q/V adapter (\cref{tab:specdec-undistilled}).

\textbf{Distillation helps more than a trained drafter at 3B.}
The adapter raises \model-3B's AL by 13.1\%, 5.4\%, and 7.2\% on GSM8K, HumanEval, and MBPP ([12.2, 14.0], [3.1, 7.6], and [5.8, 8.9]).
The distilled student, with its own head and no adapter, gains 21.8\%, 20.1\%, and 15.0\% over \model-3B ([20.1, 23.5], [8.6, 31.7], and [11.8, 18.1]).

\textbf{At 8B, the two are closer.}
The adapter raises \model-8B's AL by 16.1\%, 9.4\%, and 14.3\% ([15.1, 17.1], [6.1, 13.0], and [12.4, 16.1]), and distillation by 22.4\%, 7.9\%, and 19.9\% ([20.9, 24.0], [$-$0.5, 16.8], and [16.7, 23.2]).
Compared directly, \model\distill-8B accepts 5.4\% and 4.9\% more than \model-8B with the adapter on GSM8K and MBPP ([4.0, 6.8] and [2.0, 7.7]) and 1.6\% fewer on HumanEval ([$-$9.3, 6.4]), where it also decodes slower (2.37$\times$ against 2.73$\times$).

\textbf{The adapter leaves the output unchanged.}
It is disabled during verification, so \model-3B and \model-8B solve exactly the same problems with and without it.

\begin{table}[t]
    \centering
    \caption{
    \textbf{Undistilled \model: own head, trained Q/V adapter, and distillation.}
    AL gain is over the undistilled model with its own head, on the prompts both keep.
    }
    \label{tab:specdec-undistilled}
    \footnotesize
    \setlength{\tabcolsep}{4pt}
    \begin{tabular}{@{}llccc@{}}
    \toprule
    & & \textbf{GSM8K} & \textbf{HumanEval} & \textbf{MBPP} \\
    \midrule
    \multirow{4}{*}{3B}
    & \model-3B: AL / speedup & 7.55 / 2.54$\times$ & 4.92 / 1.77$\times$ & 7.24 / 2.38$\times$ \\
    & \textcolor{specDrafter}{+ Q/V adapter}: AL gain / speedup & +13.1\% / 2.93$\times$ & +5.4\% / 1.94$\times$ & +7.2\% / 2.46$\times$ \\
    & \textcolor{specDistilled}{\model\distill-3B}: AL gain / speedup & +21.8\% / 3.12$\times$ & +20.1\% / 2.21$\times$ & +15.0\% / 2.70$\times$ \\
    & AR accuracy (\%), \model-3B / \model\distill-3B & 81.2 / 82.9 & 49.4 / 56.1 & 56.2 / 58.2 \\
    \midrule
    \multirow{4}{*}{8B}
    & \model-8B: AL / speedup & 8.80 / 3.38$\times$ & 6.60 / 2.58$\times$ & 7.96 / 2.92$\times$ \\
    & \textcolor{specDrafter}{+ Q/V adapter}: AL gain / speedup & +16.1\% / 3.79$\times$ & +9.4\% / 2.73$\times$ & +14.3\% / 3.27$\times$ \\
    & \textcolor{specDistilled}{\model\distill-8B}: AL gain / speedup & +22.4\% / 3.92$\times$ & +7.9\% / 2.37$\times$ & +19.9\% / 3.26$\times$ \\
    & AR accuracy (\%), \model-8B / \model\distill-8B & 84.8 / 86.4 & 56.1 / 56.1 & 64.2 / 64.0 \\
    \bottomrule
    \end{tabular}
\end{table}

\subsection{Acceptance Training of the Distilled Model}\label{app:specdec-lk}

Instead of training a separate drafter, we continue the distillation with an acceptance objective that updates the whole backbone; we call the result \model\distill-\textit{spec}.
Let $\phi$ denote the parameters of the student, which starts from \model\distill, and $\bar\phi$ those of a frozen copy of \model\distill.
We keep the distillation objective and add two terms.
An LK-hybrid loss aligns the student's one-call draft with the AR head of $\bar\phi$, and a KL term holds the student's AR distribution to that of $\bar\phi$, so that the verifier, and hence the decoded output, is preserved:
\begin{align}
    \cL_{\text{acc}}
    =
    \lambda_{\text{DMD}}\,\cL_{\text{DMD}}
    +
    \lambda_{\text{LK}}\,\cL_{\text{LK}}
    +
    \hat\lambda_{\text{ret}}\,\cL_{\text{ret}}.
    \label{eq:specdec-acc-loss}
\end{align}
Training updates the shared transformer and the diffusion head; as in distillation (\cref{app:dmd-details}), the embedding matrix $\Emb$, the noise schedule, and the AR head stay frozen.

\paragraph{Distribution matching.}
$\cL_{\text{DMD}}$ is the student objective of \cref{app:distillation-details} (\cref{alg:dmd-training}), with the auxiliary and reference denoisers and the setting of the distillation that produced \model\distill: two phases, no self-conditioning, and segments that start from fresh student rollouts.
Acceptance training thus continues that distillation.

\paragraph{Acceptance.}
Each LK example takes a context $\c$ from the generations of an earlier 3B DMD student on math and code prompts, mostly GSM8K- and HumanEval-style, and drafts the next block with one call from pure noise, as in serving (\cref{app:specdec-setup}).
Both sizes use these contexts; the two runs averaged into each \model\distill-\textit{spec} model use two versions of the context set.
Test and dev problems are excluded by exact match; a 13-gram scan finds 3, 1, and 54 near-duplicates of GSM8K, HumanEval, and MBPP test problems among the contexts, and excluding those problems leaves the AL gains of acceptance training unchanged.
Slot $j\in\{1,\ldots,L'\}$ has draft distribution $\hat p^j:=\x^j_\phi(\z_1,1;\c)$ with $\z_1\sim\mathcal{N}(\zero,\I)$ and draft token $\hat v^j:=\argmax_v \hat p^j(v)$.
The verifier reads all slots in one causal pass and gives $p^j:=\x^{|\c|+j}_{\bar\phi_{\text{AR}}}(\tilde\x^{<|\c|+j})$, where $\tilde\x$ is $\c$ followed by the draft tokens.
If the first $a$ draft tokens equal the verifier's greedy tokens $v^j_\star:=\argmax_v p^j(v)$ and slot $a+1$ does not, the loss covers the accepted prefix and the first rejected slot, $\mathcal{A}=\{1,\ldots,\min(a+1,L')\}$.
Following NLD~\citep[Eqs.~6--9]{fu2026nemotron}, let $U^j$ be the union of the 200 most probable tokens under $p^j$ and under $\hat p^j$; let $p^j_\tau$ and $\hat p^j_\tau$ be $p^j$ and $\hat p^j$ restricted to $U^j$ and renormalized at temperature $\tau_{\text{LK}}=3$, i.e., $p^j_\tau(v)\propto p^j(v)^{1/\tau_{\text{LK}}}$ for $v\in U^j$; and let $\hat p^j_1$ be the same restriction of $\hat p^j$ at temperature 1. Then
\begin{align}
    \cL_{\text{LK}}
    =
    \E\Big[
    \tfrac{1}{|\mathcal{A}|}\tsum_{j\in\mathcal{A}}
    \nu^j\,\KL(p^j_\tau\,\|\,\hat p^j_\tau)
    +
    (1-\nu^j)\,\mathrm{TV}(p^j_\tau,\hat p^j_\tau)
    -
    \log \hat p^j_1(v^j_\star)
    \Big],
    \qquad
    \nu^j=\exp\!\big(-\eta\,\sg(\omega^j)\big),
    \label{eq:specdec-lk}
\end{align}
where $\omega^j=\tsum_{v\in U^j}\min\big(p^j_\tau(v),\hat p^j_\tau(v)\big)=1-\mathrm{TV}(p^j_\tau,\hat p^j_\tau)$ is the probability that a token sampled from $\hat p^j_\tau$ is accepted, and $\eta=0.5$; the weight thus shifts from the KL term toward total variation as the draft approaches the verifier.
The retention term is the forward KL from the AR distribution of $\bar\phi$ to that of the student, averaged over the set $\mathcal{R}$ of the last 128 positions of $\tilde\x$, i.e., the $L'$ draft slots and the 96 context tokens before them:
\begin{align}
    \cL_{\text{ret}}
    =
    \E\Big[
    \tfrac{1}{|\mathcal{R}|}\tsum_{l\in\mathcal{R}}
    \KL\big(
    \x^l_{\bar\phi_{\text{AR}}}(\tilde\x^{<l})
    \,\big\|\,
    \x^l_{\phi_{\text{AR}}}(\tilde\x^{<l})
    \big)
    \Big].
    \label{eq:specdec-ret}
\end{align}

\paragraph{Weights and optimization.}
We set $\lambda_{\text{LK}}=0.25$ and $\lambda_{\text{ret}}=12$, and cap the retention weight at
\begin{align}
    \hat\lambda_{\text{ret}}
    =
    \min\Big(
    \lambda_{\text{ret}},\;
    3\,\frac{\|\nabla_\phi(\lambda_{\text{DMD}}\cL_{\text{DMD}}+\lambda_{\text{LK}}\cL_{\text{LK}})\|}{\|\nabla_\phi\cL_{\text{ret}}\|}
    \Big),
    \label{eq:specdec-ret-cap}
\end{align}
with both gradient norms taken from the previous update.
Without the cap, once the AR distribution drifted, the retention gradient grew until it dominated the clipped update, and training diverged.
Each update draws 64 DMD examples and 128 drafts.
The drafts form chains of four rounds: the first round drafts after a prefix of up to 1,664 tokens that keeps the prompt's BOS token, and each later round drafts after the tokens that the previous round committed, as in serving.
The optimizers are those of \cref{app:dmd-details}, except that the student learning rate is doubled to $7.2\times10^{-7}$; the drafts never use self-conditioning, as in serving.
Each \model\distill-\textit{spec} model is a uniform weight average of two runs with $\lambda_{\text{DMD}}=1$ and $0.5$, each trained for 600 updates from \model\distill{} while the learning rate decays linearly to zero.
At 8B, the teacher-generated DMD data has 16,384 contexts instead of 65,536, and the fit of the query noise levels and the LK contexts are reused from 3B.

\paragraph{Results.}
\Cref{tab:specdec-lk} gives the gains.
Acceptance training raises AL by 5.2\%, 6.0\%, and 6.1\% at 3B and by 8.5\%, 4.8\%, and 8.1\% at 8B on GSM8K, HumanEval, and MBPP; the HumanEval intervals include zero at both sizes.
It leaves AR accuracy unchanged on all three benchmarks at both sizes (sign tests, $p\geq0.10$ at 3B and $p\geq0.69$ at 8B).

\begin{table}[t]
    \centering
    \caption{
    \textbf{Acceptance training: gain of \model\distill-\textit{spec} over \model\distill.}
    AL over the prompts both models keep, with 95\% bootstrap intervals; speedup as in \cref{tab:specdec-full}.
    }
    \label{tab:specdec-lk}
    \footnotesize
    \setlength{\tabcolsep}{4pt}
    \begin{tabular}{@{}llccc@{}}
    \toprule
    & & \textbf{GSM8K} & \textbf{HumanEval} & \textbf{MBPP} \\
    \midrule
    \multirow{3}{*}{3B}
    & AL, \model\distill-3B $\to$ \model\distill-\textit{spec}-3B & 9.21 $\to$ 9.68 & 5.92 $\to$ 6.28 & 8.36 $\to$ 8.87 \\
    & AL gain & +5.2\% {\scriptsize[4.0, 6.2]} & +6.0\% {\scriptsize[$-$0.2, 12.1]} & +6.1\% {\scriptsize[4.1, 8.2]} \\
    & Speedup & 3.12$\times$ $\to$ 3.18$\times$ & 2.21$\times$ $\to$ 2.23$\times$ & 2.70$\times$ $\to$ 2.74$\times$ \\
    \midrule
    \multirow{3}{*}{8B}
    & AL, \model\distill-8B $\to$ \model\distill-\textit{spec}-8B & 10.76 $\to$ 11.68 & 7.31 $\to$ 7.66 & 9.56 $\to$ 10.33 \\
    & AL gain & +8.5\% {\scriptsize[7.3, 9.8]} & +4.8\% {\scriptsize[$-$0.1, 10.4]} & +8.1\% {\scriptsize[5.8, 10.4]} \\
    & Speedup & 3.92$\times$ $\to$ 4.20$\times$ & 2.37$\times$ $\to$ 2.72$\times$ & 3.26$\times$ $\to$ 3.47$\times$ \\
    \bottomrule
    \end{tabular}
\end{table}

\clearpage

\end{document}